\documentclass[10pt,journal,compsoc,x11names]{IEEEtran}

\DeclareUnicodeCharacter{FF0C}{,}
\usepackage{amsmath, amsfonts}
\usepackage{algorithmic}
\usepackage{algorithm}
\usepackage{array}
\usepackage[caption=false,font=normalsize,labelfont=sf,textfont=sf]{subfig}
\usepackage{textcomp}
\usepackage{stfloats}
\usepackage{url}
\usepackage{verbatim}
\usepackage{graphicx}
\usepackage{cite}
\usepackage{multirow}
\usepackage[T1]{fontenc}
\usepackage[utf8]{inputenc}

\usepackage{hyperref}
\usepackage{fontawesome5}
\usepackage{forest}
\useforestlibrary{edges}
\usepackage{makecell}
\usetikzlibrary{shadows}
\usepackage{amsmath}

\usepackage{amssymb}
\usepackage{cleveref}
\usepackage{tikz}
\usepackage{amssymb}

\usetikzlibrary{trees,positioning,shapes,shadows,arrows.meta}

\definecolor{hidden-red}{RGB}{205, 44, 36}
\definecolor{hidden-blue}{RGB}{194,232,247}
\definecolor{hidden-orange}{RGB}{243,202,120}
\definecolor{hidden-green}{RGB}{34,139,34}
\definecolor{hidden-pink}{RGB}{255,245,247}
\definecolor{hidden-black}{RGB}{20,68,106}
\definecolor{purple}{RGB}{144,153,196}
\definecolor{yellow}{RGB}{255,228,123}
\definecolor{hidden-yellow}{RGB}{255,248,203}
\definecolor{tkcolor}{RGB}{224,223,255}
\definecolor{darkblue}{rgb}{0, 0.40, 0.75}
\newcommand{\eg}{\textit{e.g.,}\xspace}

\definecolor{lightblue}{RGB}{220,235,250}
\hypersetup{colorlinks=true, citecolor=darkblue, linkcolor=darkblue, urlcolor=darkblue}

\definecolor{refred}{HTML}{BD114A}
\definecolor{refblue}{HTML}{5E7AC4}
\definecolor{textgreen}{HTML}{508D69}
\definecolor{textred}{HTML}{FF5555}
\definecolor{textyellow}{HTML}{F0A202}
\definecolor{frameblue}{HTML}{9DBDFF}
\hypersetup{
    colorlinks=true,
    linkcolor=refblue,     % 普通链接
    citecolor=refred,      % 参考文献引用颜色
    urlcolor=.
}

\crefformat{section}{Sec. #2#1#3} 
\crefformat{subsection}{Sec. #2#1#3}
\crefformat{subsubsection}{Sec. #2#1#3}

\usepackage{amsthm}
\usepackage{amsmath}%
\usepackage{MnSymbol}%

\makeatletter
\newtheorem*{rep@theorem}{\rep@title}
\newcommand{\newreptheorem}[2]{%
\newenvironment{rep#1}[1]{%
 \def\rep@title{#2 \ref{##1}}%
 \begin{rep@theorem}}%
 {\end{rep@theorem}}}
\makeatother

\newreptheorem{theorem}{Theorem}

\newreptheorem{lemma}{Lemma}

\newreptheorem{proposition}{proposition}

\usepackage{bm}
\usepackage{xspace}

\usepackage{booktabs}
\usepackage{xcolor,pifont}
\usepackage{bbding}
\usepackage{colortbl}
\usepackage{tcolorbox}
\tcbuselibrary{skins, breakable, theorems}
\usepackage{soul}

\definecolor{TakeawayBg}{HTML}{F4F5FF} % 盒子浅紫色
\definecolor{TakeawayLabel}{HTML}{666666} % 标签灰色
\definecolor{TakeawayText}{HTML}{000000}  % 正文文字色
\sethlcolor{TakeawayBg}

\definecolor{hidden-draw}{RGB}{20,68,106}
\definecolor{hidden-pink}{RGB}{255,245,247}
\definecolor{c0}{HTML}{7d9bbc} % Soft mint green
\definecolor{c1}{HTML}{b3d2f4} % Soft sky blue98b0a8
\definecolor{c2}{HTML}{92d9da} % Warm coral orange
\definecolor{c3}{HTML}{c1f6e5}  % Dusty rose

\definecolor{light-blue}{RGB}{206,225,242}

\tcbset{
  takeaway/.style={
    enhanced,
    colback=TakeawayBg,
    colframe=TakeawayText,
    arc=2mm,
    boxrule=0.8pt,
    left=3mm,right=3mm,
    top=2mm,bottom=2mm,
    fonttitle=\bfseries,
    coltitle=white,
    attach boxed title to top left={
      yshift=-2mm,
      xshift=3mm
    },
    boxed title style={
      colback=TakeawayLabel,
      colframe=TakeawayLabel,
      arc=1mm,            % ← 将标题也设置为圆角
      left=2mm,right=2mm,
      top=0.1mm,bottom=0.1mm
    },
  }
}

\usepackage[inline,shortlabels]{enumitem}

\newcommand{\Text}{
  \begin{tikzpicture}[baseline=(char.base)]
    \fill[textgreen!50] (0,0) circle (0.18);
    \node[white, font=\scriptsize\rmfamily, inner sep=0pt] (char) at (0,0) {T};
  \end{tikzpicture}
}
\newcommand{\Image}{
  \begin{tikzpicture}[baseline=(char.base)]
    \fill[textred!50] (0,0) circle (0.18);
    \node[white, font=\scriptsize\rmfamily, inner sep=0pt] (char) at (0,0) {I};
  \end{tikzpicture}
}
\newcommand{\Video}{
  \begin{tikzpicture}[baseline=(char.base)]
    \fill[textyellow!50] (0,0) circle (0.18);
    \node[white, font=\scriptsize\rmfamily, inner sep=0pt] (char) at (0,0) {V};
  \end{tikzpicture}
}
\newcommand{\Embed}{
  \begin{tikzpicture}[baseline=(char.base)]
    \fill[frameblue!50] (0,0) circle (0.18);
    \node[white, font=\scriptsize\rmfamily, inner sep=0pt] (char) at (0,0) {E};
  \end{tikzpicture}
}
\newcommand{\Result}{
  \begin{tikzpicture}[baseline=(char.base)]
    \fill[textgreen!50] (0,0) circle (0.18);
    \node[white, font=\scriptsize\rmfamily, inner sep=0pt] (char) at (0,0) {A};
  \end{tikzpicture}
}
\newcommand{\Format}{
  \begin{tikzpicture}[baseline=(char.base)]
    \fill[textred!50] (0,0) circle (0.18);
    \node[white, font=\scriptsize\rmfamily, inner sep=0pt] (char) at (0,0) {F};
  \end{tikzpicture}
}
\newcommand{\Efficient}{
  \begin{tikzpicture}[baseline=(char.base)]
    \fill[textyellow!50] (0,0) circle (0.18);
    \node[white, font=\scriptsize\rmfamily, inner sep=0pt] (char) at (0,0) {E};
  \end{tikzpicture}
}
\newcommand{\Process}{
  \begin{tikzpicture}[baseline=(char.base)]
    \fill[frameblue!50] (0,0) circle (0.18);
    \node[white, font=\scriptsize\rmfamily, inner sep=0pt] (char) at (0,0) {P};
  \end{tikzpicture}
}

\newcommand{\Filtering}{
  \begin{tikzpicture}[baseline=(char.base)]
    \fill[textgreen!50] (0,0) circle (0.18);
    \node[white, font=\scriptsize\rmfamily, inner sep=0pt] (char) at (0,0) {F};
  \end{tikzpicture}
}

\newcommand{\Iterative }{
  \begin{tikzpicture}[baseline=(char.base)]
    \fill[textyellow!50] (0,0) circle (0.18);
    \node[white, font=\scriptsize\rmfamily, inner sep=0pt] (char) at (0,0) {I};
  \end{tikzpicture}
}
\newcommand{\Correction}{
  \begin{tikzpicture}[baseline=(char.base)]
    \fill[frameblue!50] (0,0) circle (0.18);
    \node[white, font=\scriptsize\rmfamily, inner sep=0pt] (char) at (0,0) {C};
  \end{tikzpicture}
}

\newcommand{\ghlink}[1]{\href{#1}{\faIcon{github}}}
\newcommand{\hflink}[1]{\href{#1}{\includegraphics[height=1em]{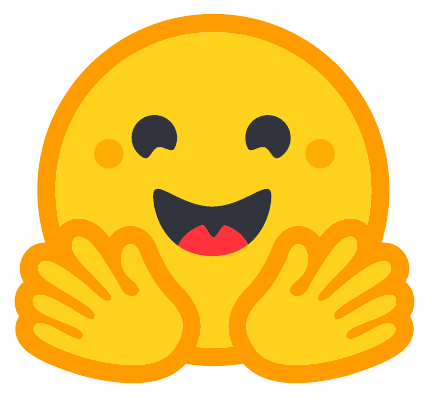}}}
\newcommand{\Yes}{\textcolor{textgreen}{\ding{51}}}
\newcommand{\No}{\textcolor{textred}{\ding{55}}}

\begin{document}

\title{Agentic Environment Engineering for Large Language Models: A Survey of Environment Modeling, Synthesis, Evaluation, and Application}
% 
% , and Agent-Environment Co-Evolution
% : \\ A Systematic Perspective on Understanding, Building, and Co-Evolving Environments

\author{
Jiachun~Li$^{*}$, 
Zhuoran~Jin$^{*}$, 
Tianyi~Men$^{*}$, 
Yupu~Hao$^{*}$, 
Kejian~Zhu$^{*}$,
Lingshuai~Wang$^{*}$, 
Dongqi~Huang$^{*}$, 
Longxiang~Wang$^{*}$,
Shengjia~Hua$^{*}$, 
Lu~Wang$^{*}$, 
Jinshan~Gao$^{*}$,
Hongbang Yuan,
Ruilin Xu,
\\
Kang Liu,
Jun Zhao$^\dagger$

% Dong~Yu\orcidA{},~\IEEEmembership{Fellow,~IEEE}, Qianli~Ma$^\dagger$,~\IEEEmembership{Member,~IEEE}% <-this % stops a space
\thanks{Version: v1 (major update on May 19, 2026)}
\thanks{$^\dagger$Corresponding author: Jun Zhao.}
\thanks{$^{*}$These authors contributed equally to this work.}
\thanks{Jiachun Li, Zhuoran Jin, Tianyi Men, Yupu Hao, Kejian Zhu, Lingshuai Wang, Dongqi Huang, Longxiang Wang, Shengjia Hua, Lu Wang, Jinshan Gao, Hongbang Yuan, Ruilin Xu, Kang Liu, Jun Zhao are with the Key Laboratory of Cognition and Decision Intelligence for Complex Systems, Institute of Automation, Chinese Academy of Sciences, Beijing 100191, China (E-mail: jiachun.li@nlpr.ia.ac.cn, zhuoran.jin@nlpr.ia.ac.cn, tianyi.men@nlpr.ia.ac.cn, kliu@nlpr.ia.ac.cn, jzhao@nlpr.ia.ac.cn ).}}
% <-this % stops a space

\markboth{Journal of \LaTeX\ Class Files, January 2025}%
{Shell \MakeLowercase{\textit{et al.}}: A Sample Article Using IEEEtran.cls for IEEE Journals}


\IEEEtitleabstractindextext{

\begin{abstract}
Environments serve as interactive systems for large language model (LLM) based agents across diverse scenarios and play a crucial role in driving the continual evolution of model capabilities. Despite this importance, existing work lacks a systematic categorization and deep analysis. This paper systematically studies current researches on agentic environments from the perspective of the environment engineering lifecycle, covering their modeling, synthesis, evaluation and application. Specifically, the paper first introduces representative environments from the perspectives of eight attributes and eight domains, providing detailed analyses of their development paths and highlighting their core capabilities.
Second, for automated environment synthesis, two paradigms are introduced, such as symbolic synthesis and neural synthesis. This paper also shows different environment evaluation methods in each paradigm. Thirdly, the corresponding environment applications from the perspective of agent-environment co-evolution are discussed. 
In specific, the paper characterizes the primary pathways for agent evolution in dynamic environments from four complementary perspectives: memory-centric experience evolution, orchestration-centric workflow evolution, trajectory-centric offline evolution, and exploration-centric online evolution.
And three paradigms of environment evolution are identified, namely neural-driven, difficulty-driven, and scaling-driven approaches. At last, several promising future directions are discussed, including Environment-as-a-Service, Multi-agent Environments, and Neural-Symbolic Environments.

\end{abstract}

}

\maketitle

% table 6 CHECK

\section{Introduction}
\label{sec:introduction}

\IEEEPARstart{L}{arge} language models (LLMs) have demonstrated remarkable performance across a wide range of tasks, from basic factual question answering to complex reasoning tasks, with their capabilities continuing to evolve \cite{Qwen3,DeepSeek-R1,GEM,HLE}. Recently, representative advanced models such as GPT-5.4 \cite{gpt-5.4}, Gemini-3.1-Pro \cite{gemini3.1}, and Kimi K2.5 \cite{Kimi_K2.5} have demonstrated strong agentic capabilities, particularly in complex tool calling, long-horizon planning, and self-improving \cite{ToolRL, TravelPlanner, Self-Refine}. To rigorously benchmark and further unlock the potential of these models, researchers draw inspiration from how humans continuously acquire experience and develop capabilities through interaction with environments, and accordingly design \textbf{agentic environments} for LLM agents.

% To evaluate and improve these models, researchers draw inspiration from the evolutionary process of human interaction with the natural environment to design \textbf{agentic environments} as interactive counterparts for the models.

An agentic environment is a dynamic system that simulates a real-world scenario that models can interact with. While models require interaction with the real world to acquire human-level skills, such interactions are often infeasible in practice due to multiple limitations, including high costs, safety risks, and data privacy concerns \cite{Agent_world_model, RWKU, Troublemaker}. Furthermore, real-world environments are inherently irreproducible, rendering operational failures exceptionally costly. Conversely, manually engineering simulated alternatives is not only resource-intensive but also suffers from constrained scenario coverage. For example, allowing an autonomous driving system to directly control a vehicle on real roads is not only highly inefficient but also poses significant risks of accidents. Agentic environments effectively address these challenges by incorporating sophisticated tools and reward signals that closely approximate real-world conditions. In addition, by interacting with simulated environments, large-scale trajectory data can be generated. In this way, continual evolution could be thereby enabled. Consequently, the agentic environment has become an inseparable twin of the agent throughout different stages, such as capability evaluation \cite{DAComp, WebShop, SWE-Bench}, inference-time reasoning enhancement \cite{ReAct,Fixing_the_Broken_Compass,Omni-Reward}, and reinforcement learning training \cite{GRPO,DAPO,reinforce_internal,towards_agentic_llm}.

% An agentic environment can be viewed as a stochastic dynamical system that interacts with the model. The model generates experience trajectories by acting according to instructions and tools provided in the environment, while feedback signals from the environment indicate the optimization direction of the corresponding trajectories. This process enables the environment to support multiple applications, including capability evaluation, inference-time reasoning enhancement, and reinforcement learning training. This interaction paradigm enables the environment to support multiple applications for the model, including capability evaluation \cite{AgentsCourt, MMR-V, MMR-Life, DAComp}, inference-time reasoning enhancement \cite{ReAct,Fixing_the_Broken_Compass,Omni-Reward,planning_survey}, and reinforcement learning training \cite{GRPO,DAPO,reinforce_internal,towards_agentic_llm}.

\begin{figure*}[t]
    \centering
    \includegraphics[width=\textwidth]{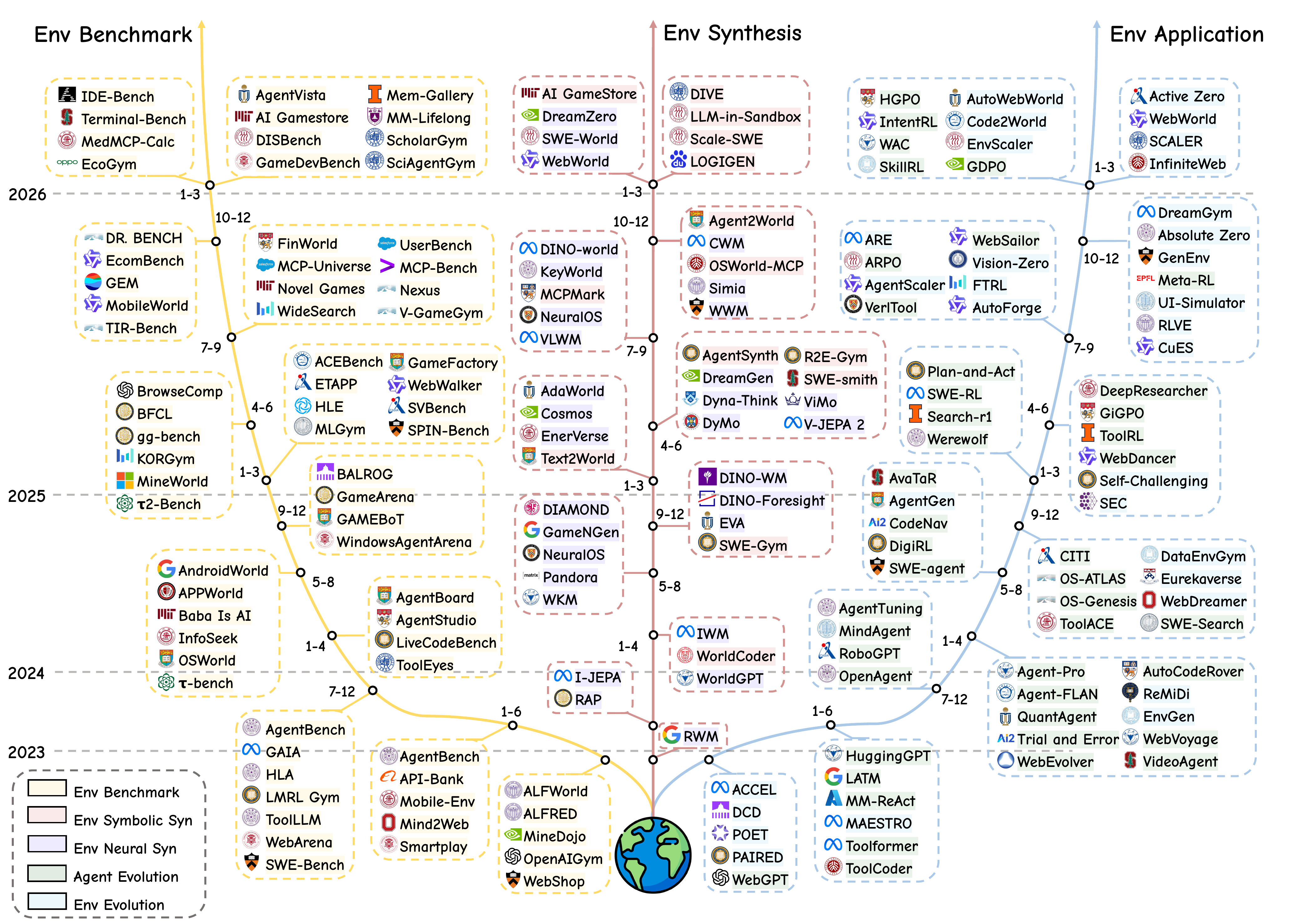}
    \caption{An overview of agentic environment engineering.} 
    \label{fig:env_attribute}
\end{figure*}

Nevertheless, agentic environments have not yet been systematically organized and analyzed in the recent literature \cite{survey_agent_reason, planning_survey, Commonsense_Survey}. To bridge this gap, this paper systematically reviews the whole lifecycle of \textbf{agentic environment engineering}, including environment modeling, automated environment synthesis, evaluation of environment quality, and environment applications.
Specifically, our survey is organized around the following three research questions:
\begin{itemize}
    \item \textbf{RQ1:} What are the key characteristics and categories of agentic environments? (See \cref{sec:env_attr} and \cref{sec:env_domain})
    \item \textbf{RQ2:} How can agentic environments be systematically constructed and evaluated? (See \cref{sec:env_syn})
    \item \textbf{RQ3:} How do agentic environments facilitate the closed-loop co-evolution process of agent and environment? (See \cref{sec:agent_evo} and \cref{sec:env_evo})
\end{itemize}

\textbf{To answer RQ1}, this paper begins by formally defining key concepts related to environments, clarifying their core components and mechanisms, and introducing the main focus of this survey. Then, existing environments are analyzed from the perspective of their attributes, including representation, feedback, timing, observability, stochasticity, continuity, modality, and cardinality.
Through comparative analysis, we observe that existing environments are inadequately suited for multi-agent settings.
Moreover, there is significant potential for improvement in striking a balance between the engineering reliability of symbolic systems and the infinite generative scalability of neural models.

To fully demonstrate the types of existing environments, the paper further introduces current benchmarks from the task domain. Specifically, we classify all environments into eight representative domains: GUI \cite{WorkArena, OSWorld, WebShop}, Deep Research \cite{SimpleQA, GAIA, BrowseComp}, Embodied \cite{ALFRED, ALFWorld, ScienceWorld}, Game \cite{GameArena, BabaIsAI, GAMEBENCH}, Tool \cite{ToolBench, tau-bench, API-Bank}, Code \cite{MBPP, Terminal-Bench, KernelBench}, Domain-Specific \cite{MedAgentBench, ScienceAgentBench, DSBench}, and Cross-Domain \cite{Openai_gym, GEM, AgentBench}. Related works are categorized into each domain, and detailed descriptions of their development trajectories and core agent capabilities are provided.

\textbf{To answer RQ2}, the paper emphasizes the environment synthesis methods and quality evaluation. \textbf{On environment synthesis}, all existing works are divided into two types based on the environment synthesis forms: 

\begin{itemize}

\item \textit{Symbolic Synthesis} uses symbolic rules with different forms (\textit{e.g.}, code) to synthesize environments, employing verifiable rubrics to ensure reliable environmental feedback. Following this approach, this survey identifies an evolution of scalability.
Specifically, the transition from task-driven to real-world–driven synthesis paradigms, and ultimately to \textit{de novo} synthesis methods, signifies a continuous expansion in both synthesis flexibility and the breadth of the environmental space.
\item \textit{Neural Synthesis} parameterizes the environment using a neural network (particularly a world model), and constructs function mappings based on the model's parameters to enable dynamic interactions with the agent. This paper systematically separates neural synthesis into three fine-grained paradigms based on the form of environment representation: Pixel-level, Word-level, and Latent-level Modeling.

\end{itemize}

\textbf{On environment evaluation}, the paper discusses commonly used quality control techniques in four perspectives: correctness, diversity, complexity, and fidelity. We highlight that, while correctness has been well-established and there exist several evaluation frameworks, the study of evaluating diversity, complexity, and fidelity remains under-researched \cite{AgentsCourt, MMR-V, MMR-Life}. Actually, these dimensions are crucial not only for assessing the environmental reliability but also for ensuring the environments' effectiveness in training agents across a wide range of tasks.

\textbf{To answer RQ3}, this survey discusses the application of environments from two complementary perspectives: agent evolution and environment evolution. 
\textbf{On agent evolution}, the agent’s capabilities advance through continuous interaction with the environment. Existing agent evolutionary methods are classified into four categories:

% Memory-Centric Experience Evolution, Orchestration-Centric Workflow Evolution, Trajectory-Centric Offline Evolution, and Exploration-Centric Online Evolution. 

\begin{itemize}

\item \textit{Memory-Centric Experience Evolution} enhances task-processing capabilities by accumulating and leveraging experiences from past interactions. 

\item \textit{Orchestration-Centric Workflow Evolution} focuses on the dynamic coordination and adaptation of task sequences and multi-agent interactions within a workflow, thereby optimizing performance and efficiency across diverse environments. 

\item \textit{Trajectory-Centric Offline Evolution} generates high-quality task interaction trajectories, which subsequently facilitate refining agent behavior by simulating high-fidelity and multifaceted environments. 

\item \textit{Exploration-Centric Online Evolution} leverages reinforcement learning  to refine agent capabilities dynamically, enabling the agent to continuously adapt and optimize its performance.

\end{itemize}

\textbf{On environment evolution}, this survey focuses on how environments are progressively evolved to support the continuous improvement of agent capability. We categorize this paradigm into three types based on the aspects of environmental changes:

\begin{itemize}

\item \textit{Neural-Driven Evolution} adjusts the environment’s internal parameters to simulate a variety of potential states and outcomes, helping the agent to prepare for diverse scenarios. 

\item \textit{Difficulty-Driven Evolution} adapts the complexity of tasks in the environment to match the agent's evolving capabilities, typically through curriculum learning. 

\item \textit{Scaling-Driven Evolution} aims to expand the range of environments by increasing scenario diversity or introducing entirely new structures, which pushes agents towards broader generalization across different contexts.

\end{itemize}

Finally, current challenges and future research directions are highlighted. Specifically, this paper emphasizes Environment-as-a-Service as a potential paradigm for standardized, scalable, and reproducible environment deployment. We then discuss the ongoing evolution of environments toward dynamic, long-horizon, open-ended, multimodal, and multi-agent settings, which impose fundamentally new requirements on agent capabilities. Moreover, emerging directions such as Neural-Symbolic Environments, sim-to-real alignment, and agent–environment co-evolution are introduced for developing more reliable and adaptive agent systems. Looking ahead, the paper argues that establishing a scientific foundation for environment engineering, including environment scaling laws and environment–capability relationships, will be crucial for advancing the next generation of LLM agents.

\vspace{3pt}

In summary, this survey is organized as follows: 
\begin{itemize}
    \item \cref{sec:prelimi} introduces the necessary preliminaries and highlights the gap between environment engineering and traditional data engineering.
    \item \cref{sec:env_attr} introduces the key properties of environments, provides their formal definitions, and offers detailed explanations with concrete examples.
    \item \cref{sec:env_domain} categorizes existing environments from the perspective of tasks, and provides a detailed overview of the specific tasks and representative environments in domains such as GUI, Deep Research, and Embodied.
    \item \cref{sec:env_syn} introduces existing methods of automatic environment synthesis in both symbolic and neural perspectives, and analyzes their respective development pathways and evaluating solutions.
    \item \cref{sec:agent_evo} presents how environments facilitate agent evolution, including Experience Utilization, Agentic Workflow Design, Synthetic Data Generation, and Reinforcement Learning Optimization.
    \item \cref{sec:env_evo} focuses on the main paradigms of environment evolution, explaining how environments adapt to the agent’s training from three perspectives (neural, difficulty, and scaling).
    \item \cref{sec:future} summarizes the main challenges of existing environment-related research and discusses potential future directions.
\end{itemize}

\tikzstyle{my-box}=[
rectangle,
draw=hidden-black,
rounded corners,
text opacity=1,
minimum height=1.5em,
minimum width=5em,
inner sep=2pt,
align=left,
fill opacity=.5,
]
\tikzstyle{leaf3}=[
my-box,
minimum height=1.5em,
fill=yellow!32,
text=black,
align=left,
font=\normalsize,
inner xsep=5pt,
inner ysep=4pt,
align=left,
text width=45em,
]
\tikzstyle{leaf6}=[
my-box,
minimum height=1.5em,
fill=purple!30,
text=black,
align=left,
font=\normalsize,
inner xsep=5pt,
inner ysep=4pt,
]
\tikzstyle{leaf4}=[
my-box,
minimum height=1.5em,
fill=hidden-blue!57,
text=black,
align=left,
font=\normalsize,
inner xsep=5pt,
inner ysep=4pt,
]
\tikzstyle{leaf2}=[
my-box,
minimum height=1.5em,
fill=hidden-green!20,
text=black,
align=left,
font=\normalsize,
inner xsep=5pt,
inner ysep=4pt,
]
\tikzstyle{leaf}=[
my-box,
minimum height=1.5em,
fill=hidden-red!20,
text=black,
align=left,
font=\normalsize,
inner xsep=5pt,
inner ysep=4pt,
]
\tikzstyle{leaf5}=[
my-box,
minimum height=1.5em,
fill=darkblue!15,
text=black,
align=left,
font=\normalsize,
inner xsep=5pt,
inner ysep=4pt,
]
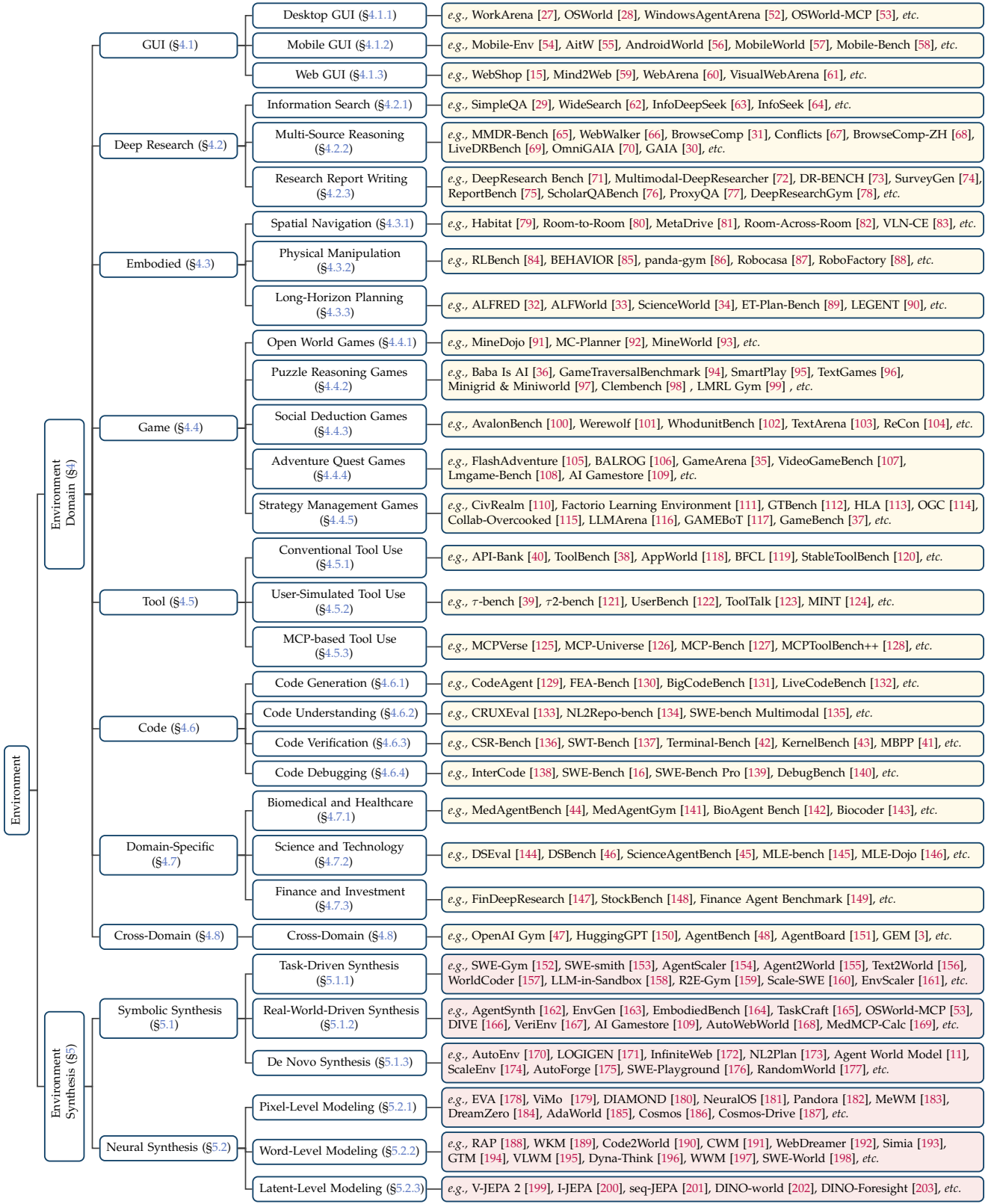
\begin{figure*}[!t]
        \vspace{-2mm}
        \centering
        \resizebox{0.96\textwidth}{!}{
                \begin{forest}
                        forked edges,
                        for tree={
                        grow=east,
                        reversed=true,
                        anchor=base west,
                        parent anchor=east,
                        child anchor=west,
                        base=left,
                        font=\normalsize,
                        rectangle,
                        draw=hidden-black,
                        rounded corners,
                        align=left,
                        minimum width=4em,
                        edge+={darkgray, line width=1pt},
                        edge path={
                          \noexpand\path[\forestoption{edge}]
                          (!u.parent anchor) -- +(6pt,0) |- (.child anchor)
                          \forestoption{edge label};
                        },
                        s sep=3pt,
                        inner xsep=2pt,
                        inner ysep=4pt,
                        line width=1.1pt,
                        ver/.style={rotate=90, child anchor=north, parent anchor=south, anchor=center},
                        },
                        where level=1{text width=12em,font=\normalsize,align=center,}{},
                        where level=2{text width=11em,font=\normalsize,align=center,}{},
                        where level=3{text width=14em,font=\normalsize,align=center,}{},
                        where level=4{text width=12em,font=\normalsize,align=left,}{},
                        where level=5{text width=50em,font=\normalsize,align=left}{},
                        [\ \ Environment\ \ \ , ver
                        % 3
                    [\ \ \ \ \ \ \ \ \   Environment \\ \ \ \ \ \ \ \ \ \  \ Domain~(\S\ref{sec:env_domain}), ver
                        % 4.1
                        [\ \ \ \ \ \ \ \ \ \  GUI~(\S\ref{GUI}) 
                            %4.1.1
                            [\ \ \ \ \ \ \  Desktop GUI~(\S\ref{Desktop GUI}) \ \ \ \ \ \
                                [\eg ~WorkArena~\cite{WorkArena}{,} 
                                OSWorld~\cite{OSWorld}{,}  WindowsAgentArena~\cite{WindowsAgentArena}{,}
                                OSWorld-MCP~\cite{OSWorld-MCP}{,}
                                \textit{etc.}, leaf3, text width=44em] 
                            ]
                            [\ \ \ \ \ \ \ \  Mobile GUI~(\S\ref{Mobile GUI}) \ \ \ \ \ \
                                [\eg ~Mobile-Env~\cite{Mobile-Env}{,}
                                AitW~\cite{AitW}{,}
                                AndroidWorld~\cite{AndroidWorld}{,}
                                MobileWorld~\cite{MobileWorld}{,}
                                Mobile-Bench~\cite{Mobile-Bench}{,}
                                \textit{etc.}, leaf3, text width=44em] 
                            ]
                            [\ \ \ \ \ \ \ \ \ \ Web GUI~(\S\ref{Web GUI}) \ \ \ \ \ \
                                [\eg ~WebShop~\cite{WebShop}{,} Mind2Web~\cite{Mind2Web}{,}
                                WebArena~\cite{WebArena}{,}
                                VisualWebArena~\cite{VisualWebArena}{,}
                                \textit{etc.}, leaf3, text width=44em] 
                            ]
                        ]
                        [\ \ \  Deep Research~(\S\ref{sec:Deep Research}) \ \ \ \ \ \ \ \ \
                            %4.1.1
                            [\ \ \   Information Search~(\S\ref{Information Search}) \ \ \ \ \ \ \ \ \ \ \ \
                                [\eg ~SimpleQA~\cite{SimpleQA}{,} WideSearch~\cite{WideSearch}{,} InfoDeepSeek~\cite{InfoDeepSeek}{,} InfoSeek~\cite{InfoSeek}{,}          
                                \textit{etc.}, leaf3, text width=44em] 
                            ]
                            [\ \ \ \ \ Multi-Source Reasoning \\ \ ~(\S\ref{Multi-Source Reasoning})
                                [\eg ~MMDR-Bench~\cite{MMDeepResearch_Bench}{,} WebWalker~\cite{WebWalker}{,} BrowseComp~\cite{BrowseComp}{,} Conflicts~\cite{Conflicts}{,} 
                                BrowseComp-ZH~\cite{BrowseComp-ZH}{,} \\ LiveDRBench~\cite{LiveDRBench}{,} OmniGAIA~\cite{OmniGAIA}{,} GAIA~\cite{GAIA}{,} 
                                \textit{etc.}, leaf3, text width=44em] 
                            ]
                            [\ \ \ \ \ Research Report Writing \\ \ ~(\S\ref{Research Report Writing})
                                [\eg ~DeepResearch Bench~\cite{DeepResearch-Bench}{,} Multimodal-DeepResearcher~\cite{Multimodal_DeepResearcher}{,} DR-BENCH~\cite{DR_BENCH}{,} SurveyGen~\cite{SurveyGen}{,} \\ ReportBench~\cite{ReportBench}{,} ScholarQABench~\cite{ScholarQABench}{,} ProxyQA~\cite{ProxyQA}{,}  DeepResearchGym~\cite{DeepResearchGym}{,}
                                \textit{etc.}, leaf3, text width=44em] 
                            ]
                        ]    
                        [\ \ \ \ \ \ Embodied~(\S\ref{Embodied})
                            %4.1.1
                            [\ \ \ \   Spatial Navigation~(\S\ref{Spatial Navigation})
                                [\eg ~Habitat~\cite{Habitat}{,} 
                                Room-to-Room~\cite{Room-to-Room}{,}
                                MetaDrive~\cite{MetaDrive}{,}
                                Room-Across-Room~\cite{Room-Across-Room}{,}
                                VLN-CE~\cite{VLN-CE}{,}
                                \textit{etc.}, leaf3, text width=44em] 
                            ]
                            [\ \ \ \ \ \ Physical Manipulation \\ \ \ \ ~(\S\ref{Physical Manipulation})
                                [\eg ~RLBench~\cite{RLBench}{,} BEHAVIOR~\cite{BEHAVIOR}{,}
                                panda-gym~\cite{panda-gym}{,}
                                Robocasa~\cite{Robocasa}{,}
                                RoboFactory~\cite{RoboFactory}{,}
                                \textit{etc.}, leaf3, text width=44em] 
                            ]
                            [\ \ \ \ \  Long-Horizon Planning \\ \ \ ~(\S\ref{Long-Horizon Planning})
                                [\eg ~ALFRED~\cite{ALFRED}{,}  
                                ALFWorld~\cite{ALFWorld}{,}
                                ScienceWorld~\cite{ScienceWorld}{,}
                                ET-Plan-Bench~\cite{ET-Plan-Bench}{,}
                                LEGENT~\cite{LEGENT}{,}
                                % EmbodiedBench~\cite{EmbodiedBench}{,}
                                % SimuHome~\cite{SimuHome}{,}
                                \textit{etc.}, leaf3, text width=44em] 
                            ]
                        ]  
                        [\ \ \ \ \ \ \ \ \  Game~(\S\ref{Game})
                            %4.1.1
                            [\ \ \    Open World Games~(\S\ref{Open World Games})
                                [\eg ~MineDojo~\cite{MineDojo}{,}  MC-Planner~\cite{MC-Planner}{,} MineWorld~\cite{MINEWORLD}{,}
                                \textit{etc.}, leaf3, text width=44em] 
                            ]
                            [\ \ \ \   Puzzle Reasoning Games \\ ~(\S\ref{Puzzle Reasoning Games})
                                [\eg ~Baba Is AI~\cite{BabaIsAI}{,} GameTraversalBenchmark~\cite{GameTraversalBenchmark}{,} SmartPlay~\cite{SMARTPLAY}{,} TextGames~\cite{TEXTGAMES}{,} \\ Minigrid \& Miniworld~\cite{Minigrid_Miniworld}{,} Clembench~\cite{clembench} {,} LMRL Gym~\cite{LMRLGym} {,}      
                                \textit{etc.}, leaf3, text width=44em] 
                            ]
                            [\ \ \ \ \  Social Deduction Games \\ ~(\S\ref{Social Deduction Games})
                                [\eg ~AvalonBench~\cite{AvalonBench}{,} Werewolf~\cite{Werewolf}{,} WhodunitBench~\cite{WhodunitBench}{,} TextArena~\cite{TextArena}{,}      ReCon~\cite{ReCon}{,}     
                                \textit{etc.}, leaf3, text width=44em] 
                            ]
                            [\ \ \ \   Adventure Quest Games \\ ~(\S\ref{Adventure Quest Games})
                                [\eg ~FlashAdventure~\cite{FlashAdventure}{,} BALROG~\cite{BALROG}{,} GameArena~\cite{GameArena}{,} VideoGameBench~\cite{VideoGameBench}{,} \\ Lmgame-Bench~\cite{LMGAME-BENCH}{,} AI Gamestore~\cite{AI_GAMESTORE}{,}          
                                \textit{etc.}, leaf3, text width=44em] 
                            ]
                            [\    Strategy Management Games \\ ~(\S\ref{Strategy Management Games})
                                [\eg ~CivRealm~\cite{CivRealm}{,} Factorio Learning Environment~\cite{FactorioLearningEnvironment}{,} GTBench~\cite{GTBENCH}{,} HLA~\cite{HLA}{,} OGC~\cite{OGC}{,} \\ Collab-Overcooked~\cite{Collab_Overcooked}{,} LLMArena~\cite{LLMARENA}{,} GAMEBoT~\cite{GAMEBoT}{,} GameBench~\cite{GAMEBENCH}{,}           
                                \textit{etc.}, leaf3, text width=44em] 
                            ]
                        ]
                        [\ \ \ \ \ \ \ \ \ \ Tool~(\S\ref{Tool})
                            %4.1.1
                            [\ \ \ \ \ \ Conventional Tool Use \\ \ ~ (\S\ref{Conventional Tool Use})
                                [\eg ~API-Bank~\cite{API-Bank}{,} 
                                ToolBench~\cite{ToolBench}{,}
                                AppWorld~\cite{AppWorld2024}{,}
                                BFCL~\cite{BFCL}{,}
                                StableToolBench~\cite{StableToolBench}{,}
                                \textit{etc.}, leaf3, text width=44em] 
                            ]
                            [\ \ \ \  User-Simulated Tool Use \\ \ ~(\S\ref{User-Simulated Tool Use})
                                [\eg ~$\tau$-bench~\cite{tau-bench}{,} 
                                $\tau$2-bench~\cite{tau2-bench}{,}
                                UserBench~\cite{UserBench}{,}
                                ToolTalk~\cite{ToolTalk}{,}
                                MINT~\cite{MINT}{,}
                                \textit{etc.}, leaf3, text width=44em] 
                            ]
                            [\ \ \ \ \ \ \  MCP-based Tool Use \\ \ \ \ ~(\S\ref{MCP-based Tool Use})
                                [\eg ~MCPVerse~\cite{MCPVerse}{,} 
                                MCP-Universe~\cite{MCP-Universe}{,}
                                MCP-Bench~\cite{MCP-Bench}{,}
                                MCPToolBench++~\cite{MCPToolBench++}{,}
                                \textit{etc.}, leaf3, text width=44em] 
                            ]
                        ]  
                        [\ \ \ \ \ \ \ \ \ Code~(\S\ref{Code})
                            %4.1.1
                            [\ \ \ \ \   Code Generation~(\S\ref{Code Generation})
                                [\eg ~CodeAgent~\cite{CodeAgent}{,} FEA-Bench~\cite{FEA_Bench}{,}     BigCodeBench~\cite{BigCodeBench}{,} LiveCodeBench~\cite{LiveCodeBench}{,}  
                                \textit{etc.}, leaf3, text width=44em] 
                            ]
                            [\ \   Code Understanding~(\S\ref{Code Understanding})
                                [\eg ~CRUXEval~\cite{CRUXEval}{,} NL2Repo-bench~\cite{NL2Repo_Bench}{,}
                                 SWE-bench Multimodal~\cite{SWE-bench_Multimodal}{,}
                                \textit{etc.}, leaf3, text width=44em] 
                            ]
                            [\ \ \ \ \  Code Verification~(\S\ref{Code Verification})
                                [\eg ~CSR-Bench~\cite{CSR-Bench}{,} SWT-Bench~\cite{SWT_Bench}{,}        Terminal-Bench~\cite{Terminal-Bench}{,}  KernelBench~\cite{KernelBench}{,} MBPP~\cite{MBPP}{,}
                                \textit{etc.}, leaf3, text width=44em] 
                            ]
                            [\ \ \ \ \  Code Debugging~(\S\ref{Code Debugging})
                                [\eg ~InterCode~\cite{InterCode}{,} SWE-Bench~\cite{SWE-Bench}{,}       SWE-Bench Pro~\cite{SWE-Bench_Pro}{,} DebugBench~\cite{DebugBench}{,}   
                                \textit{etc.}, leaf3, text width=44em] 
                            ]
                        ] 
                        [\ \ \ \ \ \ Domain-Specific \\ \ \ \ ~(\S\ref{Domain-Specific})
                            %4.1.1
                            [\ \ \    Biomedical and Healthcare \\ ~(\S\ref{Biomedical and Healthcare})
                                [\eg ~MedAgentBench~\cite{MedAgentBench}{,}
                                MedAgentGym~\cite{MedAgentGym}{,}         
                                BioAgent Bench~\cite{BioAgent_Bench}{,}
                                Biocoder~\cite{Biocoder}{,}
                                \textit{etc.}, leaf3, text width=44em] 
                            ]
                            [\ \ \ \ \  Science and Technology \\ \ ~(\S\ref{Science and Technology})
                                [\eg ~DSEval~\cite{DSEval}{,} 
                                DSBench~\cite{DSBench}{,}
                                ScienceAgentBench~\cite{ScienceAgentBench}{,}
                                MLE-bench~\cite{MLE-bench}{,}
                                MLE-Dojo~\cite{MLE-Dojo}{,}
                                \textit{etc.}, leaf3, text width=44em] 
                            ]
                            [\ \ \ \ \  Finance and Investment \\ \ ~(\S\ref{Finance})
                                [\eg ~FinDeepResearch~\cite{FinDeepResearch}{,} StockBench~\cite{StockBench}{,}
                                % CRMArena-Pro~\cite{CRMArena-Pro}{,}
                                Finance Agent Benchmark~\cite{Finance_Agent_Benchmark}{,}
                                \textit{etc.}, leaf3, text width=44em]
                            ]
                        ] 
                        [\ \ \    Cross-Domain~(\S\ref{Cross-Domain})
                            %4.1.1
                            [\ \ \ \ \ \ \ \   Cross-Domain~(\S\ref{Cross-Domain})
                                [\eg ~OpenAI Gym~\cite{Openai_gym}{,}
                                HuggingGPT~\cite{HuggingGPT}{,}
                                AgentBench~\cite{AgentBench}{,}
                                AgentBoard~\cite{AgentBoard}{,}
                                % AgentGym~\cite{AgentGym}{,}
                                GEM~\cite{GEM}{,}
                                \textit{etc.}, leaf3, text width=44em] 
                            ]
                        ] 
                    ]
                    % 4
                    [\ \ \ \ \ \ \ \ \   Environment \\ \ \ \ \ \ \ \ \ \ \ Synthesis~(\S\ref{sec:env_syn}),ver
                        % 3.1
                        [ \ \ \ \  Symbolic Synthesis \\ \ \ ~(\S\ref{sec:Symbolic Synthesis})\ \ \
                            %3.1.1
                            [\ \ \ \ \ \ Task-Driven Synthesis \\ \ \ \ ~(\S\ref{sec:task driven synthesis})
                                [\eg ~SWE-Gym~\cite{SWE-Gym}{,} 
                                SWE-smith~\cite{SWE-smith}{,} 
                                AgentScaler~\cite{AgentScaler}{,} 
                                Agent2World~\cite{Agent2World}{,} 
                                Text2World~\cite{Text2World}{,} \\ 
                                WorldCoder~\cite{WorldCoder}{,} 
                                LLM-in-Sandbox~\cite{LLM-in-Sandbox}{,} 
                                R2E-Gym~\cite{R2E-Gym}{,}
                                Scale-SWE~\cite{Scale-SWE}{,}
                                EnvScaler~\cite{EnvScaler}{,}
                                \textit{etc.}, leaf, text width=44em]  
                            ]
                            %3.1.2
                            [\ \ Real-World-Driven Synthesis \\   ~(\S\ref{sec:real world driven synthesis})
                                [\eg ~AgentSynth~\cite{AgentSynth}{,} 
                                EnvGen~\cite{EnvGen}{,} 
                                EmbodiedBench~\cite{EmbodiedBench}{,}
                                TaskCraft~\cite{TaskCraft}{,} 
                                OSWorld-MCP~\cite{OSWorld-MCP}{,} \\
                                DIVE~\cite{DIVE}{,} 
                                VeriEnv~\cite{VeriEnv}{,} 
                                AI Gamestore~\cite{AI_GAMESTORE}{,} 
                                AutoWebWorld~\cite{AutoWebWorld}{,} 
                                MedMCP-Calc~\cite{MedMCP-Calc}{,} 
                                \textit{etc.}, leaf, text width=44em]  
                            ]
                            [\ \ \  De Novo Synthesis~(\S\ref{sec:synthesis from scratch})
                                [\eg ~AutoEnv~\cite{AutoEnv}{,}
                                LOGIGEN~\cite{LOGIGEN}{,} 
                                InfiniteWeb~\cite{InfiniteWeb}{,} 
                                NL2Plan~\cite{NL2Plan}{,} 
                                Agent World Model~\cite{Agent_world_model}{,} \\
                                ScaleEnv~\cite{ScaleEnv}{,} 
                                AutoForge~\cite{AutoForge}{,} 
                                SWE-Playground~\cite{SWE-Playground}{,} 
                                RandomWorld~\cite{RandomWorld}{,} 
                                \textit{etc.}, leaf, text width=44em]  
                            ]
                        ]
                        [ \ Neural Synthesis~(\S\ref{sec:Neural Synthesis})\ \ \
                            %3.1.1
                            [ \ Pixel-Level Modeling~(\S\ref{sec:Pixel-level Modeling})
                                [\eg ~EVA~\cite{EVA}{,} 
                                ViMo ~\cite{ViMo}{,} 
                                DIAMOND~\cite{DIAMOND}{,}
                                NeuralOS~\cite{NeuralOS}{,}
                                Pandora~\cite{Pandora}{,}
                                MeWM~\cite{MeWM}{,} \\
                                DreamZero~\cite{DreamZero}{,}
                                AdaWorld~\cite{AdaWorld}{,}
                                Cosmos~\cite{Cosmos}{,}
                                Cosmos-Drive~\cite{Cosmos-Drive}{,}
                                \textit{etc.}, leaf, text width=44em]  
                            ]
                            %3.1.2
                            [\  Word-Level Modeling~(\S\ref{sec:Word-level Modeling})
                                [\eg ~RAP~\cite{RAP}{,}
                                WKM~\cite{WKM}{,} 
                                Code2World~\cite{Code2World}{,} 
                                CWM~\cite{CWM}{,}
                                WebDreamer~\cite{WebDreamer}{,}
                                Simia~\cite{Simia}{,} \\
                                GTM~\cite{GTM}{,}
                                VLWM~\cite{VLWM}{,}
                                Dyna-Think~\cite{Dyna-Think}{,}
                                WWM~\cite{WWM}{,}
                                SWE-World~\cite{SWE-World}{,}
                                \textit{etc.}, leaf, text width=44em]  
                            ]
                            [\ Latent-Level Modeling~(\S\ref{sec:Latent-level Modeling})
                                [\eg ~V-JEPA 2~\cite{V-JEPA_2}{,} 
                                I-JEPA~\cite{I-JEPA}{,} 
                                seq-JEPA~\cite{seq-JEPA}{,}
                                DINO-world~\cite{DINO-world}{,}
                                DINO-Foresight~\cite{DINO-Foresight}{,}
                                \textit{etc.}, leaf, text width=44em]  
                            ]
                        ]
                    ]
                ]
                \end{forest}
        }
        \caption{Taxonomy of the main content of this survey.}\label{fig:taxonomy}
\end{figure*}
\tikzstyle{my-box}=[
rectangle,
draw=hidden-black,
rounded corners,
text opacity=1,
minimum height=1.5em,
minimum width=5em,
inner sep=2pt,
align=left,
fill opacity=.5,
]
\tikzstyle{leaf6}=[
my-box,
minimum height=1.5em,
fill=purple!30,
text=black,
align=left,
font=\normalsize,
inner xsep=5pt,
inner ysep=4pt,
]
\tikzstyle{leaf4}=[
my-box,
minimum height=1.5em,
fill=hidden-blue!57,
text=black,
align=left,
font=\normalsize,
inner xsep=5pt,
inner ysep=4pt,
]
\tikzstyle{leaf2}=[
my-box,
minimum height=1.5em,
fill=hidden-green!20,
text=black,
align=left,
font=\normalsize,
inner xsep=5pt,
inner ysep=4pt,
]
\tikzstyle{leaf}=[
my-box,
minimum height=1.5em,
fill=hidden-red!20,
text=black,
align=left,
font=\normalsize,
inner xsep=5pt,
inner ysep=4pt,
]
\tikzstyle{leaf5}=[
my-box,
minimum height=1.5em,
fill=darkblue!15,
text=black,
align=left,
font=\normalsize,
inner xsep=5pt,
inner ysep=4pt,
]
\begin{figure*}[!t]
        \vspace{-2mm}
        \centering
        \resizebox{0.96\textwidth}{!}{
                \begin{forest}
                        forked edges,
                        for tree={
                        grow=east,
                        reversed=true,
                        anchor=base west,
                        parent anchor=east,
                        child anchor=west,
                        base=left,
                        font=\normalsize,
                        rectangle,
                        draw=hidden-black,
                        rounded corners,
                        align=left,
                        minimum width=4em,
                        edge+={darkgray, line width=1pt},
                        edge path={
                          \noexpand\path[\forestoption{edge}]
                          (!u.parent anchor) -- +(6pt,0) |- (.child anchor)
                          \forestoption{edge label};
                        },
                        s sep=3pt,
                        inner xsep=2pt,
                        inner ysep=4pt,
                        line width=1.1pt,
                        ver/.style={rotate=90, child anchor=north, parent anchor=south, anchor=center},
                        },
                        where level=1{text width=12em,font=\normalsize,align=center,}{},
                        where level=2{text width=11em,font=\normalsize,align=center,}{},
                        where level=3{text width=14em,font=\normalsize,align=center,}{},
                        where level=4{text width=12em,font=\normalsize,align=left,}{},
                        where level=5{text width=50em,font=\normalsize,align=left}{},
                        [\ \ Evolution\ \ \ , ver
                        % 3
                    [\ \ \ \ \ \ \ \ \   Agent \\ \ \ \ \ \ \ \ \ \ Evolution~(\S\ref{sec:agent_evo}), ver
                        % 4.1
                        [\ \ \ \ Memory-Centric \\ \ \ Experience Evolution \\ \ ~(\S\ref{sec:agent_evo_experience_utilization})
                            %4.1.1
                            [\ \ \ \ \ \ \ \ \   Instance Trajectory \\ \ \ \ \ \ \ \ \   Experience ~(\S\ref{sec:instance_trajectory_experience})
                                [\eg ~ OpenAgent~\cite{OpenAgent}{,} CoPS~\cite{CoPS}{,} ELLMER~\cite{ELLMER}{,}           Synapse~\cite{Synapse}{,} WorldMM~\cite{WorldMM}{,} 
                                \textit{etc.}, leaf2, text width=44em] 
                            ]
                            [\ \ \ \ \ \ \ \ \   Abstract Scripts \\ \ \ \ \ \ \ \ \  Experience~(\S\ref{sec:abstract_scripts_experience})
                                [\eg ~Reasoning Bank~\cite{Reasoning-Bank}{,} Agent-KB~\cite{AGENTKB_2}{,} DeepAgent~\cite{DeepAgent}{,} FLEX~\cite{FLEX}{,} O-Mem~\cite{O-Mem}{,} \\ Reme~\cite{Reme}{,} AWM~\cite{AWM}{,} PhysMem~\cite{PhysMem}{,} Agent-Pro~\cite{Agent-Pro}{,} BIFROST~\cite{BIFROST}{,}           
                                \textit{etc.}, leaf2, text width=44em] 
                            ]
                            [\ \ \ \ \ \ \ \ \   Structured Skill \\ \ \ \ \ \ \ \ \  Experience~(\S\ref{sec:structure_skill_experience})
                                [\eg ~SAGE~\cite{SAGE}{,} ASI~\cite{ASI}{,} SkillWeaver~\cite{SkillWeaver}{,} SkillRL~\cite{SkillRL_2}{,} SkillOrchestra~\cite{SkillOrchestra}{,}            
                                \textit{etc.}, leaf2, text width=44em] 
                            ]
                        ]
                        [ \ \ Orchestration-Centric \\ \ \ Workflow Evolution \\ \ ~(\S\ref{sec:agent_evo_agentic_workflow_design})
                            %4.1.1
                            [\ \ \ \ \  Fixed Workflow 
                            (\S\ref{sec:fixed_workflow})
                                [\eg~MetaGPT~\cite{METAGPT}{,} Agentless~\cite{Agentless}{,} 
                                CAMEL~\cite{CAMEL}{,} 
                                WebVoyager~\cite{WebVoyager}{,} 
                                MedAgent-Zero~\cite{MedAgent_Zero}{,} \\
                                OpenHands~\cite{OpenHands}{,}
                                SWE-agent~\cite{SWE_agent}{,}
                                \textit{etc.}, leaf2, text width=44em] 
                            ]
                            [\ \ Automated Workflow
                         (\S\ref{sec:automated_workflow})
                                [\eg~MaAS~\cite{MaAS}{,} HuggingGPT~\cite{HuggingGPT}{,}  
                                AutoFlow~\cite{AutoFlow}{,}
                                Workforce~\cite{OWL}{,}
                                SWE-search~\cite{SWE_Search}{,} \\
                                Plan-and-act~\cite{PLAN_AND_ACT}{,}
                            HyperAgent~\cite{HYPERAGENT}{,}
                                Webpilot~\cite{Webpilot}{,}
                        RepairAgent~\cite{RepairAgent}{,} 
                                \textit{etc.}, leaf2, text width=44em] 
                            ]
                            [\ \ \ Evolving Workflow
                            (\S\ref{sec:evolving_workflow})
                                [\eg~AFlow~\cite{AFlow}{,} 
                                ADAS~\cite{ADAS}{,}
                                ReasonRAG~\cite{ReasonRAG}{,} 
                                TextGrad~\cite{TextGrad}{,}
                                GPTSwarm~\cite{GPTSwarm}{,} 
                                DSPy~\cite{DSPy}{,} \\
                                Chain-of-Agents~\cite{Chain-of-Agents}{,}
                                Criticize-Reflect~\cite{Criticize_Reflect}{,}
                                \textit{etc.}, leaf2, text width=44em] 
                            ]
                        ]    
                        [\ \ \ \ \ Trajectory-Centric \\ \ \ \ \ Offline Evolution \\ \ \ \ ~(\S\ref{sec:agent_evo_synthetic_data_generation})
                            %4.1.1
                            [\ \ \ \ \ \ Task Synthesis~(\S\ref{sec:task_synthesis})\ \ \
                                [\eg ~BAGEL~\cite{BAGEL}{,} OS-Genesis~\cite{OS-Genesis}{,} Insta~\cite{Insta}{,} APIGen-MT~\cite{APIGen-MT}{,} WebShaper~\cite{WebShaper}{,}\\ WebWatcher~\cite{WebWatcher}{,}
                                WebExplorer~\cite{WebExplorer}{,} AutoPlay~\cite{AutoPlay}{,} CRMWeaver~\cite{CRMWeaver}{,} WebLeaper~\cite{WebLeaper}{,}              
                                \textit{etc.}, leaf2, text width=44em] 
                            ]
                            [\ \ Trajectory Synthesis~(\S\ref{sec:Trajectory_synthesis})\ \ \
                                [\eg ~ToolAlpaca~\cite{ToolAlpaca}{,} Lingma SWE-GPT~\cite{Lingma_SWE-GPT}{,} Aguvis~\cite{Aguvis}{,} FlowReasoner~\cite{FlowReasoner}{,}\\ WebSynthesis~\cite{WebSynthesis}{,} AgentFold~\cite{AgentFold}{,} ToolACE-MCP~\cite{ToolACE-MCP}{,} ProAct~\cite{ProAct}{,}  
                                \textit{etc.}, leaf2, text width=44em] 
                            ]
                            [\  Trajectory Refinement~(\S\ref{sec:trajectory_refinement})\ \ \
                                [\eg ~Toolformer~\cite{Toolformer}{,} ETO~\cite{ETO}{,} Self-Improvement~\cite{Self-Improvement}{,} GUI-Reflection~\cite{GUI-Reflection}{,} TiG~\cite{TiG}{,}\\ AgentFrontier~\cite{AgentFrontier}{,} WebSTAR~\cite{WebSTAR}{,} 
                                SynthAgent~\cite{SynthAgent}{,} TopoCurate~\cite{TopoCurate}{,}            
                                \textit{etc.}, leaf2, text width=44em] 
                            ]
                        ]  
                        [\ \ \ Exploration-Centric \\ \ \ \ Online Evolution \\ \ \ ~(\S\ref{sec:agent_evo_reinforcement_learning_optimization})
                            %4.1.1
                            [\ \ Reasoning Structure~(\S\ref{sec:reasoning_structure})
                                [\eg ~DeepRetrieval~\cite{Deepretrieval}{,} Search-R1~\cite{Search-r1}{,} AutoRefine~\cite{AutoRefine}{,} SEEA-R1~\cite{SEEA-R1}{,} M3-Agent~\cite{M3-Agent}{,} \\ Video-Thinker~\cite{Video-Thinker}{,} ReSearch~\cite{ReSearch_2}{,}      
                                \textit{etc.}, leaf2, text width=44em] 
                            ]
                            [\ \ \ \ \  Reward Shaping~(\S\ref{sec:reward_shaping})
                                [\eg ~Agent-R1~\cite{Agent-R1}{,}  ToolRL~\cite{ToolRL}{,} Chain-of-Agents~\cite{Chain-of-Agents}{,} VRAG-RL~\cite{VRAG-RL}{,} GDPO~\cite{GDPO}{,}  \\ FlowSteer~\cite{FlowSteer}{,} Tool-N1~\cite{Tool-N1}{,} ToolOrchestra~\cite{ToolOrchestra}{,}     
                                \textit{etc.}, leaf2, text width=44em] 
                            ]
                            [\ \ \ \ \    Algorithmic \\ \ \  \ \ \ \ \  Optimization~(\S\ref{sec:algorithmic_optimization})
                                [\eg ~RAGEN~\cite{RAGEN}{,} ZeroSearch~\cite{ZeroSearch}{,} EvolveSearch~\cite{EvolveSearch}{,} GiGPO~\cite{GiGPO}{,} ARPO~\cite{ARPO}{,} \\ MobileGUI-RL~\cite{MobileGUI-RL}{,} SPEAR~\cite{SPEAR}{,} VAGEN~\cite{VAGEN}{,} SeeUPO~\cite{SeeUPO}{,}     
                                \textit{etc.}, leaf2, text width=44em] 
                            ]
                        ]
                    ]
                    % 4
                    [\ \ \ \ \ \ \ \ \ Environment \\ \ \ \ \ \ \ \ \ \ Evolution~(\S\ref{sec:env_evo}),ver
                        % 3.1
                        [\ \ \ \ \ Neural-Driven \\ \ \ \ \ \ \ Evolution~(\S\ref{sec:neural_evo})\ \ \
                            %3.1.1
                            [\ \ \ \ \ \ \ \ \ \ \ Self-Play~(\S\ref{sec:self-play})
                                [\eg ~Absolute Zero~\cite{Absolute_zero}{,}
                                Self-Challenging~\cite{Self-Challenging}{,}
                                Active Zero~\cite{Active_Zero}{,}
                                Vision-zero~\cite{Vision-zero}{,}
                                \textit{etc.}, leaf4, text width=44em]  
                            ]
                            %3.1.2
                            [\ \ \ \ \ \ \ World Model~(\S\ref{sec:world_model})
                                [\eg ~WebDreamer~\cite{WebDreamer}{,} 
                               UI-Simulator~\cite{UI-Simulator}{,} Code2World~\cite{Code2World}{,} WebWorld~\cite{WebWorld} {,}
                                \textit{etc.}, leaf4, text width=44em]  
                            ]
                        ]
                        [\ \ \ \ \ Difficulty-Driven \\ \ \ \ \ \ \ Evolution~(\S\ref{sec:diff_evo})\ \ \
                            %3.1.1
                            [\ \ \ \ \ \ \ \ Explicit Curriculum \\ \ \ \ \ \ \ \ \ \ Signals~(\S\ref{sec:Explicit Curriculum Signals})
                                [\eg ~POET~\cite{POET}{,} 
                                    AgentGen~\cite{AgentGen}{,} 
                                    Environment Tuning~\cite{Environment_Tuning}{,} 
                                    SEC~\cite{SEC}{,} 
                                    Reasoning Core~\cite{Reasoning_Core}{,} \\ 
                                    DreamGym~\cite{DreamGym} {,}
                                \textit{etc.}, leaf4, text width=44em]  
                            ]
                            %3.1.2
                            [\ \ \ \ \ \ \ \ Implicit Curriculum \\ \ \ \ \ \ \ \ \ Mechanisms~(\S\ref{sec:Implicit Curriculum Mechanisms})
                                [\eg ~PAIRED~\cite{PAIRED}{,} 
                                        DCD~\cite{DCD}{,} 
                                        ACCEL~\cite{ACCEL}{,} 
                                        MAESTRO~\cite{MAESTRO}{,} 
                                        ReMiDi~\cite{ReMiDi}{,} 
                                        EnvGen~\cite{EnvGen}{,} \\
                                        DataEnvGym~\cite{DataEnvGym}{,}
                                        Eurekaverse~\cite{Eurekaverse}{,} 
                                        RLVE~\cite{RLVE}{,} 
                                        CuES~\cite{CuES}{,} 
                                        GenEnv~\cite{GenEnv}{,} 
                                        SCALER~\cite{SCALER} {,}
                                \textit{etc.}, leaf4, text width=44em]  
                            ]
                        ]
                        [\ \ \ \ \ Scaling-Driven \\ \ \ \ \ \ \ Evolution~(\S\ref{sec:scaling_evo})\ \ \
                            %3.1.1
                            [\ \ \ \ \ \ \ \ \ \ \ Scenario-Level \\ \ \ \ \ \ \ \ \ \ \ \ Scaling~(\S\ref{sec:Scenario-Level Scaling})
                                [\eg ~FTRL~\cite{FTRL}{,} 
                                        AgentScaler~\cite{AgentScaler}{,} 
                                        AutoForge~\cite{AutoForge}{,} 
                                        EnvScaler~\cite{EnvScaler}{,} 
                                        InfiniteWeb~\cite{InfiniteWeb}{,} \\
                                        Agent World Model~\cite{Agent_world_model}{,} 
                                        AutoWebWorld~\cite{AutoWebWorld} {,}
                                \textit{etc.}, leaf4, text width=44em]  
                            ]
                            %3.1.2
                            [\ \ \ \ \ \ \ \ Environment-Level \\ \ \ \ \ \ \ \ \ Scaling~(\S\ref{sec:Environment-Level Scaling})
                                [\eg ~ARE~\cite{ARE}{,} 
                                AutoEnv~\cite{AutoEnv} {,} \textit{etc.}, leaf4, text width=44em]  
                            ]
                        ]
                    ]
                ]
                \end{forest}
                }
        \caption{Taxonomy of the main content of this survey.}\label{fig:taxonomy}
\end{figure*}
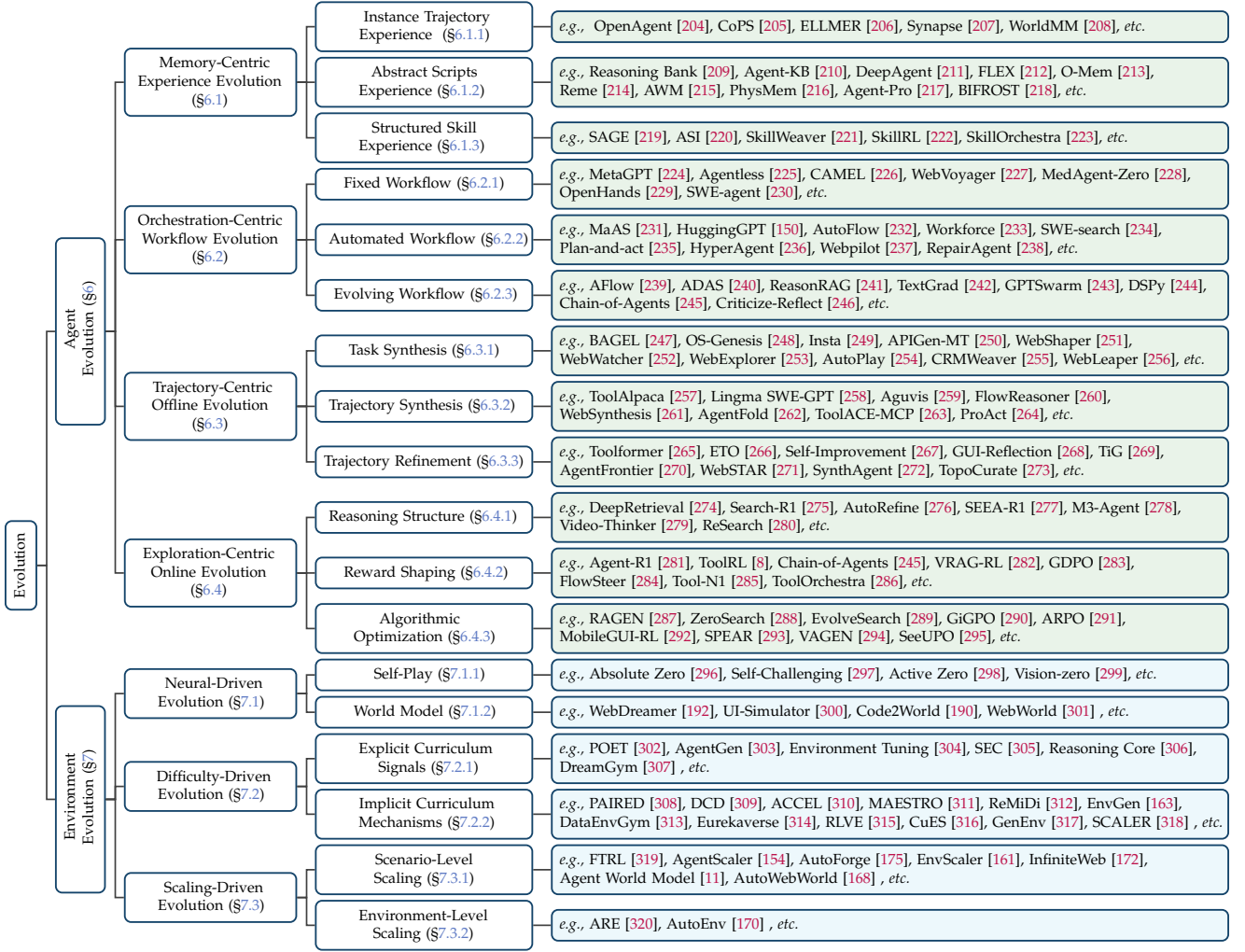

\section{Preliminaries} \label{sec:prelimi}
\subsection{Definition of Key Concepts}

\subsubsection{Environment}
We define the \textbf{Environment} $\mathcal{E}$ as a stochastic dynamical system with which the agent $\pi$ interacts. Formally, we represent $\mathcal{E}$ as a Partially Observable Markov Decision Process (POMDP), defined by the tuple $\langle \mathcal{S}, \mathcal{A}, \mathcal{P}, \mathcal{R}, \Omega, \mathcal{O}, \gamma \rangle$:

\begin{itemize}
    \item \textbf{State Space} $\mathcal{S}$: The set of all possible latent states of the environment.
    \item \textbf{Action Space} $\mathcal{A}$: The set of all possible actions available to the agent.
    \item \textbf{Transition Function} $\mathcal{P}$: Defined as $\mathcal{S} \times \mathcal{A} \rightarrow \Delta(\mathcal{S})$, this probabilistic kernel governs the environment's physical dynamics, where $\mathcal{P}(s_{t+1} | s_t, a_t)$ denotes the probability of transitioning to state $s_{t+1}$.
    \item \textbf{Reward Function} $\mathcal{R}$: Defined as $\mathcal{S} \times \mathcal{A} \rightarrow \mathbb{R}$, given the current state $s_t$ and action $a_t$, the mechanism providing scalar feedback $r_t$ to the agent.
    \item \textbf{Observation Space} $\Omega$ and \textbf{Observation Function} $\mathcal{O}$: Defined as $\mathcal{S} \times \mathcal{A} \rightarrow \Delta(\Omega)$, the interface denotes the agent's perception of the latent state, where $\mathcal{O}(o_t | s_t, a_t)$ denotes the probability of receiving observation $o_t$ given the current state and action.
    \item  \textbf{Discount Factor} $\gamma$: A constant $\in [0, 1)$ that weights the importance of future rewards, ensuring the mathematical convergence of the expected total return in infinite-horizon tasks.
\end{itemize}
In this work, we focus on \textbf{agentic environments}, defined as a class of dynamic, interactive systems designed for evaluating and training LLM agents. 
Unlike traditional reinforcement learning simulators which operate over predefined state-action spaces and fixed transition dynamics, agentic environments enable more open-ended, language-centric, and tool-augmented interactions.
Such environments define a set of permissible actions that the LLM can execute (e.g., generating natural language responses \cite{ALFWorld, MLE-bench, Openai_gym} or invoking external tools \cite{ToolACE, Toolformer, ToolRL}), and return corresponding observations (e.g., code execution results \cite{SWE-Bench,Terminal-Bench, LiveCodeBench} or HTML pages \cite{WebShop, WebArena, WebVoyager}) and rewards (e.g., answer correctness or format correctness \cite{LLMARENA, HLE, GAIA}), thereby enabling iterative interaction with the model.

\subsubsection{Agent}

The \textbf{Agent} is defined as an autonomous decision-making entity that interacts with $\mathcal{E}$. In a POMDP setting, the underlying state $s_t$ is not directly observable. Therefore, the agent conditions its next action on the \textbf{History} $h_t \in \mathcal{H}$, which represents the sequence of all previous observations, actions, and rewards:
\begin{equation}
    h_t = (o_0, a_0, r_0, \dots, a_{t-1}, r_{t-1}, o_t) \in \mathcal{H}
\end{equation}
The agent's behavior is governed by a \textbf{Policy} $\pi: \mathcal{H} \rightarrow \Delta(\mathcal{A})$, which maps the history to a probability distribution over actions. The probability of selecting action $a_t$ is given by:
\begin{equation}
    \pi(a_t | h_t) = P(A_t = a_t | H_t = h_t).
\end{equation}
The primary objective of the agent is to find an optimal policy $\pi^*$ that maximizes the \textbf{Expected Discounted Return} $J(\pi)$:
\begin{equation}
    J(\pi) = \mathbb{E}_{\pi, \mathcal{P}, \mathcal{O}} \left[ \sum_{k=0}^{\infty} \gamma^k r_{t+k} \right]
\end{equation}
where the expectation is taken over the trajectories induced by the interaction between the policy $\pi$ and the environment's transition and observation dynamics.

As this survey focuses on agents built upon LLMs, the policy $\pi$ corresponds to the internal parameters of the model, while $h_t$ represents the interaction history between the model and the environment. Unlike traditional reinforcement learning, where actions are typically discrete, the action $a_t$ in this setting primarily consists of natural language tokens generated by the LLM. As a result, trajectories are largely formed by chains of thought produced during the interaction between the model and the environment.

\subsubsection{Environment-Agent Alignment}
% \subsubsection{Agentic RL}
In the previous subsection, we note that the primary objective of environment-agent interaction is to optimize the policy $\pi$ for maximizing expected returns, which we term \textbf{Environment–Agent Alignment}. How to achieve this objective is also one of the main focuses of this survey.

Traditional approaches widely adopt teacher models to synthesize high-quality offline trajectories, which are then used for supervised fine-tuning (SFT) to enable imitation learning for the agent's policy $\pi$ \cite{ToolBench, Toolformer, Self-Improvement}. However, recent efforts have increasingly focused on a new paradigm known as Agentic Reinforcement Learning (RL).
This approach is particularly instrumental for developing reasoning models (e.g., DeepSeek-R1 \cite{DeepSeek-R1}), which refine multi-turn cognitive trajectories or tool-mediated interactions through RL \cite{GiGPO, ARPO, Search-r1}. Here, we introduce several representative Agentic RL algorithms as preliminary background:

\textbf{PPO.} \textbf{Proximal Policy Optimization (PPO)} \cite{PPO} has long been the de facto standard for LLM alignment and reinforcement learning. PPO typically employs an actor-critic framework, maintaining a value network $V_\phi$ to estimate the expected return, which serves as a baseline for variance reduction. The training objective involves a clipped surrogate loss to ensure stable updates by constraining the policy shift between iterations. For a given state-action pair, the PPO objective is defined as:
\begin{equation}
\mathcal{L}^{\text{PPO}}(\theta) = \mathbb{E}_{t} \left[ \min(r_t(\theta) \hat{A}_t, \text{clip}(r_t(\theta), 1-\epsilon, 1+\epsilon) \hat{A}_t) \right]
\end{equation}
where $r_t(\theta) = \frac{\pi_\theta(a_t|s_t)}{\pi_{\theta_{\text{old}}}(a_t|s_t)}$ is the probability ratio, with $\pi_{\theta_{\text{old}}}$ representing the reference policy from the previous iteration.
The advantage $\hat{A}_t$ is typically computed using Generalized Advantage Estimation (GAE) based on the rewards and the learned value function:
\begin{equation}
\hat{A}_t = \sum_{k=0}^{\infty} (\gamma \lambda)^k \delta_{t+k}, \quad \delta_t = r_t + \gamma V_\phi(s_{t+1}) - V_\phi(s_t)
\end{equation}
where $\gamma$ and $\lambda$ are hyperparameters for discounting and smoothing. While effective, the requirement of a dedicated critic network introduces significant memory and computational overhead, posing challenges for scaling in resource-constrained environments.

\textbf{GRPO.} While PPO serves as a foundational baseline, \textbf{Group Relative Policy Optimization (GRPO)} \cite{GRPO} has recently emerged as a highly efficient alternative for agentic RL works. In contrast to the standard actor-critic architecture, GRPO eliminates the need for an explicit value network by estimating the advantage from a sample-based baseline. Specifically, for a set of $G$ outputs $\{y_i\}_{i=1}^G$ sampled from the same prompt $q$, the GRPO objective is formulated as:
\begin{equation}
\mathcal{L}^{\text{GRPO}}(\theta) = \mathbb{E}_{q} \left[ \frac{1}{G} \sum_{i=1}^G \frac{1}{|y_i|} \sum_{t=1}^{|y_i|}\left( \mathcal{L}_{i,t}^{\text{CLIP}}(\theta) - \beta \mathbb{D}_{\text{KL}}(\pi_\theta \,\|\, \pi_{\text{ref}}) \right) \right]
\end{equation}
where the surrogate clipping objective is defined as: \begin{equation} \mathcal{L}_{i,t}^{\text{CLIP}}(\theta) = \min \left( r_{i,t}(\theta) \hat{A}_i, \, \text{clip}(r_{i,t}(\theta), 1-\epsilon, 1+\epsilon) \hat{A}_i \right) 
\end{equation}
with $r_{i,t}(\theta) = \frac{\pi_\theta(y_{i,t} \mid q, y_{i, < t})}{\pi_{\theta_{\text{old}}}(y_{i,t} \mid q, y_{i, < t})}$ denoting the importance sampling ratio.
The group-normalized advantage $\hat{A}_i$ is computed as:
\begin{equation}
\hat{A}_i = \frac{r_i - \mu}{\sigma + \delta}, \quad \mu = \frac{1}{G}\sum_{j=1}^G r_j, \quad \sigma = \sqrt{\frac{1}{G}\sum_{j=1}^G (r_j - \mu)^2}
\end{equation}
where $\delta > 0$ is a small constant for numerical stability. This relative reward mechanism effectively reduces variance without the computational overhead of a critic network, facilitating more stable optimization in high-dimensional discrete action spaces.

\textbf{DAPO.} To further enhance the stability and efficiency of group-relative methods, \textbf{Decoupled Clip and Dynamic Sampling Policy Optimization (DAPO)} \cite{DAPO} builds upon the GRPO framework by introducing several key architectural refinements. Addressing the entropy collapse and gradient vanishing issues in standard GRPO, DAPO decouples the clipping mechanism into separate upper and lower bounds, allowing for more aggressive updates when moving away from suboptimal states while maintaining strict constraints on positive shifts. Furthermore, DAPO shifts from sample-level to token-level policy gradient loss to ensure a more fine-grained credit assignment. The DAPO objective is refined as:
\begin{equation}\mathcal{L}^{\text{DAPO}}(\theta) = \mathbb{E}_{q} \left[ \frac{1}{\sum_{i=1}^{G}|y_i|} \sum_{i=1}^G \sum_{t=1}^{|y_i|} \mathcal{L}_{i,t}^{\text{D-CLIP}}(\theta) \right]
\end{equation}
where the decoupled surrogate clipping objective is defined as:\begin{equation}\mathcal{L}_{i,t}^{\text{D-CLIP}}(\theta) = \min \left( r_{i,t}(\theta) \hat{A}_i, \text{clip}(r_{i,t}(\theta), 1-\epsilon_{\text{low}}, 1+\epsilon_{\text{high}}) \hat{A}_i \right)
\end{equation}
The $\hat{A}_i$ and $r_{i,t}(\theta)$ follows the definition in GRPO. By integrating these refinements, DAPO effectively mitigates gradient vanishing and overshooting in complex reasoning tasks, ensuring more robust convergence in large-scale agentic training.

\begin{figure}[t]
    \centering
    \includegraphics[width=0.95\linewidth]{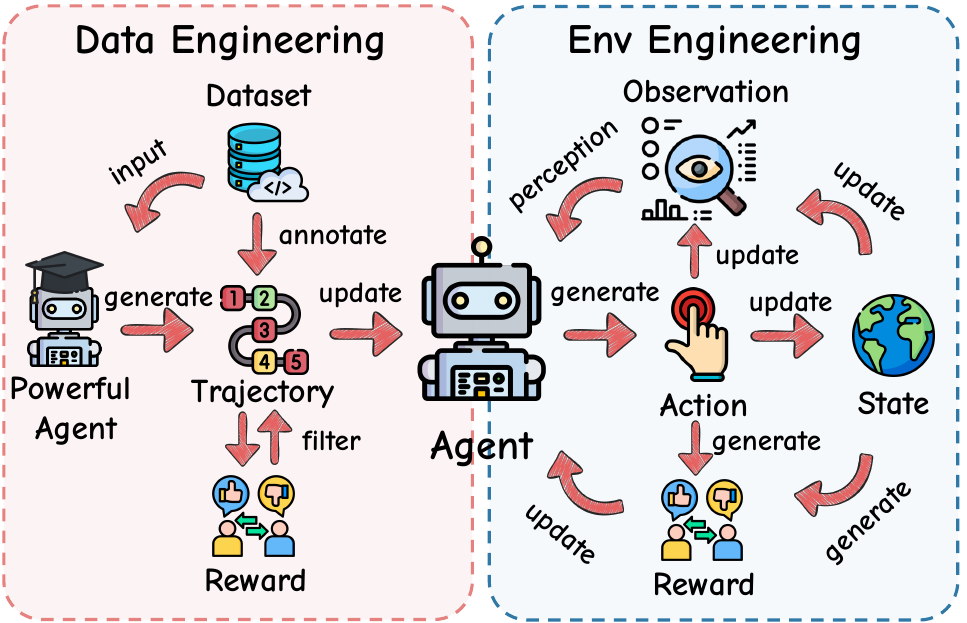}
    \caption{A comparison between data engineering and environment engineering.} 
    \label{fig:prelimi}
\end{figure}

\subsection{From Data Engineering to Environment Engineering}
In traditional tasks, we train and test LLMs based on datasets, and data-centric engineering has already matured significantly. So why, at this point in time, do we need environments to replace the role that data previously played? In this section, we aim to further introduce the characteristics of the environment and explain its value by comparing it with traditional data engineering.

\subsubsection{From Passive Learning to Collaborative Evolution}
In the paradigm of traditional data engineering, the training process is characterized by a unidirectional transfer of information from a static corpus to the model. The data acts as a fixed boundary of knowledge which forces the model to act as a passive recipient of pre-collected trajectories (see Fig. \ref{fig:prelimi}, left). This static nature creates a significant mismatch between the fixed difficulty of the dataset and the evolving proficiency of the learner. If the provided data is too advanced for the current state of the model, the optimization process often becomes unstable or fails to converge. Conversely, when the model encounters trajectories that are relatively simple, it quickly reaches a point of diminishing returns where additional training iterations no longer yield performance gains because the dataset lacks the necessary complexity to drive further improvement.

The introduction of an environment shifts this relationship toward a collaborative evolution, where the data distribution is dynamically adjusted based on the real-time performance of the agent. Instead of being restricted to a fixed set of samples, the environment acts as a generative engine that can modulate the complexity of tasks to match the growing capabilities of the model. For instance, some environments can regulate their complexity to match the agent’s progressively enhancing capability during the training phases \cite{AgentGen, DreamGym, SCALER}. This synchronization ensures that the learning signal remains within an optimal range of difficulty, allowing the model to continuously transcend its previous performance limits through a process of mutual progression between the learner and the simulated world.

\subsubsection{From Single-turn Q\&A to Multi-turn Interaction}
In traditional data engineering, most samples interact with the model in a single-turn question-and-answer format, such as in GSM8K\cite{GSM8k}, MMLU\cite{mmlu}, MMMU\cite{mmmu}, and others\cite{RAG-RewardBench}. However, this format inherently overlooks the procedural complexity required to navigate open-ended problems that cannot be solved in a single step. By isolating the question from a broader context of trial and error, traditional datasets encourage the model to find a direct mapping from the prompt to the answer. This often results in a model that is proficient at pattern matching within limited contexts but lacks the resilience to handle tasks that require long-term reasoning or error correction.

In contrast, an environment provides a rich interactive framework that supports multi-turn engagement and external tool integration. Within this ecosystem, the agent does not work in isolation but rather collaborates with various functional roles such as code executors \cite{SWE-Bench, Terminal-Bench, IDE-Bench}, knowledge retrievers \cite{WideSearch,BrowseComp,GAIA}, and other specialized agents \cite{RoboFactory, LLMARENA, unlocking_future, MCPVerse}. These multi-turn interactions allow the agent to decompose a complex objective into a series of manageable sub-tasks while receiving immediate feedback at each juncture. By interacting with an executor to verify a logic step or querying a search engine to fill a knowledge gap, the agent learns to manage a dynamic workflow \cite{AutoFlow, MUSE, AgentFlow}. This transition enables the model to move beyond the rote memorization of static solutions and instead master the general strategies of problem-solving and strategic planning that are essential for real-world applications.

\subsubsection{From Open-loop System to Closed-loop System}
Traditional data-driven training is essentially an open-loop system where the learning process is decoupled from the consequences of the model's outputs. In this paradigm, the model generates a response based on a static prompt and the loss is calculated against a fixed gold standard. This lack of a feedback loop means that the model cannot perceive how its predicted tokens influence the subsequent state of a task. Consequently, any minor error in the initial steps will accumulate without a mechanism for correction because the static data cannot provide a reactive signal to guide the model back to a successful path.

An environment transforms this process into a closed-loop control system by establishing a continuous coupling between the agent's actions and the resulting state changes. Every decision made by the model triggers a response from the environment which then serves as the new input for the next iteration. This reciprocal interaction forces the model to internalize the transition dynamics of the task and learn robust recovery strategies. By experiencing the objective outcomes of its own errors, the model moves beyond the simple memorization of surface patterns and develops a functional understanding of goal-oriented logic. This closed-loop mechanism is the fundamental prerequisite for achieving long-term stability and autonomous self-improvement in complex reasoning chains.

\subsection{Positioning Our Survey}
Consistent with the notion that \textit{man is shaped by his environment}, an agent and its environment are intrinsically coupled and cannot be considered in isolation. However, despite the increasing attention in recent surveys to the evolution of agents within their environments, existing studies largely adopt an agent-centric view \cite{survey_agent_adaptation,survey_agent_reason, survey_agent_search, survey_agent_system, survey_agent_rl, survey_agent_efficient, survey_agent_evo}. Some studies focus on agent reasoning methods \cite{survey_agent_rl, survey_agent_search, survey_agent_reason}, investigating how algorithms such as reinforcement learning can be applied to effectively improve agent performance across different environments.
Other studies take a system-level perspective \cite{survey_agent_adaptation,survey_agent_evo,survey_agent_system}, examining the evolution of agents by investigating how different components in agentic systems, such as context, tools, and algorithms, can be coordinated effectively. Our survey is organized around the lifecycle of environments, systematically covering the full pipeline of environment modeling, construction, evaluation and application, and providing perspectives complementary to those absent in prior work.

\section{Environment Attribute} \label{sec:env_attr}

As shown in Fig. \ref{fig:env_attribute}, the landscape of agent evaluation environments is characterized by a multidimensional taxonomy of attributes. These underlying attributes change how agents perceive, interact, and make decisions with the environments. To capture the fundamental differences among these environments, it is necessary to analyze them based on their basic mechanics and mathematical formulations. Therefore, we break down the basic attributes of these environments into several key pairs, including \textbf{Symbolic vs. Neural} (\cref{attribute:Symbolic vs. Neural}), \textbf{Open-Loop vs. Closed-Loop} (\cref{attribute:Open-Loop vs. Closed-Loop}), \textbf{Online vs. Offline} (\cref{attribute:Online vs. Offline}), \textbf{MDP vs. POMDP} (\cref{attribute:MDP vs. POMDP}), \textbf{Deterministic vs. Nondeterministic} (\cref{attribute:Deterministic vs. Nondeterministic}), \textbf{Discrete vs. Continuous} (\cref{attribute:Discrete vs. Continuous}), \textbf{Unimodal vs. Multimodal} (\cref{attribute:Unimodal vs. Multimodal}), and \textbf{Single-Agent vs. Multi-Agent} (\cref{attribute:Single-Agent vs. Multi-Agent}). This taxonomy establishes a systematic framework for analyzing these environments, highlighting the diverse complexities and fundamental attributes that shape agent-environment interactions.

\begin{figure*}[t]
    \centering
    \includegraphics[width=\textwidth]{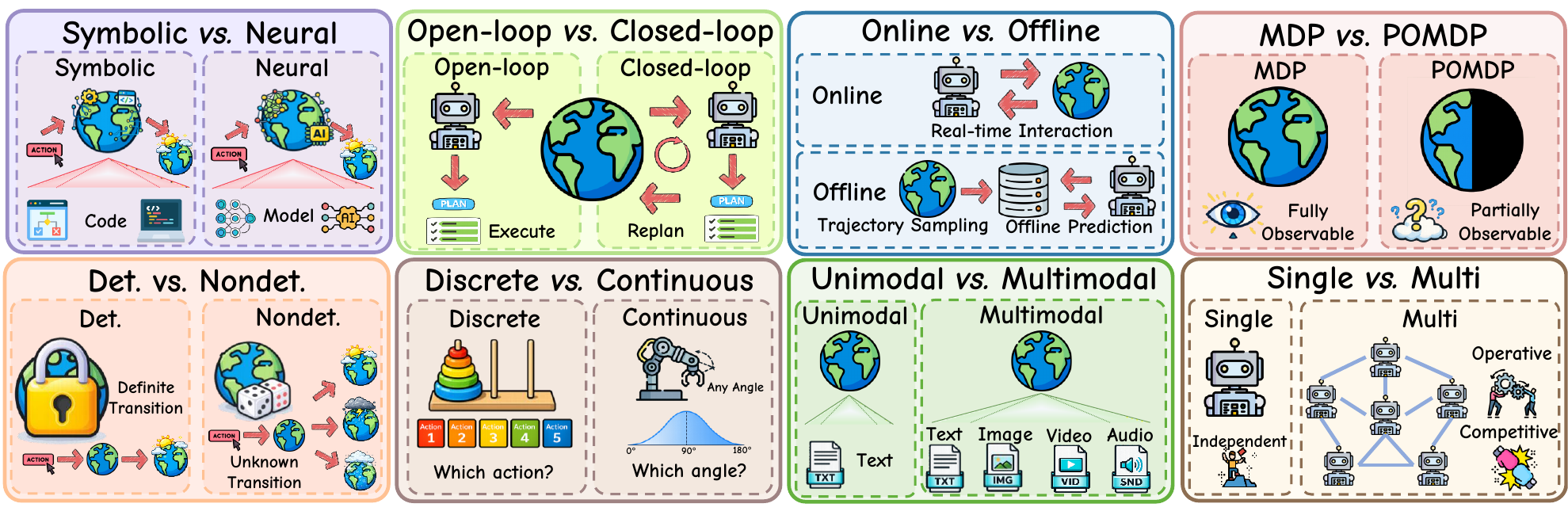}
    \caption{An overview of environment attributes.} 
    \label{fig:env_attribute}
\end{figure*}

\subsection{Symbolic vs. Neural}
\label{attribute:Symbolic vs. Neural}
The core distinction between these paradigms lies in the underlying implementation of the transition function $\mathcal{P}$ and how the environment dynamics are generated. This determines whether the system relies on programmed code to update the states or uses a neural model to predict future states.

\subsubsection{Symbolic Environment}
In a symbolic environment, the transition dynamics are governed by explicit programmed logic and predefined rules. The transition function $\mathcal{P}(s_{t+1} | s_t, a_t)$ is realized through executing source code, software scripts, or physics engines. As a foundational standard in this paradigm, the planning domain definition language (PDDL) governs transition dynamics through explicit symbolic action schemas, utilizing logical preconditions and deterministic effects to update environment states. While PDDL provides a strict logical foundation, modern environments predominantly realize these transitions through flexible programming code. Agent World Model~\cite{Agent_world_model} synthesizes large executable environments using code-driven backends and SQL databases to ensure reliable state transitions and consistency for agent training. EnvScaler~\cite{EnvScaler} programmatically synthesizes executable tool-interactive environments where transition dynamics and state updates are governed by explicit Python code and predefined rule-based logic. AutoEnv~\cite{AutoEnv} utilizes domain-specific languages and code abstractions to automate the synthesis of executable environments with explicit, programmed transition and reward logic.

\subsubsection{Neural Environment}
In a neural environment, the transition dynamics are approximated by a model parameterized by $\theta$. The transition function $\mathcal{P}_{\theta}(s_{t+1} | s_t, a_t)$ is realized through the forward propagation of these learned parameters, typically utilizing neural networks. This approach often serves as a surrogate simulator to generate state transitions where the true underlying physical rules are unknown or overly complex. WebDreamer~\cite{WebDreamer} uses LLMs as learned world models to simulate web state transitions, approximating environment dynamics through neural network parameters rather than explicit code. DreamGen~\cite{DreamGen} utilizes neural video world models to approximate environment transitions and generate synthetic trajectories for robot training through the forward propagation of learned parameters. WebEvolver~\cite{WebEvolver} employs concurrently optimized LLMs to function as virtual web servers, forecasting future webpage representations based on present conditions and executed actions to synthesize multi-step interaction trajectories.

\subsection{Open-Loop vs. Closed-Loop}
\label{attribute:Open-Loop vs. Closed-Loop}
The core distinction between the two paradigms lies in how the agent utilizes the observation sequence to formulate its actions over time. This determines whether the agent simply executes a fixed plan based on the starting conditions, or continuously adjusts its next steps based on new feedback from the environment.

\subsubsection{Open-Loop Environment}
In an open-loop system, the agent receives only a single initial observation $o_0$ and executes a predetermined sequence of actions. It does not incorporate subsequent feedback from the environment. The policy simplifies to depend on a sequence predetermined from the initial observation $\pi(a_t | o_0)$ in a non-adaptive manner. TaskLAMA~\cite{TaskLAMA} decomposes complex tasks into structured task graphs from initial goal prompts without incorporating real-time environmental observations or adaptive feedback mechanisms. HuggingGPT~\cite{HuggingGPT} adopts a global planning strategy to predetermine a complete task sequence from the initial request without incorporating subsequent feedback from the environment during execution. TaskBench~\cite{TaskBench} assesses task automation capabilities by requiring models to generate a tool invocation graph directly from the initial user instruction, omitting interactive feedback from actual tool execution.

\subsubsection{Closed-Loop Environment}
In a closed-loop system, the agent receives new observations $o_t$ at each time step and adapts its actions accordingly. The policy utilizes the dynamically updating history $h_t$ to make reactive decisions, allowing the agent to respond to environmental stochasticity and correct deviations. Mobile-Bench~\cite{Mobile-Bench} implements an iterative execution framework where agents dynamically update their thoughts and actions based on real-time UI feedback and historical environment observations. MCPVerse~\cite{MCPVerse} utilizes an automated agentic loop where models interact with executable tools and leverage real-time feedback to iteratively refine their planning and multi-step actions. WebWalker~\cite{WebWalker} employs an explore-and-critic framework that iteratively adapts its navigation based on real-time observations and dynamic memory to resolve complex web traversal tasks.

\subsection{Online vs. Offline}
\label{attribute:Online vs. Offline}
The core distinction between the two paradigms lies entirely in the interaction mechanism. This mechanism dictates whether an agent actively drives state transitions through sequential exploration with the environment, or is evaluated solely on its ability to mimic expert behavior from pre-recorded trajectories.

\subsubsection{Online Environment}
In an online environment, the agent actively interacts with the true dynamical system $\mathcal{E}$. At each time step $t$, the agent executes an action $a_t$, and the environment sequentially yields the subsequent observation $o_{t+1}$. The agent evaluates and refines its behavior through real-time feedback. For example, WebArena~\cite{WebArena} provides a suite of fully functional websites where agents interact dynamically and receive real-time feedback to achieve complex tasks. EmbodiedBench~\cite{EmbodiedBench} assesses vision-driven agents through real-time interaction across diverse simulated environments , where models must actively process sequential feedback and visual observations to refine their task execution. Terminal-Bench~\cite{Terminal-Bench} benchmarks agents on interactive, long-horizon tasks within real terminal environments, requiring models to autonomously manipulate containerized systems and refine actions based on sequential execution feedback.

\subsubsection{Offline Environment}
In an offline environment, direct interaction with the system $\mathcal{E}$ is strictly prohibited. The agent relies instead on a static dataset of previously sampled trajectories $\mathcal{D}$. Evaluation is conducted statically by comparing the predicted single step action $\pi(a_t | h_t)$ against the expert action $a^*_t$ from human or strong models, severing the standard sequential feedback loop. For example, Mind2Web~\cite{Mind2Web} evaluates agents on cached snapshots of real-world websites by comparing predicted actions against human-annotated trajectories without real-time sequential feedback. ALFRED~\cite{ALFRED} uses a static dataset of expert demonstrations to evaluate models, comparing predicted actions against human-annotated ground truth without real-time environmental feedback. AndroidControl~\cite{AndroidControl} evaluates agents on a static dataset across apps, measuring single-step action accuracy against expert ground truth rather than through real-time environmental interaction.

\subsection{MDP vs. POMDP}
\label{attribute:MDP vs. POMDP}

The core distinction between these two paradigms lies in the mapping relationship between the environment's state space $\mathcal{S}$ and the agent's observation space $\Omega$. Essentially, this determines whether the agent has full access to everything happening in the environment, or if it must rely on incomplete observations to understand the true situation.

\subsubsection{Fully Observable Environment}
In a fully observable environment, typically formalized as a Markov Decision Process (MDP), the agent has complete and transparent access to the true dynamics. The observation space perfectly aligns with the state space, formally $\Omega = \mathcal{S}$. Here, the observation $o_t$ is entirely identical to the latent state $s_t$, allowing the agent to base its policy strictly on the current state without needing historical context, formulated as $\pi(a_t | s_t)$. For example, KOR-Bench~\cite{KOR-Bench} evaluates reasoning capabilities through tasks where all logical rules and complete problem states are explicitly provided in the text prompt. GVGAI-LLM~\cite{GVGAI-LLM} directly presents the evolving game status and foundational principles through formatted text descriptions during every single step. V-GameGym~\cite{V-GameGym} evaluates LLMs on synthesizing complete visual games from explicit text instructions where the sandbox execution provides transparent access to all underlying game states.

\subsubsection{Partially Observable Environment}
In a partially observable environment, modeled as a POMDP, the agent lacks direct access to the complete latent state. The observation $o_t \in \Omega$ serves as an incomplete or noisy projection of $s_t$, strictly governed by the observation function $\mathcal{O}(o_t | s_t, a_t)$. Consequently, the agent is required to aggregate information across its entire interactive trajectory $h_t$ to infer the underlying state distribution and execute optimal decisions. For example, WebArena~\cite{WebArena} evaluates web agents that observe only the focused tab or a limited viewport. Consequently, agents must rely on their interaction history to infer the full underlying website state. GAIA~\cite{GAIA} evaluates general AI assistants on complex real-world questions where agents must iteratively gather information from a diverse environment. EmbodiedBench~\cite{EmbodiedBench} evaluates multimodal agents in 3D simulated tasks where the complete environment state is hidden , requiring models to make decisions relying entirely on egocentric visual observations and interaction history.

\subsection{Deterministic vs. Nondeterministic}
\label{attribute:Deterministic vs. Nondeterministic}
The distinction between deterministic and nondeterministic environments is characterized by the properties of the transition function $\mathcal{P}$ and the reward function $\mathcal{R}$. Essentially, this determines whether taking a specific action in a given situation always produces the exact same, predictable states, or if the outcome involves randomness and uncertainty.

\subsubsection{Deterministic Environment}
In a deterministic environment, executing a specific action $a_t \in \mathcal{A}$ in a given state $s_t \in \mathcal{S}$ invariably results in a single, predictable next state $s_{t+1}$ and reward $r_t$. Formally, the transition kernel collapses into a deterministic mapping, such that $\mathcal{P}(s_{t+1} | s_t, a_t) = 1$ for exactly one target state. GameTraversalBenchmark~\cite{GameTraversalBenchmark} evaluates planning on static 2D maps where discrete directional actions predictably determine the next state. Baba Is AI~\cite{BabaIsAI} introduces a gridworld puzzle game where manipulating objects and textual rules results in predictable and exact state transitions. 
VSP~\cite{Vsp} features classical maze navigation and block-moving scenarios, which serve as deterministic environments because they are governed by strict, predefined rules where every allowed action yields a guaranteed, predictable outcome.

\subsubsection{Nondeterministic Environment}
In a nondeterministic environment, the environmental dynamics are governed by probability distributions. Taking an action $a_t$ in state $s_t$ results in a next state $s_{t+1}$ that is sampled from the distribution defined by the transition kernel $\mathcal{P}(\cdot | s_t, a_t) \in \Delta(\mathcal{S})$. Consequently, the identical state-action pair $(s_t, a_t)$ can yield different resulting states and rewards across independent interactions. Furthermore, environments in which the agent's state transition function is fundamentally difficult to obtain or intractable to model are also practically classified as nondeterministic. Frozen Lake~\cite{Openai_gym} in the slippery ice setting introduces inherent stochasticity through its slippery grid-world mechanics, where intended directional movements have a probability of resulting in transitions to unintended adjacent states. BrowseComp~\cite{BrowseComp} requires agents to persistently navigate the open web to find obscure information, operating within a vast internet environment where state transitions are fundamentally intractable to model. WebVoyager~\cite{WebVoyager} navigates live real-world websites where dynamic elements like constant updates and pop up windows make state transitions hard to predict.

\subsection{Discrete vs. Continuous}
\label{attribute:Discrete vs. Continuous}
The core distinction between the two paradigms lies entirely in the action space $\mathcal{A}$. This determines whether the agent chooses from a fixed, limited set of options or generates exact numerical values across an infinite range.

\subsubsection{Discrete Action Space}
In an environment with a discrete action space, the available actions form a finite and countable set like $\mathcal{A} = \{a_1, a_2, \dots, a_n\}$. The agent's policy typically outputs a categorical probability distribution over these specific choices. ALFWorld~\cite{ALFWorld} evaluates agents on embodied household tasks by restricting interactions to a finite set of text commands and physical action primitives. WebShop~\cite{WebShop} simulates online shopping where agents operate by selecting from a finite set of textual search and button click actions. APPWorld~\cite{AppWorld2024} evaluates interactive coding agents on everyday digital tasks where the agent must select from a finite set of predefined API calls.

\subsubsection{Continuous Action Space}
In an environment with a continuous action space, the available actions span a real-valued vector space: $\mathcal{A} \subseteq \mathbb{R}^d$. The agent cannot enumerate the actions and must instead rely on its policy to output a continuous probability density function to select an action vector. RoboFactory~\cite{RoboFactory} evaluates collaborative manipulation tasks where multiple embodied agents must output continuous joint control signals across eight dimensions to execute precise physical interactions. Scenario Dreamer~\cite{Scenario_Dreamer} evaluates autonomous vehicle planners that navigate using a continuous control space of steering and acceleration commands applied to a realistic bicycle model. EmbodiedBench~\cite{EmbodiedBench} evaluates vision-driven agents on low-level navigation and manipulation tasks that require precise control over real-valued vectors representing translational and rotational displacements.

\subsection{Unimodal vs. Multimodal}
\label{attribute:Unimodal vs. Multimodal}
The core distinction between the two paradigms lies in the modality of the observation space $\Omega$. This determines whether the agent perceives the environment through a single modality or combine several different modalities to understand the environment.

\subsubsection{Unimodal Environment}
In a unimodal environment, the observation space $\Omega$ consists of a single modality, predominantly textual or visual data. The observation $o_t$ is presented in a homogeneous linguistic format, requiring the agent to process and infer the state $s_t$ entirely through text. MetaDrive~\cite{MetaDrive} provides diverse driving scenarios where agents perceive the visual environment through a homogeneous observation modality to benchmark generalizability across unseen traffic scenes. API-Bank~\cite{API-Bank} evaluates tool-augmented capabilities by requiring models to plan and execute API calls through a textual interface consisting of natural language dialogues and structured documentation. OSWorld-MCP~\cite{OSWorld-MCP} evaluates computer-use agents by standardizing tool invocation through the text-based Model Context Protocol (MCP) to facilitate structured context ingestion and interaction.

\subsubsection{Multimodal Environment}
In a multimodal environment, the observation space $\Omega$ spans across heterogeneous sensory modalities, typically integrating text, images, and videos. The observation becomes a composite representation, formally $o_t = \langle o_t^{\text{text}}, o_t^{\text{image}}, o_t^{\text{video}} \rangle$. Consequently, the agent is required to align and fuse these diverse observational streams to accurately comprehend the environment. VisualWebArena~\cite{VisualWebArena} evaluates multimodal agents on realistic web tasks requiring the integration of images and html to navigate and execute actions on complex websites. AgentStudio~\cite{AgentStudio} offers a realistic environment with universal observation spaces that integrate text, image, and video modalities to evaluate general virtual agents. MLE-bench~\cite{MLE-bench} evaluates AI agents on machine learning engineering tasks requiring the processing of diverse data modalities including text, images, and signal data. 

\subsection{Single-Agent vs. Multi-Agent}
\label{attribute:Single-Agent vs. Multi-Agent}
The key difference is the number of agents interacting with the environment. This determines whether the environment's state transitions and rewards depend on a single action or a combination of actions from multiple agents.
\subsubsection{Single-Agent Environment}
In a single-agent environment, there exists only one decision-making entity $\pi$ interacting with the system. The environment's physical dynamics and the resulting state transitions $\mathcal{P}(s_{t+1} | s_t, a_t)$ are exclusively influenced by the single agent's action $a_t \in \mathcal{A}$. For example, ScienceWorld~\cite{ScienceWorld} evaluates the scientific reasoning of a single autonomous agent completing elementary science experiments within an interactive text environment. SWE-Bench~\cite{SWE-Bench} tasks a single LLM with autonomously editing large codebases to resolve real-world GitHub issues. OSWorld-MCP~\cite{OSWorld-MCP} assesses a single agent that autonomously selects between graphical actions and tool invocations to complete real-world computer tasks.

\subsubsection{Multi-Agent Environment}
In a multi-agent environment, a set of $N$ autonomous agents, indexed by $i \in \{1, \dots, N\}$, interact simultaneously within the shared environment $\mathcal{E}$. The action space expands to a joint action space formed by the Cartesian product of individual action spaces, $\boldsymbol{\mathcal{A}} = \mathcal{A}^1 \times \dots \times \mathcal{A}^N$. Consequently, the environment's transition function is driven by the joint action $\boldsymbol{a}_t = \langle a_t^1, \dots, a_t^N \rangle \in \boldsymbol{\mathcal{A}}$, formulated as $\mathcal{P}(s_{t+1} | s_t, \boldsymbol{a}_t)$. Each agent $i$ is governed by its own policy $\pi^i$ and receives a reward $\mathcal{R}^i(s_t, \boldsymbol{a}_t)$, which may be competitive, cooperative, or mixed, depending on the multi-agent formulation. For example, Generative agents~\cite{Generative-agents} presents a simulated village containing twenty five unique characters powered by LLMs. It operates as an environment with multiple agents because these digital entities autonomously converse, form relationships, and organize collective activities. Collab-overcooked~\cite{Collab-overcooked} introduces an operative cooking simulation where distinct LLMs manage separated resources and uneven information. To succeed, these isolated entities must actively converse and deliver items to achieve shared goals. AvalonBench~\cite{AvalonBench} proposes a social deduction environment where LLMs are assigned hidden roles with asymmetrical knowledge. To succeed in this competitive game, agents must engage in strategic dialogue, utilizing deception and logical reasoning to either cooperate on tasks or actively sabotage the group's missions.

\begin{tcolorbox}[takeaway,title={Takeaway 3}]
\begin{itemize}
    \item \textbf{Multi-Agent Environment:} While most existing frameworks are limited to single-agent scenarios, future designs should incorporate multi-agent cooperative and competitive environments to realize collective intelligence and enhance parallel processing efficiency.
    \item \textbf{Symbolic-Neural Environment Joint Modeling:} Current agents face a deployment bottleneck, caught between rigid, schema-bound traditional frameworks (stable but inflexible) and fully generative world models (open but uncontrollable and inconsistent). Future infrastructures must bridge this gap by providing an intermediate environment that combines the engineering reliability of symbolic systems with the infinite generative scalability of neural models.
\end{itemize}
\end{tcolorbox}

\section{Environment Domain} \label{sec:env_domain}

\begin{figure*}[t]
    \centering
    \includegraphics[width=1\textwidth]{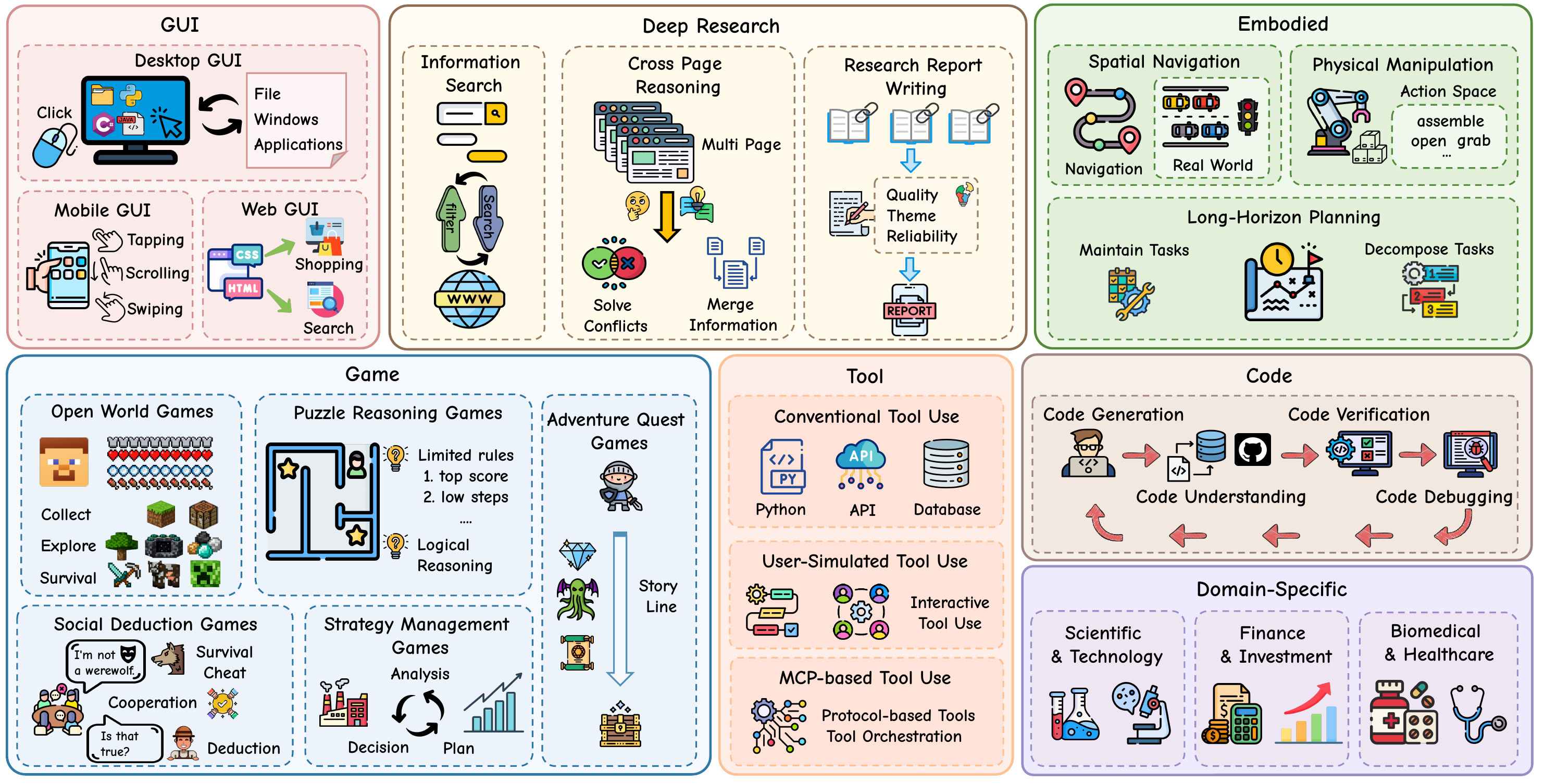}
    \caption{
    An overview of environment domains, including GUI, Deep Research, Embodied, Game, Tool, Code, and Domain-Specific.
    }
    \label{fig:environment_domain_overview}
\end{figure*}

As shown in Fig. \ref{fig:environment_domain_overview}, agent evaluation spans a diverse set of environments, with different domains placing substantially different demands on agents. Some environments focus on grounded perception and interface manipulation, some emphasize long-horizon planning and interactive decision making, and others require evidence synthesis, tool use, or domain-specific professional reasoning. Consequently, existing evaluation environments can be broadly organized into several domains, including \textbf{GUI} (\cref{GUI}), \textbf{Deep Research} (\cref{sec:Deep Research}), \textbf{Embodied} (\cref{Embodied}), \textbf{Game} (\cref{Game}), \textbf{Tool} (\cref{Tool}), \textbf{Code} (\cref{Code}), \textbf{Domain-Specific} (\cref{Domain-Specific}), and \textbf{Cross-Domain} (\cref{Cross-Domain}). This taxonomy reflects a broader shift in agent evaluation from narrow, task-specific benchmarks toward more diverse, realistic, and capability-compositional environments.

\begin{table*}
\caption{An overview of various environments categorized into \textbf{GUI} and \textbf{Deep Research} domains. Within the Modality column, specific icons are utilized to denote the supported data types where \protect\Text represents text, \protect\Image represents images, and \protect\Video represents videos. Additionally, the \protect\Yes and \protect\No icons indicate the presence or absence of specific attributes across the columns, while the resource icons provide direct links to the respective GitHub \protect\ghlink{https://github.com/} or Hugging Face \protect\hflink{https://huggingface.co/} repositories.}
\resizebox{\textwidth}{!}{
\rowcolors{2}{white}{gray!10} 
\renewcommand{\arraystretch}{1.3}
\begin{tabular}{c|lccccccc} \toprule
\textbf{Domain} & \textbf{Name} & \textbf{Size} & \textbf{Modality} & \textbf{Observability} & \textbf{Multi-agent} & \textbf{Continuity} & \textbf{Online} & \textbf{Resource} \\ \midrule

\cellcolor{white} & WebShop~\cite{WebShop} & 500 & \Text\hspace{-0.4em}\Image & Partially & \No & Discrete & \Yes & \ghlink{https://github.com/princeton-nlp/WebShop}\\
\cellcolor{white} & Mind2Web~\cite{Mind2Web} & 1,341 & \Text\hspace{-0.4em}\Image & Partially & \No & Discrete & \No & \ghlink{https://github.com/OSU-NLP-Group/Mind2Web}\\
\cellcolor{white} & Mind2Web2~\cite{Mind2Web2} & 120 & \Text\hspace{-0.4em}\Image & Partially & \No & Discrete & \Yes & \ghlink{https://github.com/OSU-NLP-Group/Mind2Web-2}\\
\cellcolor{white} & WebArena~\cite{WebArena} & 812 & \Text\hspace{-0.4em}\Image & Partially & \No & Discrete & \Yes & \ghlink{https://github.com/web-arena-x/webarena}\\
\cellcolor{white} & VisualWebArena~\cite{VisualWebArena} & 910 & \Text\hspace{-0.4em}\Image & Partially & \No & Discrete & \Yes & \ghlink{https://github.com/web-arena-x/visualwebarena}\\

\cellcolor{white} & WebVoyager~\cite{WebVoyager} & 643 & \Text\hspace{-0.4em}\Image & Partially & \No & Discrete & \Yes & \ghlink{https://github.com/MinorJerry/WebVoyager}\\
\cellcolor{white} & Mobile-Env~\cite{Mobile-Env} & 74 & \Text\hspace{-0.4em}\Image & Partially & \No & Mixed & \Yes & \ghlink{https://github.com/X-LANCE/Mobile-Env}\\
\cellcolor{white} & AitW~\cite{AitW} & 715,142 & \Text\hspace{-0.4em}\Image & Partially & \No & Mixed & \No & \ghlink{https://github.com/google-research/google-research/tree/master/android_in_the_wild}\\
\cellcolor{white} & AndroidWorld~\cite{AndroidWorld} & 116 & \Text\hspace{-0.4em}\Image & Partially & \No & Mixed & \Yes & \ghlink{https://github.com/google-research/android_world}\\
\cellcolor{white} & AndroidControl~\cite{AndroidControl} & 721 & \Text\hspace{-0.4em}\Image & Partially & \No & Mixed & \No & \ghlink{https://github.com/google-research/google-research/tree/master/android_control}\\

\cellcolor{white} & MobileWorld~\cite{MobileWorld} & 201 & \Text\hspace{-0.4em}\Image & Partially & \No & Mixed & \Yes & \ghlink{https://github.com/Tongyi-MAI/MobileWorld}\\
\cellcolor{white} & AgentStudio~\cite{AgentStudio} & 205 & \Text\hspace{-0.4em}\Image\hspace{-0.4em}\Video & Partially & \No & Mixed & \Yes & \ghlink{https://ltzheng.github.io/agent-studio/}\\
\cellcolor{white} & WorkArena~\cite{WorkArena} & 19,912 & \Text\hspace{-0.4em}\Image & Partially & \No & Mixed & \Yes & \ghlink{https://github.com/ServiceNow/WorkArena}\\
\cellcolor{white} & OSWorld~\cite{OSWorld} & 369 & \Text\hspace{-0.4em}\Image & Partially & \No & Mixed & \Yes & \ghlink{https://github.com/xlang-ai/OSWorld}\\
\cellcolor{white} & WindowsAgentArena~\cite{WindowsAgentArena} & 154 & \Text\hspace{-0.4em}\Image & Partially & \No & Mixed & \Yes & \ghlink{https://github.com/microsoft/WindowsAgentArena}\\

\cellcolor{white} & OSWorld-MCP~\cite{OSWorld-MCP} & 361 & \Text & Partially & \No & Mixed & \Yes & \ghlink{https://github.com/X-PLUG/OSWorld-MCP}\\
\cellcolor{white} & Mobile-Bench~\cite{Mobile-Bench} & 832 & \Text\hspace{-0.4em}\Image 
& Partially & \No & Discrete & \Yes & \ghlink{https://github.com/XiaoMi/MobileBench}\\
\cellcolor{white} & MT-Mind2Web~\cite{MT-Mind2Web} & 120 & \Text & Partially & \No & Discrete & \No & \hflink{https://huggingface.co/datasets/magicgh/MT-Mind2Web}\\
\cellcolor{white} & VideoWebArena~\cite{VideoWebArena} & 2,021 & \Text\hspace{-0.4em}\Image\hspace{-0.4em}\Video & Partially & \No & Mixed & \Yes & \ghlink{https://github.com/ljang0/videowebarena}\\
\cellcolor{white} & Mind2Web-Live~\cite{Mind2Web-Live} & 104 & \Text\hspace{-0.4em}\Image & Partially & \No & Discrete & \Yes & \hflink{https://huggingface.co/datasets/iMeanAI/Mind2Web-Live}\\
\cellcolor{white} & Online-Mind2Web~\cite{Online-Mind2Web} & 300 & \Text\hspace{-0.4em}\Image & Partially & \No & Discrete & \Yes & \hflink{https://huggingface.co/datasets/osunlp/Online-Mind2Web}\\
\cellcolor{white} \multirow{-22}{*}{\makecell{\textbf{GUI} \\ (\cref{GUI})}} & MobileAgentBench~\cite{MobileAgentBench} & 100 & \Text\hspace{-0.4em}\Image & Partially & \No & Mixed & \Yes & \ghlink{https://github.com/MobileAgentBench/mobile-agent-bench}\\
\midrule

\cellcolor{white} & HLE~\cite{HLE} & 2,500 & \Text\hspace{-0.4em}\Image & Partially & \No & Discrete & \Yes & \raisebox{-0.2em}{\hflink{https://huggingface.co/datasets/cais/hle}}\\
\cellcolor{white} & DeepResearch Bench~\cite{DeepResearch-Bench} & 100 & \Text & Partially & \No & Discrete & \Yes & \ghlink{https://github.com/Ayanami0730/deep_research_bench}\\
\cellcolor{white} & GAIA~\cite{GAIA} & 466 & \Text & Partially & \No & Discrete & \Yes & \raisebox{-0.2em}{\hflink{https://huggingface.co/gaia-benchmark}}\\
\cellcolor{white} & WideSearch~\cite{WideSearch} & 200 & \Text & Partially & \No & Discrete & \Yes & \raisebox{-0.2em}{\hflink{https://huggingface.co/datasets/ByteDance-Seed/WideSearch}}\\
\cellcolor{white} & WebWalker~\cite{WebWalker} & 680 & \Text & Partially & \No & Discrete & \Yes & \ghlink{https://github.com/Alibaba-NLP/DeepResearch}\\

\cellcolor{white} & BrowseComp~\cite{BrowseComp} & 1,266 & \Text & Partially & \No & Discrete & \Yes & \ghlink{https://github.com/openai/simple-evals}\\
\cellcolor{white} & Conflicts~\cite{Conflicts} & 458 & \Text & Partially & \No & Discrete & \Yes & \ghlink{https://github.com/google-research-datasets/rag_conflicts}\\
\cellcolor{white} & DeepResearchGym~\cite{DeepResearchGym} & 1,000 & \Text & Partially & \No & Discrete & \Yes & \ghlink{https://github.com/cxcscmu/deepresearch_benchmarking}\\
\cellcolor{white} & InfoDeepSeek~\cite{InfoDeepSeek} & 245 & \Text & Partially & \No & Discrete & \Yes & \ghlink{https://github.com/YunjiaXi/InfoDeepSeek}\\
\cellcolor{white} & InfoSeek~\cite{InfoSeek} & 830 & \Text & Partially & \No & Discrete & \Yes & \ghlink{https://github.com/VectorSpaceLab/Infomatica}\\
\cellcolor{white}   & BrowseComp-ZH~\cite{BrowseComp-ZH} & 289 & \Text & Partially & \No & Discrete & \Yes & \ghlink{https://github.com/PALIN2018/BrowseComp-ZH}\\

\cellcolor{white} & SurveyGen~\cite{SurveyGen} & 4,200 & \Text & Partially & \No & Discrete & \No & \ghlink{https://github.com/tongbao96/SurveyGen}\\
\cellcolor{white} & ReportBench~\cite{ReportBench} & 100 & \Text & Partially & \No & Discrete & \Yes & \ghlink{https://github.com/ByteDance-BandAI/ReportBench}\\
\cellcolor{white} & LiveDRBench~\cite{LiveDRBench} & 100 & \Text & Partially & \No & Discrete & \Yes & \ghlink{https://github.com/microsoft/LiveDRBench}\\
\cellcolor{white} & ScholarQABench~\cite{ScholarQABench} & 2,967 & \Text & Partially & \No & Discrete & \No & \ghlink{https://github.com/AkariAsai/ScholarQABench}\\
\cellcolor{white} & ResearcherBench~\cite{ResearcherBench} & 65 & \Text & Partially & \No & Discrete & \Yes & \ghlink{https://github.com/GAIR-NLP/ResearcherBench}\\
\cellcolor{white} & ProxyQA~\cite{ProxyQA} & 100 & \Text & Partially & \No & Discrete & \No & \ghlink{https://github.com/Namco0816/ProxyQA}\\
\cellcolor{white} & AI Idea Bench 2025~\cite{AI_Idea_Bench_2025} & 3,495 & \Text & Partially & \No & Discrete & \No & \ghlink{https://github.com/yansheng-qiu/AI_Idea_Bench_2025}\\
\cellcolor{white} & DeepReview~\cite{DeepReview} & 13,378 & \Text & Partially & \No & Discrete & \No & \ghlink{https://github.com/ResearAI/DeepReviewer-v2}\\
\cellcolor{white} & PaperBench~\cite{PaperBench} & 20 & \Text & Partially & \No & Discrete & \No & \ghlink{https://github.com/openai/preparedness/tree/main/project/paperbench}\\
\cellcolor{white} & Multimodal-DeepResearcher~\cite{Multimodal_DeepResearcher} & 100 & \Text\hspace{-0.4em}\Image & Partially & \No & Discrete & \Yes & \ghlink{https://github.com/rickyang1114/multimodal-deepresearcher}\\
\cellcolor{white} & Vision-DeepResearch~\cite{Vision_DeepResearch} & 2,000 & \Text\hspace{-0.4em}\Image & Partially & \No & Discrete & \Yes & \ghlink{https://github.com/Osilly/Vision-DeepResearch}\\
\cellcolor{white} & MMDR-Bench~\cite{MMDeepResearch_Bench} & 140 & \Text\hspace{-0.4em}\Image & Partially & \No & Discrete & \Yes & \ghlink{https://github.com/AIoT-MLSys-Lab/MMDeepResearch-Bench}\\
\cellcolor{white} \multirow{-25}{*}{\makecell{\textbf{Deep Research} \\ (\cref{sec:Deep Research})}} & OmniGAIA~\cite{OmniGAIA} & 360 & \Text\hspace{-0.4em}\Image\hspace{-0.4em}\Video & Partially & \No & Discrete & \Yes & \ghlink{https://github.com/RUC-NLPIR/OmniGAIA}\\

 \bottomrule
\end{tabular}}
\end{table*}

\subsection{GUI}
\label{GUI}
\textbf{GUI} environments denote settings in which agents complete tasks through the perception and manipulation of graphical user interfaces~\cite{OSWorld,WindowsAgentArena,AgentStudio,AitW,Mobile-Env,AndroidWorld,MobileAgentBench,WebShop,Mind2Web,WebArena,VisualWebArena}. These environments emphasize the ability to comprehend, localize and interact with interface elements, as well as to make sequential decisions and execute actions based on interface feedback. Based on the primary interaction platforms, existing work can be broadly categorized into three groups: \textbf{Desktop GUI} (\cref{Desktop GUI}), \textbf{Mobile GUI} (\cref{Mobile GUI}), and \textbf{Web GUI} (\cref{Web GUI}).

\subsubsection{Desktop GUI}
\label{Desktop GUI}
\textbf{Desktop GUI} concerns tasks carried out in operating systems, where agents need to interact with files, windows, applications, and system tools. Compared with other GUI settings, this domain places greater demands on long-horizon planning, cross application coordination, and stable grounding in complex visual interfaces. OSWorld~\cite{OSWorld} evaluates open-ended tasks in realistic desktop environments and covers a wide range of everyday computer operations. WindowsAgentArena~\cite{WindowsAgentArena} further focuses on practical Windows workflows and highlights the challenges of interacting with native system components and desktop software. AgentStudio~\cite{AgentStudio} complements these environments by providing a more general platform for building and evaluating GUI agents, which supports broader research on desktop interaction. Overall, Desktop GUI is moving toward more realistic and more general computer use, making it an important domain for evaluating general purpose agents.

\subsubsection{Mobile GUI}
\label{Mobile  GUI}
\textbf{Mobile GUI} focuses on agent interaction with smartphone interfaces, where tasks are completed through tapping, typing, scrolling, and swiping on small screens. Compared with desktop settings, mobile environments are more constrained by limited display space, deeper page transitions, and stronger dependence on app specific layouts. AitW~\cite{AitW} studies Android control from human demonstrations and natural language instructions, establishing mobile GUI as an important grounded interaction problem. Mobile-Env~\cite{Mobile-Env} moves this line of research toward controllable evaluation by providing an interactive Android environment for task construction and testing. More recent benchmarks such as AndroidWorld~\cite{AndroidWorld} and MobileAgentBench~\cite{MobileAgentBench} further emphasize executable and reproducible evaluation, pushing the field from static demonstration data toward realistic mobile task completion. Overall, Mobile GUI has gradually evolved from static demonstration-based settings toward more interactive and realistic mobile environments, making it a key domain for evaluating end-to-end mobile agents.

\subsubsection{Web GUI}
\label{Web GUI}
\textbf{Web GUI} centers on agent interaction with browser interfaces to perform tasks such as navigation, information retrieval, form filling, and shopping. Compared with other GUI settings, this domain combines textual content, visual layout, structured page elements, and dynamic interaction logic, which makes it one of the main testbeds for GUI agents. WebShop~\cite{WebShop} frames online shopping as sequential decision making over webpages. Mind2Web~\cite{Mind2Web} extends this setting to real websites from diverse domains and highlights generalization to unseen sites and tasks. WebArena~\cite{WebArena} further advances the field by providing executable web environments for realistic long-horizon workflows, while VisualWebArena~\cite{VisualWebArena} shows that rendered visual information is often necessary for successful web interaction. Web GUI now serves as a central benchmark for studying agent performance in dynamic, information-rich, and workflow-oriented online environments.

\subsection{Deep Research}
\label{sec:Deep Research}
% Deep Research 是指智能体围绕开放式研究目标，在网页、搜索引擎、数据库和文档库中持续进行信息检索、证据筛选、来源核验与结果整合，最终形成有依据的回答或研究报告。与只要求单步检索或单页问答的任务不同，Deep Research 更强调长程信息搜集、跨来源证据整合以及面向用户目标的结果组织能力。按照任务链路中的主要瓶颈，现有工作大致可以分为 \textbf{Information Search}、\textbf{Cross Page Reasoning} 和 \textbf{Research Report Writing} 三类。前一类关注能否找到信息，中间一类关注能否整合信息，后一类关注能否产出结构化研究结果。

\textbf{Deep Research} refers to the process where an agent continuously retrieves information, screens evidence, and synthesizes findings from diverse sources to address a research goal, ultimately generating answers grounded in evidence. Based on the major bottlenecks involved, existing work can be broadly divided into three categories (\cref{Information Search,Multi-Source Reasoning,Research Report Writing}). The first~\cite{WebWalker,WideSearch,InfoDeepSeek} centers on finding relevant information, referred to as \textbf{Information Search}; the second~\cite{GAIA,BrowseComp,Conflicts,BrowseComp-ZH} emphasizes synthesizing information across sources, referred to as \textbf{Multi-Source Reasoning}; and the third~\cite{DeepResearch-Bench,DeepResearchGym,DR_BENCH} targets the production of structured research outputs, referred to as \textbf{Research Report Writing}.

\subsubsection{Information Search}
\label{Information Search}
\textbf{Information Search} examines whether an agent can effectively search the open web and progressively extend its exploration to uncover high quality evidence. SimpleQA~\cite{SimpleQA} evaluates the ability to retrieve precise factual information for question answering, capturing the most basic form of web information seeking. 
WideSearch~\cite{WideSearch} focuses on broad information collection across sources, highlighting the gathering and organization of large numbers of atomic facts. 
InfoDeepSeek~\cite{InfoDeepSeek} evaluates information gathering in Agentic RAG as a standalone capability, focusing on how agents decide whether to search, what to search for, and which evidence to retain.
Overall, this line of work shifts Deep Research from single retrieval to persistent exploration and from finding a single answer to covering a body of valuable evidence.

\subsubsection{Multi-Source Reasoning}
\label{Multi-Source Reasoning}
\textbf{Multi-Source Reasoning} concerns whether an agent can synthesize information from multiple sources by comparing evidence, aligning claims, drawing inferences, and resolving conflicts. GAIA~\cite{GAIA} requires models to combine reasoning, web browsing, and tool use to solve real world problems that often cannot be answered from a single page. 
WebWalker~\cite{WebWalker} also relates to this setting, as solving its tasks often requires agents to move through hierarchical website structures and connect evidence distributed across multiple linked pages. 
BrowseComp~\cite{BrowseComp} further highlights this ability by requiring sustained navigation and the integration of scattered clues, making it well suited for evaluating cross page evidence synthesis. 
Conflicts~\cite{Conflicts} focuses explicitly on contradictions across sources, testing whether models can detect conflicts and distinguish their types. With benchmarks such as BrowseComp-ZH~\cite{BrowseComp-ZH}, this line of work has also expanded from English webpages to more complex Chinese information environments. Overall, the key capability in Deep Research is not reproducing a single source but relating multiple pieces of evidence to form a grounded conclusion.

\subsubsection{Research Report Writing}
\label{Research Report Writing}
\textbf{Research Report Writing} concerns whether an agent can transform collected evidence into structured, verifiable research outputs that are aligned with user needs. DeepResearch Bench~\cite{DeepResearch-Bench} evaluates deep research agents with an emphasis on iterative web exploration, targeted retrieval, advanced synthesis, and report quality. DeepResearchGym~\cite{DeepResearchGym} highlights reproducible evaluation by providing stable search interfaces and protocols for more controlled comparison of report generation. DR.BENCH~\cite{DR_BENCH} focuses more directly on long research reports, using a multidimensional framework that evaluates not only content quality but also topical focus and retrieval reliability. Overall, while earlier work mainly evaluates key stages of the research process, such as information search and multi-source reasoning, this line of work shifts the focus toward the quality of the final deliverable.

\begin{table*}
\caption{An overview of various environments categorized into \textbf{Embodied} and \textbf{Game} domains.}
\resizebox{\textwidth}{!}{
\rowcolors{2}{white}{gray!10} 
\renewcommand{\arraystretch}{1.3}
\begin{tabular}{c|lccccccc} 
\toprule
\textbf{Domain} & \textbf{Name} & \textbf{Size}  & \textbf{Modality} & \textbf{Observability} & \textbf{Multi-agent} & \textbf{Continuity} & \textbf{Online} & \textbf{Resource} \\ \midrule
\cellcolor{white} & Habitat~\cite{Habitat} & -- & \Image & Partially & \No & Discrete & \Yes & \ghlink{https://github.com/peteanderson80/Matterport3DSimulator}\\
\cellcolor{white} & Room-to-Room~\cite{Room-to-Room} & 4173 & \Text\hspace{-0.4em}\Image & Partially & \No & Discrete & \Yes & \ghlink{https://github.com/peteanderson80/Matterport3DSimulator}\\
\cellcolor{white} & RLBench~\cite{RLBench} & 100 & \Text\hspace{-0.4em}\Image & Partially & \No & Mixed & \Yes & \ghlink{https://github.com/stepjam/RLBench}\\
\cellcolor{white} & ALFRED~\cite{ALFRED} & 1,529 & \Text\hspace{-0.4em}\Image & Partially & \No & Discrete & \No & \ghlink{https://github.com/askforalfred/alfred}\\
\cellcolor{white} & ALFWorld~\cite{ALFWorld} & 134 & \Text & Partially & \No & Discrete & \Yes & \ghlink{https://github.com/alfworld/alfworld}\\
\cellcolor{white} & Robocasa~\cite{Robocasa} & 100 & \Image & Partially & \No & Continuous & \Yes & \ghlink{https://github.com/robocasa/robocasa}\\
\cellcolor{white} & BEHAVIOR~\cite{BEHAVIOR} & 100 & \Image & Partially & \No & Continuous & \Yes & \ghlink{https://stanfordvl.github.io/behavior/intro.html}\\
\cellcolor{white} & panda-gym~\cite{panda-gym} & 80 & \Text & Completely & \No & Continuous & \Yes & \ghlink{https://github.com/qgallouedec/panda-gym}\\
\cellcolor{white} & MetaDrive~\cite{MetaDrive} & 100 & \Image & Partially & \Yes & Continuous & \Yes & \ghlink{https://github.com/metadriverse/metadrive}\\
\cellcolor{white} & ScienceWorld~\cite{ScienceWorld} & 1,800 & \Text & Partially & \No & Discrete & \Yes & \ghlink{https://github.com/allenai/ScienceWorld}\\
\cellcolor{white} & ET-Plan-Bench~\cite{ET-Plan-Bench} & 11,838 & \Text\hspace{-0.4em}\Image & Partially & \No & Discrete & \Yes & \ghlink{https://github.com/ET-Plan-Bench/ET-Plan-Bench}\\
\cellcolor{white} & LEGENT~\cite{LEGENT} & 100 & \Text\hspace{-0.4em}\Image & Partially & \No & Mixed & \Yes & \ghlink{https://github.com/thunlp/LEGENT}\\
\cellcolor{white} & EmbodiedBench~\cite{EmbodiedBench} & 1,128 & \Text\hspace{-0.4em}\Image & Partially & \No & Mixed & \Yes & \ghlink{https://github.com/EmbodiedBench/EmbodiedBench}\\
\cellcolor{white} & RoboFactory~\cite{RoboFactory} & 1,100 & \Image & Partially & \Yes & Continuous & \Yes & \ghlink{https://github.com/MARS-EAI/RoboFactory}\\
\cellcolor{white} & Scenario Dreamer~\cite{Scenario_Dreamer} & 50,000 & \Image & Partially & \No & Continuous & \Yes & \ghlink{https://github.com/princeton-computational-imaging/scenario-dreamer}\\
\cellcolor{white} & SimuHome~\cite{SimuHome} & 600 & \Text & Partially & \No & Discrete & \Yes & \ghlink{https://github.com/holi-lab/SimuHome}\\
\cellcolor{white} & Sari Sandbox~\cite{Sari_Sandbox} & 100 & \Text\hspace{-0.4em}\Image & Partially & \No & Mixed & \Yes & \ghlink{https://github.com/upeee/sari-sandbox-env}\\
\cellcolor{white} \multirow{-18}{*}{\makecell{\textbf{Embodied}\\(\cref{Embodied}) }} & Nexus~\cite{Nexus} & 1,090 & \Image & Completely & \Yes & Continuous & \No & \ghlink{https://github.com/OpenDriveLab/Nexus}\\ 
\midrule

\cellcolor{white} & MineDojo~\cite{MineDojo} & 3,142 & \Text\hspace{-0.4em}\Image & Partially & \No & Discrete & \Yes & \ghlink{https://github.com/MineDojo/MineDojo}\\
\cellcolor{white} & KORGym~\cite{KORGym} & 55,200 & \Text\hspace{-0.4em}\Image & Partially & \No & Discrete & \Yes & \ghlink{https://github.com/multimodal-art-projection/KORGym}\\
\cellcolor{white} & TextArena~\cite{TextArena} & 74 & \Text & Partially & \Yes & Discrete & \Yes & \ghlink{https://github.com/TextArena/TextArena}\\
\cellcolor{white} & V-GameGym~\cite{V-GameGym} & 2,219 & \Text & Completely & \No & Discrete & \No & \raisebox{-0.2em}{\hflink{https://huggingface.co/datasets/alibabagroup/SKYLENAGE-GameCodeGym}}\\
\cellcolor{white} & Kor-Bench~\cite{KOR-Bench} & 1,250 & \Text & Completely & \No & Discrete & \No & \ghlink{https://github.com/KOR-Bench/KOR-Bench}\\
\cellcolor{white} & AI Gamestore~\cite{AI_GAMESTORE} & 2,100 & \Image & Partially & \No & Mixed & \Yes & \ghlink{https://github.com/lance-ying/aigamestore_harness}\\
\cellcolor{white} & ING-VP~\cite{ING-VP} & 300 & \Image & Partially & \No & Discrete & \Yes & \ghlink{https://github.com/thisisus7/ing-vp}\\
\cellcolor{white} & VideoGameBench~\cite{VideoGameBench} & 50 & \Image & Partially & \No & Mixed & \Yes & \ghlink{https://github.com/alexzhang13/videogamebench}\\
\cellcolor{white} & GVGAI-LLM~\cite{GVGAI-LLM} & 540 & \Text & Completely & \No & Discrete & \Yes & \ghlink{https://github.com/zmuhls/gvgai-web}\\
\cellcolor{white} & BALROG~\cite{BALROG} & 425 & \Text\hspace{-0.4em}\Image & Partially & \No & Discrete & \Yes & \ghlink{https://github.com/balrog-ai/BALROG}\\
\cellcolor{white} & Smartplay~\cite{SMARTPLAY} & 931,900 & \Text & Partially & \No & Discrete & \Yes & \ghlink{https://github.com/microsoft/SmartPlay}\\
\cellcolor{white} & DSGBench~\cite{DSGBench} & 160 & \Image & Partially & \Yes & Discrete & \Yes & \ghlink{https://github.com/DeciBrain-Group/DSGBench}\\
\cellcolor{white} & GTBench~\cite{GTBENCH} & 500 & \Text & Partially & \Yes & Discrete & \Yes & \ghlink{https://github.com/jinhaoduan/GTBench}\\
%\cellcolor{white} & Lmage-Bench~\cite{LMGAME-BENCH} & 319 & \Image & Partially & \No & Discrete & \Yes & \ghlink{https://github.com/lmgame-org/GamingAgent}\\
%\cellcolor{white} & Factorio Learning Environment~\cite{FactorioLearningEnvironment} & 96 & \Text & Partially & \No & Discrete & \Yes & \ghlink{https://github.com/JackHopkins/factorio-learning-environment}\\
\cellcolor{white} & Pok\'eLLMon~\cite{PokeLLMon} & 1,909 & \Text & Partially & \Yes & Discrete & \Yes & \ghlink{https://github.com/git-disl/PokeLLMon}\\
\cellcolor{white} & MC-Planner~\cite{MC-Planner} & 2,130 & \Text\hspace{-0.4em}\Image & Partially & \No & Mixed & \Yes & \ghlink{https://github.com/CraftJarvis/MC-Planner}\\
\cellcolor{white} & GameTraversalBenchmark~\cite{GameTraversalBenchmark} & 150 & \Text & Completely & \No & Discrete & \Yes & \ghlink{https://github.com/umair-nasir14/Game-Traversal-Benchmark}\\
%\cellcolor{white} & FlashAdventure~\cite{FlashAdventure} & 238 & \Image & Partially & \No & Mixed & \Yes & \ghlink{https://github.com/ahnjaewoo/FlashAdventure}\\
\cellcolor{white} & LMRL Gym~\cite{LMRLGym} & 6,541 & \Text & Partially & \No & Discrete & \Yes & \ghlink{https://github.com/abdulhaim/LMRL-Gym}\\
%\cellcolor{white} & MAgIC~\cite{MAgIC} & 375 & \Text & Partially & \Yes & Discrete & \Yes & \ghlink{https://github.com/cathyxl/MAgIC}\\
\cellcolor{white} & Clembench~\cite{clembench} & 250 & \Text & Partially & \Yes & Discrete & \Yes & \ghlink{https://github.com/clp-research/clembench}\\
\cellcolor{white} & Baba Is AI~\cite{BabaIsAI} & 1,275 & \Image & Completely & \No & Discrete & \Yes & \ghlink{https://github.com/nacloos/baba-is-ai}\\
\cellcolor{white} & HLA~\cite{HLA} & 330 & \Text\hspace{-0.4em}\Image & Completely & \Yes & Discrete & \Yes & \ghlink{https://github.com/HosnLS/Hierarchical-Language-Agent}\\
\cellcolor{white} & LLMArena~\cite{LLMARENA} & 5,600 & \Text & Partially & \Yes & Discrete & \Yes & \ghlink{https://github.com/THU-BPM/LLMArena}\\
%\cellcolor{white} & WhodunitBench~\cite{WhodunitBench} & 3,050 & \Text\hspace{-0.4em}\Image & Partially & \No & Discrete & \Yes & \ghlink{https://github.com/jun0wanan/WhodunitBench-Murder_Mystery_Games}\\
\cellcolor{white} & TextGames~\cite{TEXTGAMES} & 24,000 & \Text & Completely & \No & Discrete & \Yes & \ghlink{https://github.com/fhudi/textgames}\\
\cellcolor{white} & GameArena~\cite{GameArena} & 2,240 & \Text & Partially & \Yes & Discrete & \Yes & \ghlink{https://github.com/lmgame-org/ai-space-escape-engine}\\
\cellcolor{white} & SPIN-Bench~\cite{SPIN-Bench} & 1,280 & \Text & Partially & \Yes & Discrete & \Yes & \ghlink{https://github.com/spinbench/spinbench}\\
\cellcolor{white} & GAMEBoT~\cite{GAMEBoT} & 2,720 & \Text & Partially & \Yes & Discrete & \Yes & \ghlink{https://github.com/Visual-AI/GAMEBoT}\\
\cellcolor{white} & GameBench~\cite{GAMEBENCH} & 277 & \Text & Partially & \Yes & Discrete & \Yes & \ghlink{https://github.com/Joshuaclymer/GameBench}\\
\cellcolor{white} & MineWorld~\cite{MINEWORLD} & 1,000 & \Image & Partially & \No & Mixed & \Yes & \ghlink{https://github.com/microsoft/mineworld}\\
\cellcolor{white} & AvalonBench~\cite{AvalonBench} & 1,140 & \Text & Partially & \Yes & Discrete & \Yes & \ghlink{https://github.com/jonathanmli/Avalon-LLM}\\
\cellcolor{white} & Werewolf~\cite{Werewolf} & 1,200 & \Text & Partially & \Yes & Discrete & \Yes & \ghlink{https://github.com/boluoweifenda/werewolf}\\
%\cellcolor{white} & gg-bench~\cite{gg-bench} & 3,780 & \Text & Completely & \Yes & Discrete & \Yes & \ghlink{https://github.com/vivek3141/gg-bench}\\
\cellcolor{white} & GameFactory~\cite{GAMEFACTORY} & 7,407 & \Image\hspace{-0.4em}\Video & Partially & \No & Mixed & \No & \ghlink{https://github.com/KlingAIResearch/GameFactory}\\
\cellcolor{white} & GameNGen~\cite{GameNGen} & 8,912 & \Image\hspace{-0.4em}\Video & Partially & \No & Mixed & \Yes & \ghlink{https://github.com/GameNGen/GameNGen.github.io}\\
%\cellcolor{white} & DeepPHY~\cite{DeepPHY} & 1,336 & \Image & Partially & \No & Discrete & \Yes & \ghlink{https://github.com/ADAPT-Chase/deepphy}\\
%\cellcolor{white} & GameDevBench~\cite{GameDevBench} & 132 & \Text\hspace{-0.4em}\Image\hspace{-0.4em}\Video & Partially & \No & Discrete & \Yes & \ghlink{https://github.com/waynchi/gamedevbench}\\
%\cellcolor{white} & CivRealm~\cite{CivRealm} & 100,000 & \Text & Partially & \Yes & Discrete & \Yes & \ghlink{https://github.com/bigai-ai/civrealm}\\
% \cellcolor{white}  & Novel Games~\cite{NovelGames} & -- & \Text\hspace{-0.4em}\Image & Partially & \No & Discrete & \Yes & --\\ 

\cellcolor{white} & OGC~\cite{OGC} & 37 & \Image & Partially & \Yes & Discrete & \Yes & \ghlink{https://git.hcics.simtech.uni-stuttgart.de/public-projects/OGC}\\
%\cellcolor{white} & Collab-Overcooked~\cite{Collab_Overcooked} & 30 & \Text & Partially & \Yes & Discrete & \Yes & \ghlink{https://github.com/YusaeMeow/Collab-Overcooked}\\
% \cellcolor{white} &ReCon~\cite{ReCon} & 6 & \Text & Partially & \Yes & Discrete & \Yes & \ghlink{https://github.com/shenzhi-wang/avalon_recon}\\
\cellcolor{white} \multirow{-27}{*}{\makecell{\textbf{Game} \\ (\cref{Game})}} & Minigrid \& Miniworld~\cite{Minigrid_Miniworld} & 35 & \Text\hspace{-0.4em}\Image & Partially & \No & Discrete & \Yes & \ghlink{https://github.com/Farama-Foundation/Minigrid}\\

\bottomrule
\end{tabular}}
\end{table*}

\subsection{Embodied}
\label{Embodied}
\textbf{Embodied} environments refer to settings in which an agent acts as a robot or virtual character situated in a 3D environment. In these settings, the agent completes tasks through perception, movement, and interaction~\cite{Habitat,Room-to-Room,MetaDrive,RLBench,Robocasa,BEHAVIOR,ALFRED,ALFWorld,TEAch,ReALFRED}. According to the primary task forms and capability requirements, existing work can be broadly categorized into three types: \textbf{Spatial Navigation} (\cref{Spatial Navigation}), \textbf{Physical Manipulation} (\cref{Physical Manipulation}), and \textbf{Long-Horizon Planning} (\cref{Long-Horizon Planning}).

\subsubsection{Spatial Navigation}
\label{Spatial Navigation}
\textbf{Spatial Navigation} environments center on tasks that require agents to move, explore, localize, and reach target positions in 3D spaces. This line of research mainly examines whether an agent can build usable spatial representations from egocentric observations and act reliably in unseen environments. 
Habitat~\cite{Habitat} provides efficient photorealistic 3D simulation and supports standard embodied tasks such as point goal navigation. Room-to-Room~\cite{Room-to-Room} further extends this setting to language guided navigation in real building scale environments, highlighting the need to ground natural language instructions in perception and action. MetaDrive~\cite{MetaDrive} broadens navigation oriented embodied evaluation to driving scenarios by generating diverse road environments and testing generalization under dynamic traffic conditions. Spatial navigation has become a core setting for evaluating spatial perception and generalization in embodied agents.

\subsubsection{Physical Manipulation}
\label{Physical Manipulation}
\textbf{Physical Manipulation} environments focus on embodied interaction between agents and objects, including tasks such as grasping, placing, assembling and tool use. Compared with navigation tasks, this setting places greater emphasis on precise control, complex contact-rich interaction, and reasoning about object states. RLBench~\cite{RLBench} provides a large scale manipulation benchmark with diverse vision-guided tasks and supports research on multi-task and imitation-based robot learning. Robocasa~\cite{Robocasa} extends this direction toward realistic everyday household tasks and language-conditioned mobile manipulation. BEHAVIOR~\cite{BEHAVIOR} further improves realism by modeling a wide range of everyday household activities, while also showing that current embodied agents still exhibit clear limitations in complex physical interaction. Physical manipulation has expanded from narrow skill evaluation to more general and realistic object-centric interaction settings.

\subsubsection{Long-Horizon Planning}
\label{Long-Horizon Planning}
\textbf{Long-Horizon Planning} environments center on multi step and compositional tasks whose success depends on maintaining task state, decomposing goals, and adapting to changing environments. In this setting, the main challenge is not only to execute individual actions, but also to organize them into coherent plans over long decision horizons.
ALFRED~\cite{ALFRED} is a representative benchmark in this area, as it combines egocentric vision and natural language instructions with long household tasks that involve irreversible state changes. ALFWorld~\cite{ALFWorld} complements this direction by aligning text based and embodied environments, making it possible to study abstract planning together with grounded execution. TEACh~\cite{TEAch} further extends long-horizon embodied tasks to dialogue-enabled settings, where agents must complete household goals through both interaction and clarification. ReALFRED~\cite{ReALFRED} pushes this line toward more realistic environments by reducing the visual and scene gap between simulation and real-world deployment. Long-horizon planning has become an important setting for evaluating planning consistency and long-term task execution in embodied agents.

\begin{table*}[tbp]
\caption{An overview of various environments categorized into \textbf{Tool} and \textbf{Code} domains.}
\resizebox{\textwidth}{!}{
\rowcolors{2}{white}{gray!10} 
\renewcommand{\arraystretch}{1.3}
\begin{tabular}{c|lccccccc} \toprule
\textbf{Domain} & \textbf{Name} & \textbf{Size}  & \textbf{Modality} & \textbf{Observability} & \textbf{Multi-agent} & \textbf{Continuity} & \textbf{Online} & \textbf{Resource} \\ \midrule

\cellcolor{white} & API-Bank~\cite{API-Bank} & 314 & \Text & Completely & \No & Mixed & \Yes & \ghlink{https://github.com/AlibabaResearch/DAMO-ConvAI/tree/main/api-bank}\\
\cellcolor{white} & ToolBench~\cite{ToolBench} & 600 & \Text & Completely & \No & Mixed & \Yes & \ghlink{https://github.com/OpenBMB/ToolBench}\\
\cellcolor{white} & ToolEyes~\cite{ToolEyes} & 382 & \Text & Partially & \No & Discrete & \Yes & \ghlink{https://github.com/Junjie-Ye/ToolEyes}\\
\cellcolor{white} & AppWorld~\cite{AppWorld2024} & 585 & \Text & Partially & \No & Discrete & \Yes & \ghlink{https://github.com/stonybrooknlp/appworld}\\
\cellcolor{white} & $\tau$-bench~\cite{tau-bench} & 165 & \Text & Partially & \No & Mixed & \Yes & \ghlink{https://github.com/sierra-research/tau-bench}\\
\cellcolor{white} & ACEBench~\cite{ACEBench} & 2,000 & \Text & Partially & \No & Discrete & \Yes & \ghlink{https://github.com/chenchen0103/ACEBench}\\
\cellcolor{white} & FlowBench~\cite{FlowBench} & 536 & \Text & Partially & \No & Discrete & \Yes & \ghlink{https://github.com/Justherozen/FlowBench}\\
\cellcolor{white} & TRAJECT-Bench~\cite{TRAJECT-Bench} & 200 & \Text & Partially & \No & Discrete & \No & \hflink{https://huggingface.co/datasets/bigboss24/TRAJECT-Bench}\\
\cellcolor{white} & ETAPP~\cite{ETAPP} & 800 & \Text & Partially & \No & Mixed & \Yes & \ghlink{https://github.com/hypasd-art/ETAPP}\\
\cellcolor{white} & BFCL~\cite{BFCL} & 5,551 & \Text & Partially & \No & Discrete & \Yes & \ghlink{https://github.com/ShishirPatil/gorilla/tree/main}\\
\cellcolor{white} & UserBench~\cite{UserBench} & 471 & \Text & Partially & \No & Discrete & \Yes & 
\ghlink{https://github.com/SalesforceAIResearch/UserBench/tree/main}\\
\cellcolor{white} & $\tau$2-bench~\cite{tau2-bench} & 279 & \Text & Partially & \Yes & Discrete & \Yes & \ghlink{https://github.com/sierra-research/tau2-bench}\\
\cellcolor{white} & MCPVerse~\cite{MCPVerse} & 250 & \Text & Partially & \No & Mixed & \Yes & \ghlink{https://github.com/hailsham/mcpverse}\\
\cellcolor{white} & MCPToolBench++~\cite{MCPToolBench++} & 1,509 & \Text & Partially & \No & Discrete & \Yes & \ghlink{https://github.com/mcp-tool-bench/MCPToolBenchPP}\\
\cellcolor{white} & MCP-Universe~\cite{MCP-Universe} & 231 & \Text & Partially & \No & Mixed & \Yes & \hflink{https://huggingface.co/datasets/AfterQuery/MCP-Universe}\\
\cellcolor{white} & MCP-Bench~\cite{MCP-Bench} & 104 & \Text & Partially & \No & Discrete & \Yes & \ghlink{https://github.com/Accenture/mcp-bench}\\
\cellcolor{white} & MCPMark~\cite{MCPMark} & 127 & \Text & Partially & \No & Mixed & \Yes & \ghlink{https://github.com/eval-sys/mcpmark}\\
\cellcolor{white} \multirow{-18}{*}{\makecell{\textbf{Tool} \\ (\cref{Tool})}} & M$^3$-Bench~\cite{M^3-Bench}
& 208 & \Text\hspace{-0.4em}\Image & Partially & \No & Discrete & \Yes & \ghlink{https://github.com/EtaYang10th/Open-M3-Bench}\\

\midrule

\cellcolor{white} & SWE-Bench~\cite{SWE-Bench} & 2,294 & \Text & Partially & \No & Discrete & \Yes & \raisebox{-0.2em}{\hflink{https://huggingface.co/datasets/SWE-bench/SWE-bench}}\\
\cellcolor{white} & InterCode~\cite{InterCode} & 1,351 & \Text & Partially & \No & Discrete & \Yes & \ghlink{https://github.com/princeton-nlp/intercode}\\
\cellcolor{white} & CodeAgent~\cite{CodeAgent} & 101 & \Text & Partially & \No & Discrete & \Yes & \ghlink{https://github.com/zkcpku/CodeAgent}\\
\cellcolor{white} & BigCodeBench~\cite{BigCodeBench} & 1,140 & \Text & Completely & \No & Discrete & \Yes & \raisebox{-0.2em}{\hflink{https://huggingface.co/datasets/bigcode/bigcodebench}}\\
\cellcolor{white} & CodeElo~\cite{CODEELO} & 387 & \Text & Completely & \No & Discrete & \Yes & \raisebox{-0.2em}{\hflink{https://huggingface.co/datasets/Qwen/CodeElo}}\\
\cellcolor{white} & LiveCodeBench~\cite{LiveCodeBench} & 511 & \Text & Completely & \No & Discrete & \Yes & \raisebox{-0.2em}{\hflink{https://huggingface.co/datasets/livecodebench/code_generation}}\\
\cellcolor{white} & CSR-Bench~\cite{CSR-Bench} & 100 & \Text & Partially & \No & Discrete & \Yes & \ghlink{https://github.com/amazon-science/CSR-Bench}\\
\cellcolor{white} & SWE-Bench Pro~\cite{SWE-Bench_Pro} & 1,865 & \Text & Partially & \No & Discrete & \Yes & \raisebox{-0.2em}{\hflink{https://huggingface.co/datasets/ScaleAI/SWE-bench_Pro}}\\
\cellcolor{white} & Terminal-Bench~\cite{Terminal-Bench} & 89 & \Text & Partially & \No & Discrete & \Yes & \ghlink{https://github.com/harbor-framework/terminal-bench}\\
\cellcolor{white} & SWE-bench Multimodal~\cite{SWE-bench_Multimodal} & 517 & \Text\hspace{-0.4em}\Image & Partially & \No & Discrete & \Yes & \raisebox{-0.2em}{\hflink{https://huggingface.co/datasets/SWE-bench/SWE-bench_Multimodal}}\\
\cellcolor{white} & KernelBench~\cite{KernelBench} & 250 & \Text & Completely & \No & Discrete & \Yes & \raisebox{-0.2em}{\hflink{https://huggingface.co/datasets/ScalingIntelligence/KernelBench}}\\
\cellcolor{white} & IDE-Bench~\cite{IDE-Bench} & 6,000 & \Text & Partially & \No & Discrete & \Yes & \ghlink{https://github.com/AfterQuery/IDE-Bench}\\
\cellcolor{white}  & MBPP~\cite{MBPP} & 2,383 & \Text & Completely & \No & Discrete & \Yes & \raisebox{-0.2em}{\hflink{https://huggingface.co/datasets/Muennighoff/mbpp}}\\ 
\cellcolor{white} & CRUXEval~\cite{CRUXEval} & 800 & \Text & Completely & \No & Discrete & \Yes & \ghlink{https://github.com/facebookresearch/cruxeval}\\
\cellcolor{white} & DebugBench~\cite{DebugBench} & 4,253 & \Text & Completely & \No & Discrete & \Yes & \ghlink{https://github.com/thunlp/DebugBench}\\
\cellcolor{white} & CodeRAG-Bench~\cite{CodeRAG_Bench} & 9,000 & \Text & Partially & \No & Discrete & \Yes & \ghlink{https://github.com/code-rag-bench/code-rag-bench}\\
\cellcolor{white} & SWT-Bench~\cite{SWT_Bench} & 1,983 & \Text & Partially & \No & Discrete & \Yes & \ghlink{https://github.com/logic-star-ai/swt-bench}\\
\cellcolor{white} & FEA-Bench~\cite{FEA_Bench} & 1,401 & \Text & Partially & \No & Discrete & \Yes & \ghlink{https://github.com/microsoft/FEA-Bench}\\
\cellcolor{white} & Multi-SWE-bench~\cite{Multi_SWE_bench} & 1,632 & \Text & Partially & \No & Discrete & \Yes & \ghlink{https://github.com/multi-swe-bench/multi-swe-bench}\\
\cellcolor{white} \multirow{-21}{*}{\makecell{\textbf{Code} \\ (\cref{Code})}} & NL2Repo-Bench~\cite{NL2Repo_Bench} & 104 & \Text & Partially & \No & Discrete & \Yes & \ghlink{https://github.com/multimodal-art-projection/NL2RepoBench}\\

\bottomrule
\end{tabular}
}
\vspace{-3pt}
\end{table*}

\subsection{Game}
\label{Game}
\textbf{Game} refers to scenarios in which an agent operates in a game world with explicit rules~\cite{BabaIsAI,GameTraversalBenchmark,SMARTPLAY}, goal constraints~\cite{MineDojo,MC-Planner,MINEWORLD}, and evolving states, continuously perceiving the environment, selecting actions, and adjusting its strategy based on feedback to accomplish interactive decision tasks. Based on differences in game mechanics and primary capabilities, existing work can be broadly categorized into five types: \textbf{Open World Games} (\cref{Open World Games}), \textbf{Puzzle Reasoning Games} (\cref{Puzzle Reasoning Games}), \textbf{Social Deduction Games} (\cref{Social Deduction Games}), \textbf{Adventure Quest Games} (\cref{Adventure Quest Games}), and \textbf{Strategy Management Games} (\cref{Strategy Management Games}).

\subsubsection{Open World Games} 
\label{Open World Games}
\textbf{Open World Games} examine whether an agent can sustain exploration and accomplish goals in highly open worlds with large state spaces and sparse feedback. MineDojo~\cite{MineDojo} is a representative benchmark in this direction, building a large suite of open world tasks on top of Minecraft that covers exploration, collection, crafting, and survival. 
MC Planner~\cite{MC-Planner} further emphasizes multiple steps planning in Minecraft, highlighting the need for more detailed subgoal ordering and feedback correction in long-horizon tasks. MineWorld~\cite{MINEWORLD} shows that open world games can further evolve into persistent real time environments, where agents are no longer evaluated only on fixed task sets but instead make long term decisions in more continuous open worlds. Overall, open world games provide a testbed that more closely resembles complex real world environments.

\subsubsection{Puzzle Reasoning Games}
\label{Puzzle Reasoning Games}
\textbf{Puzzle Reasoning Games} focus on whether an agent can reason effectively in tasks with constrained rules but high logical demands.
Baba Is AI~\cite{BabaIsAI} is especially suited for evaluating rule rewriting and compositional reasoning. Agents must manipulate not only objects but also the rules themselves. GameTraversalBenchmark~\cite{GameTraversalBenchmark} abstracts planning as a two dimensional grid traversal problem, emphasizing whether an agent can reach the goal in as few steps as possible, and is therefore well suited for assessing path planning and state tracking. 
SmartPlay~\cite{SMARTPLAY} decomposes multiple capabilities through mini games, enabling finer evaluation of abilities such as object based reasoning, spatial reasoning, and planning ahead. Overall, these environments make reasoning weaknesses easier to identify by placing complex capabilities within clearer rule structures.

\subsubsection{Social Deduction Games}
\label{Social Deduction Games}
\textbf{Social Deduction Games} focus on communication, strategic interaction, and identity inference among multiple agents under incomplete information. 
AvalonBench~\cite{AvalonBench} uses hidden roles, negotiation, and deception to evaluate multi agent language gaming abilities. 
Werewolf~\cite{Werewolf} further illustrates the importance of reasoning in environments where hidden roles and public discussion coexist. These social game settings require both logical analysis and effective dialogue generation.
WhodunitBench~\cite{WhodunitBench} extends this line to multimodal murder mystery games, unifying perception, interaction, and cognition in one dynamic environment. In addition, more general competitive text game platforms such as TextArena~\cite{TextArena} make it possible to compare negotiation, theory of mind, and deception capabilities across a broader range of games in a unified way. Overall, these environments make games an important testbed for collective reasoning and social interaction.

\subsubsection{Adventure Quest Games}
\label{Adventure Quest Games}
\textbf{Adventure Quest Games} focus on agents completing full task chains in scenarios involving story progression, clue collection, and multi stage objectives. 
FlashAdventure~\cite{FlashAdventure} directly evaluates whether an agent can complete an entire storyline, emphasizing that it must not only perform isolated actions but also sustain progress throughout the whole adventure progression. BALROG~\cite{BALROG} takes a broader view of long-horizon interactive game tasks, integrating spatial reasoning, long term planning, and continued exploration across a range of challenging games. GameArena~\cite{GameArena} further shows that interactive games can serve not only as task environments but also as dynamic evaluation platforms, capturing models’ step by step reasoning during real gameplay. 
Overall, this direction shifts evaluation from one-shot problem solving to continuous task completion, thereby placing greater emphasis on capabilities such as retaining information, integrating clues, and planning over multiple stages.

\subsubsection{Strategy Management Games}
\label{Strategy Management Games}
\textbf{Strategy Management Games} focus on high level decision making under resource constraints, long term returns, and complex competition. CivRealm~\cite{CivRealm} is built on a Civilization style environment. It unifies resource management, diplomacy, warfare, and long-term societal development within a single platform. This makes it well suited for studying macro level decision making. The Factorio Learning Environment~\cite{FactorioLearningEnvironment} emphasizes long term planning in open factory systems through automation, resource optimization, and scalable production chains. 
While PokeLLMon~\cite{PokeLLMon} is set in a more strongly competitive tactical game scenario, it likewise highlights knowledge use, online feedback, and long term decision adjustment. Overall, these environments extend games from local rule solving to high level platforms for global utility optimization.

\subsection{Tool}
\label{Tool}
\textbf{Tool Use} refers to settings in which agents invoke external tools, such as functions, APIs, databases, or execution engines, to acquire information or execute actions to complete the task~\cite{API-Bank,BFCL,ACEBench,ToolBench,tau-bench,UserBench,tau2-bench,MCPVerse,MCPToolBench++,MCP-Bench}. Unlike GUI environments, tool use does not require visual interface manipulation. Instead, the interaction proceeds through structured tool calls and their corresponding outputs. According to whether tool use involves simulated users or is built upon MCP servers, existing work can be broadly categorized into three types: \textbf{Conventional Tool Use} (\cref{Conventional Tool Use}), \textbf{User-Simulated Tool Use} (\cref{User-Simulated Tool Use}), and \textbf{MCP-based Tool Use} (\cref{MCP-based Tool Use}).

\subsubsection{Conventional Tool Use}
\label{Conventional Tool Use}
\textbf{Conventional Tool Use} focuses on settings where agents invoke external tools without interacting with explicit simulated users. In this setting, the main difficulty lies in selecting the correct tool, producing valid arguments, and executing tool calls in a reliable manner. API-Bank~\cite{API-Bank} evaluates whether models can select appropriate APIs and invoke them correctly under a fixed tool inventory. BFCL~\cite{BFCL} further strengthens this setting by introducing a more systematic evaluation of function calls, covering serial and parallel calls, as well as structured verification of output correctness.  %ACEBench~\cite{ACEBench} extends this line of evaluation to more realistic instructions and multi-turn interactions, while still preserving a bounded action space. %
ToolBench~\cite{ToolBench} expands the scale of tool space by exposing models to a large collection of real-world APIs and evaluating both single-tool and multi-tool problem solving. AppWorld~\cite{AppWorld2024} further advances tool-use evaluation toward executable app-based environments, where agents need to interact with multiple application APIs and complete tasks under state-based verification. In general, conventional tool use has become a core setting for evaluating whether agents can accurately use tools in standard tool call environments.

\subsubsection{User-Simulated Tool Use}
\label{User-Simulated Tool Use}
\textbf{User-Simulated Tool Use} studies settings in which tool use is embedded in interactions with simulated users. Compared with conventional tool use, these tasks place greater emphasis on responding to evolving user intentions, clarifying underspecified requests, and adjusting tool-use behavior according to conversational feedback. $\tau$-bench~\cite{tau-bench} highlights this setting by embedding tool use into dynamic user interactions, where agents must follow policies while updating their actions according to conversation progress. UserBench~\cite{UserBench} further pushes this setting toward more user-centered workflows, as agents are required to clarify underspecified goals and refine their decisions as user preferences are gradually revealed. $\tau^2$-bench~\cite{tau2-bench} further extends this paradigm by introducing a dual-control interactive environment, where both agents and simulated users can influence the environment through tool use, requiring more adaptive and coordinated decision-making. Overall, this class of benchmarks shifts evaluation from isolated tool use to interactive, user-centered settings, emphasizing adaptability and alignment with user intent.

\subsubsection{MCP-based Tool Use}
\label{MCP-based Tool Use}
\textbf{MCP-based Tool Use} concerns settings in which agents interact with tools provided through MCP servers. The central challenge in this setting is not only correct tool invocation, but also effective tool discovery, task decomposition, and coordination across a large collection of MCP tools. MCPVerse~\cite{MCPVerse} significantly expands the scale of tool space by incorporating hundreds of executable tools, making tool retrieval and orchestration substantially more demanding. MCPToolBench++~\cite{MCPToolBench++} further evaluates model performance in MCP-based environments with diverse tools and execution settings. MCP-Bench~\cite{MCP-Bench} emphasizes realistic multi-step tasks over live MCP servers, requiring agents to identify relevant tools, plan execution sequences, and integrate outputs across different sources. Overall, these works extend tool-use evaluation from general tool calling to standardized MCP-based tool ecosystems.

\begin{table*}
\caption{An overview of various environments categorized into \textbf{Domain-Specific} and \textbf{Cross-Domain}.}
\resizebox{\textwidth}{!}{
\rowcolors{2}{white}{gray!10} 
\renewcommand{\arraystretch}{1.3}
\begin{tabular}{c|lccccccc} \toprule
\textbf{Domain} & \textbf{Name} & \textbf{Size}  & \textbf{Modality} & \textbf{Observability} & \textbf{Multi-agent} & \textbf{Continuity} & \textbf{Online} & \textbf{Resource} \\ \midrule

\cellcolor{white} & Biocoder~\cite{Biocoder} & 460 & \Text & Completely & \No & Discrete & \Yes & \ghlink{https://github.com/gersteinlab/biocoder}\\
\cellcolor{white} & TravelPlanner~\cite{TravelPlanner} & 1,225 & \Text & Partially & \No & Discrete & \Yes & \hflink{https://huggingface.co/datasets/osunlp/TravelPlanner}\\
\cellcolor{white} & DSEval~\cite{DSEval} & 825 & \Text & Partially & \No & Discrete & \Yes & \ghlink{https://github.com/MetaCopilot/dseval}\\
\cellcolor{white} & Natural Plan~\cite{Natural_Plan} & 3,600 & \Text & Completely & \No & Discrete & \No & \ghlink{https://github.com/google-deepmind/natural-plan}\\
\cellcolor{white} & ScienceAgentBench~\cite{ScienceAgentBench} & 102 & \Text & Completely & \No & Discrete & \Yes & \hflink{https://huggingface.co/datasets/osunlp/ScienceAgentBench}\\
\cellcolor{white} & DSBench~\cite{DSBench} & 540 & \Text\hspace{-0.4em}\Image & Partially & \No & Discrete & \Yes & \hflink{https://huggingface.co/datasets/liqiang888/DSBench}\\
\cellcolor{white} & MLE-bench~\cite{MLE-bench} & 75 & \Text\hspace{-0.4em}\Image\hspace{-0.4em}\Video & Partially & \No & Discrete & \Yes & \ghlink{https://github.com/openai/mle-bench}\\
\cellcolor{white} & DiscoveryWorld~\cite{DiscoveryWorld} & 120 & \Text\hspace{-0.4em}\Image & Partially & \No & Discrete & \Yes & \ghlink{https://github.com/allenai/discoveryworld}\\
\cellcolor{white} & MedAgentBench~\cite{MedAgentBench} & 300 & \Text & Partially & \No & Discrete & \Yes & \hflink{https://huggingface.co/datasets/shaafsalman/MedAgentBench-Tasks}\\
\cellcolor{white} & StockBench~\cite{StockBench} & 82 & \Text & Partially & \No & Mixed & \Yes & \ghlink{https://github.com/ChenYXxxx/stockbench}\\
\cellcolor{white} & MLE-Dojo~\cite{MLE-Dojo} & 50 & \Text & Partially & \No & Discrete & \Yes & \ghlink{https://github.com/MLE-Dojo/MLE-Dojo}\\
\cellcolor{white} & FinDeepResearch~\cite{FinDeepResearch} & 64 & \Text & Partially & \No & Mixed & \Yes & \hflink{https://huggingface.co/datasets/OpenFinArena/FinDeepResearch}\\
\cellcolor{white} & PaperArena~\cite{PaperArena} & 784 & \Text\hspace{-0.4em}\Image & Partially & \No & Discrete & \Yes & \ghlink{https://github.com/Melmaphother/PaperArena}\\
\cellcolor{white} & BixBench~\cite{BixBench} & 205 & \Text\hspace{-0.4em}\Image & Partially & \No & Discrete & \Yes & \raisebox{-0.2em}{\hflink{https://huggingface.co/datasets/futurehouse/BixBench}}\\
\cellcolor{white} & CRMArena-Pro~\cite{CRMArena-Pro} & 8,614 & \Text & Partially & \No & Discrete & \Yes & \raisebox{-0.2em}{\hflink{https://huggingface.co/datasets/Salesforce/CRMArenaPro}}\\
\cellcolor{white} & MedAgentGym~\cite{MedAgentGym} & 13,238 & \Text & Partially & \No & Discrete & \Yes & \raisebox{-0.2em}{\ghlink{https://github.com/wshi83/MedAgentGym}}\\
\cellcolor{white} & Finance Agent Benchmark~\cite{Finance_Agent_Benchmark} & 337 & \Text & Partially & \No & Discrete & \Yes & \raisebox{-0.2em}{\hflink{https://huggingface.co/datasets/vals-ai/finance_agent_benchmark}}\\
\cellcolor{white} & EcomBench~\cite{EcomBench} & 100 & \Text & Partially & \No & Discrete & \No & \raisebox{-0.2em}{\hflink{https://huggingface.co/datasets/Alibaba-NLP/EcomBench}}\\
\cellcolor{white} & MedAgentBench v2~\cite{MedAgentBench_v2} & 300 & \Text & Partially & \No & Discrete & \Yes & \ghlink{https://github.com/ericoericochen/medagentbenchv2}\\
\cellcolor{white} & MedMCP-Calc~\cite{MedMCP-Calc} & 118 & \Text & Partially & \No & Mixed & \Yes & \ghlink{https://github.com/SPIRAL-MED/MedMCP-Calc}\\
\cellcolor{white} & ESG~\cite{ESG} & 291 & \Text\hspace{-0.4em}\Video & Partially & \Yes & Discrete & \Yes & \ghlink{https://github.com/ElaineZhao92/ESGAgent-and-Benchmark}\\
\cellcolor{white} & BioAgent Bench~\cite{BioAgent_Bench} & -- & \Text & Partially & \No & Discrete & \Yes & \ghlink{https://github.com/bioagent-bench/bioagent-bench}\\
\cellcolor{white} & WoW-bench~\cite{WoW-bench} & 234 & \Text & Partially & \No & Discrete & \Yes & \ghlink{https://github.com/Skyfall-Research/world-of-workflows}\\
\cellcolor{white} & EnterpriseOps-Gym~\cite{EnterpriseOps-Gym} & 1,150 & \Text & Partially & \No & Discrete & \Yes & \hflink{https://huggingface.co/datasets/ServiceNow-AI/EnterpriseOps-Gym}\\
\cellcolor{white} & MetaClaw~\cite{MetaClaw} & 934 & \Text & Partially & \No & Discrete & \Yes & \ghlink{https://github.com/aiming-lab/MetaClaw}\\
\cellcolor{white} & Claw-Eval~\cite{Claw-Eval} & 300 & \Text\hspace{-0.4em}\Image\hspace{-0.4em}\Video & Partially & \No & Discrete & \Yes & \hflink{https://huggingface.co/datasets/claw-eval/Claw-Eval}\\
\cellcolor{white} \multirow{-26}{*}{\makecell{\textbf{Domain-Specific}\\ (\cref{Domain-Specific})}} & ClawArena~\cite{ClawArena} & 1,879 & \Text & Partially & \No & Discrete & \Yes & \ghlink{https://github.com/aiming-lab/ClawArena}\\
\midrule

\cellcolor{white} & OpenAI Gym~\cite{Openai_gym} & -- & \Text\hspace{-0.4em}\Image & Partially & \No & Mixed & \Yes & \ghlink{https://github.com/openai/gym}\\
\cellcolor{white} & HuggingGPT~\cite{HuggingGPT} & 3,497 & \Text\hspace{-0.4em}\Image & Completely & \No & Discrete & \Yes & \ghlink{https://github.com/microsoft/JARVIS}\\
\cellcolor{white} & AgentBench~\cite{AgentBench} & 1,014 & \Text & Partially & \No & Mixed & \Yes & \ghlink{https://github.com/THUDM/AgentBench}\\
\cellcolor{white} & TaskLAMA~\cite{TaskLAMA} & 478 & \Text & Completely & \No & Discrete & \Yes & \hflink{https://huggingface.co/datasets/Spico/TaskLAMA}\\
\cellcolor{white} & TaskBench~\cite{TaskBench} & 17,331 & \Text\hspace{-0.4em}\Image\hspace{-0.4em}\Video & Completely & \No & Discrete & \No & \ghlink{https://github.com/microsoft/JARVIS/tree/main/taskbench}\\
\cellcolor{white} & AgentBoard~\cite{AgentBoard} & 1,012 & \Text & Partially & \No & Discrete & \Yes & \hflink{https://huggingface.co/datasets/hkust-nlp/agentboard}\\
\cellcolor{white} & AgentGym~\cite{AgentGym} & 20,509 & \Text & Partially & \No & Discrete & \Yes & \ghlink{https://github.com/WooooDyy/AgentGym}\\
\cellcolor{white} & WorfBench~\cite{WorfBench} & 2,146 & \Text & Completely & \No & Discrete & \No & \ghlink{https://github.com/zjunlp/WorfBench}\\
\cellcolor{white} & GEM~\cite{GEM} & -- & \Text\hspace{-0.4em}\Image & Partially & \No & Discrete & \Yes & \ghlink{https://github.com/axon-rl/gem}\\
\cellcolor{white} & MLGym~\cite{MLGym} & -- & \Text & Partially & \No & Discrete & \Yes & \ghlink{https://github.com/facebookresearch/MLGym}\\
\cellcolor{white} & MedBrowseComp~\cite{MedBrowseComp} & 1,139 & \Text & Partially & \No & Mixed & \Yes & \hflink{https://huggingface.co/datasets/AIM-Harvard/MedBrowseComp}\\
\cellcolor{white} & WebMMU~\cite{WebMMU} & 4,240 & \Text\hspace{-0.4em}\Image & Completely & \No & Discrete & \No & \hflink{https://huggingface.co/datasets/McGill-NLP/WebMMU}\\
\cellcolor{white} & TPS-Bench~\cite{TPS-Bench} & 200 & \Text & Partially & \No & Discrete & \Yes & \ghlink{https://github.com/hanwenxu1/mcp-agent}\\
\cellcolor{white} & AutoEnv~\cite{AutoEnv} & -- & \Text & Partially & \No & Discrete & \Yes & \ghlink{https://github.com/FoundationAgents/AutoEnv}\\
\cellcolor{white} & AgencyBench~\cite{AgencyBench} & 138 & \Text & Partially & \No & Discrete & \Yes & \hflink{https://huggingface.co/datasets/GAIR/AgencyBench}\\
\cellcolor{white} \multirow{-17}{*}{\makecell{ \textbf{Cross-Domain} \\ (\cref{Cross-Domain})}}  & AgentVista~\cite{AgentVista} & 209 & \Text & Partially & \No & Discrete & \Yes & \hflink{https://huggingface.co/datasets/Warrieryes/AgentVista}\\
\bottomrule
\end{tabular}}
\end{table*}

\subsection{Code}
\label{Code}
\textbf{Code} refers to scenarios in which an agent interacts continuously with code, repositories~\cite{CodeAgent,CSR-Bench,SWE-bench_Multimodal}, test harnesses~\cite{InterCode,CSR-Bench}, and other related programming artifacts in order to accomplish a programming objective. Existing work covers different stages of the software development process, and the boundaries between benchmarks are often not strict. Based on the primary bottlenecks involved, they can be broadly categorized into four types: \textbf{Code Generation} (\cref{Code Generation}), \textbf{Code Understanding} (\cref{Code Understanding}), \textbf{Code Verification} (\cref{Code Verification}), and \textbf{Code Debugging} (\cref{Code Debugging}).

\subsubsection{Code Generation}
\label{Code Generation}
\textbf{Code Generation} focuses on producing new code from natural language requirements or problem descriptions. MBPP~\cite{MBPP} is an early benchmark for function level program synthesis and serves as a basic reference for code generation ability. BigCodeBench~\cite{BigCodeBench} increases task complexity by moving beyond short algorithmic problems to emphasize complex instruction understanding and multi-function calling, making evaluation closer to real development needs. CodeAgent~\cite{CodeAgent} further extends this setting to repository scale code generation, where models must generate solutions under complex project structures and dependencies. LiveCodeBench~\cite{LiveCodeBench} highlights continual updates and contamination control, helping distinguish whether models solve tasks through memorization or genuine coding ability. 
Overall, this direction has evolved from simple static code generation to more complex, dynamic, and realistic settings.

\subsubsection{Code Understanding}
\label{Code Understanding}
\textbf{Code Understanding} focuses on whether agents can understand repository structure, interface relationships, dependency paths, and task context. 
NL2Repo-bench~\cite{NL2Repo_Bench} evaluates whether models can understand natural language requirements, locate relevant files and modules, and reason over project structure rather than isolated code snippets.
CSR-Bench~\cite{CSR-Bench} highlights repository understanding in deployment tasks by requiring agents to read documentation, analyze directory structures, and generate executable commands. SWE-bench-Multimodal~\cite{SWE-bench_Multimodal} further shows that software context may also include interfaces, screenshots, and other visual information. 
Overall, code understanding is expanding from reading functions to interpreting repositories, documentation, and multimodal software context.

\subsubsection{Code Verification}
\label{Code Verification}
\textbf{Code Verification} focuses on how agents use tests, compilation, execution results, and performance metrics to verify whether a candidate solution is correct. 
MBPP~\cite{MBPP} and BigCodeBench~\cite{BigCodeBench} rely on executable tests for stable requirement checking. 
LiveCodeBench~\cite{LiveCodeBench} further integrates execution, self-repair, and test evaluation in a dynamic setting, making verification part of the full coding process rather than just a final scoring step. 
KernelBench~\cite{KernelBench} extends this idea to performance oriented scenarios by evaluating both correctness and efficiency gains. 
Overall, code verification provides a valuable closed loop through objective execution feedback.

\subsubsection{Code Debugging}
\label{Code Debugging}
\textbf{Code Debugging} involves locating and fixing problems based on error messages, test failures, execution anomalies, and environmental feedback.
InterCode~\cite{InterCode} makes debugging an interactive repair process through execution feedback. 
SWE-Bench~\cite{SWE-Bench} requires root cause analysis and multi-step fixes in real software contexts. 
CSR-Bench~\cite{CSR-Bench} further extends debugging to repository deployment scenarios involving dependencies, scripts, and experimental environments. This shows that \textbf{Code Debugging} has expanded from single file bug fixing to system level and environment level problem solving.

\subsection{Domain-Specific}
\label{Domain-Specific}
\textbf{Domain-specific} environments evaluate whether agents can effectively operate in specialized disciplinary or industrial settings~\cite{MedAgentBench,MedAgentGym,BioAgent_Bench,Biocoder,DiscoveryWorld,ScienceAgentBench,MLE-bench,StockBench,FinDeepResearch,Finance_Agent_Benchmark}. They require agents to understand domain-specific knowledge and operate in accordance with professional practices. In this section, we broadly divide existing work into three categories: \textbf{Biomedicine and Healthcare} (\cref{Biomedical and Healthcare}), \textbf{Science and Technology} (\cref{Science and Technology}), and \textbf{Finance and Investment} (\cref{Finance}).

\subsubsection{Biomedical and Healthcare}
\label{Biomedical and Healthcare}
\textbf{Biomedical and Healthcare} environments evaluate whether agents can operate effectively in clinical and biomedical settings. Agents in these environments need to understand clinical terminology, biomedical data, and bioinformatics workflows, while acting reliably under professional constraints. MedAgentBench~\cite{MedAgentBench} studies agent behavior in clinical tasks derived from electronic health records and physician workflows. MedAgentGym~\cite{MedAgentGym} provides an executable environment for training and evaluating medical agents at scale. Beyond clinical settings, BioAgent Bench~\cite{BioAgent_Bench} focuses on bioinformatics workflows, while BioCoder~\cite{Biocoder} evaluates coding ability on biomedical programming tasks. 

\subsubsection{Science and Technology}
\label{Science and Technology}
\textbf{Science and Technology} environments study whether agents can support research activities such as literature understanding, experiment design, and result analysis. DiscoveryWorld~\cite{DiscoveryWorld} is a representative benchmark that places agents in interactive scientific tasks requiring multi-step reasoning and experimentation. ScienceAgentBench~\cite{ScienceAgentBench} evaluates agents on realistic research-oriented problems grounded in scientific practice. MLE-bench~\cite{MLE-bench} further extends this setting to machine learning research and engineering, where agents must handle data, models, and iterative experimentation.

\subsubsection{Finance and Investment}
\label{Finance}
\textbf{Finance and Investment} environments evaluate whether agents can analyze financial information and make decisions in dynamic and risk sensitive settings. StockBench~\cite{StockBench} focuses on stock trading tasks and tests whether agents can act on market signals, news, and financial indicators. FinDeepResearch~\cite{FinDeepResearch} studies deep financial analysis and emphasizes evidence collection and structured reasoning. Finance Agent Benchmark~\cite{Finance_Agent_Benchmark} further evaluates research oriented financial tasks built around realistic analyst style workflows.

\subsection{Cross-Domain}
\label{Cross-Domain}
\textbf{Cross-Domain} environments evaluate whether agents can generalize across heterogeneous tasks and settings, rather than operate only within a single domain. OpenAI Gym~\cite{Openai_gym} established a unified interface for benchmarking agent behavior across diverse environments. AgentBench~\cite{AgentBench} evaluates reasoning and decision making in multiple interactive environments under a common evaluation setting. AgentBoard~\cite{AgentBoard} further extends this direction by providing a unified analytical framework with finer-grained process-level metrics, making it easier to compare agent performance beyond final success rate. Recent work also begins to study broader transfer and learning across environments. GEM~\cite{GEM} provides a Gym-style framework for training and evaluating agentic language models in diverse environments, while AutoEnv~\cite{AutoEnv} moves a step further by automatically generating heterogeneous worlds for measuring cross-environment adaptation. Overall, this line of research shifts evaluation from single-domain competence to cross-environment robustness, transferability, and scalable learning.

\subsection{Summary}
\begin{tcolorbox}[takeaway,title={Takeaway 4}]
\begin{itemize}
    \item \textbf{Capability Focus Across Domains:} GUI, Tool, and Code domains stress grounded external operation and workflow execution; Embodied and Game domains emphasize perception, control, and planning; Deep Research and Domain-Specific domains require evidence synthesis, knowledge-grounded reasoning, and reliable decision making;
    while Cross-Domain settings evaluate transfer across heterogeneous environments.
    \item \textbf{Evolution of Environment Design:} Agent environments are evolving from static and narrow benchmarks toward executable, multimodal, and long-horizon settings. Future environment design should better balance realism, diversity, controllability and verifiability to support both robust training and reliable evaluation of general agents.
\end{itemize}
\end{tcolorbox}

\section{Environment Synthesis}\label{sec:env_syn}

\begin{figure*}[t]
    \centering
    \includegraphics[width=1\textwidth]{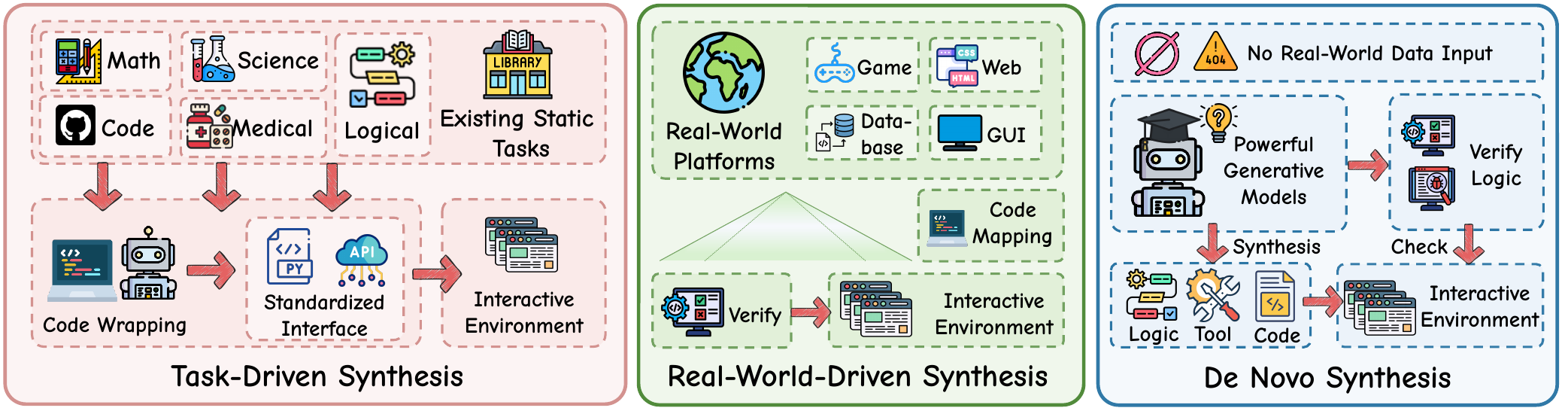 }
    \caption{
    Three symbolic environment synthesis methods are presented: Task-Driven Synthesis, Real-World-Driven Synthesis, and De Novo Synthesis. From left to right, the methods offer increasing degrees of freedom and require more verification logic.
    }
    \label{fig:symbolic_synthesis}
\end{figure*}

% \renewcommand{\arraystretch}{1.5} 
% --- END OF LINK ---
% \begin{table*}
% \resizebox{\textwidth}{!}{
% \begin{tabular}{lcccccc}
% \toprule
% \textbf{Name} &
%   \textbf{Modality} &
%  \textbf{Architecture} &   \textbf{Base Model} &
%   \textbf{Task Domain} &
%   \textbf{Data Source}  & \textbf{Application}\\ \midrule

% \rowcolor{gray!10!white} \multicolumn{7}{c}{\textbf{\textit{Task-Driven Synthesis}}}\\

% \rowcolor{gray!10!white} \multicolumn{7}{c}{\textbf{\textit{Real-World-Driven Synthesis}}}\\

% \rowcolor{gray!10!white} \multicolumn{7}{c}{\textbf{\textit{Synthesis from Scratch}}}\\
%  gg-bench\cite{gg-bench} & \Text & LLM & OpenAI o1 & Game & - & Evaluation \\
%  AutoEnv\cite{AutoEnv} & \Text & LLM & Claude-4-Sonnet & Domain & - & Training,Evaluation \\
%  AutoForge\cite{AutoForge} & \Text & LLM & Qwen3-Thinking & Code, Tool & - & Training,Evaluation \\
%  SWE-Playground\cite{SWE-Playground} & \Text & LLM & Gemini 2.5, GPT-4.1... & Code & - & Training \\
%  Endless Terminals\cite{Endless_Terminals} & \Text & LLM & o3 & Code & - & Training,Evaluation \\
%  Agent World Model\cite{Agent_world_model} & \Text & LLM & GPT-5 & Tool & - & Training \\
%  ScaleEnv\cite{ScaleEnv} & \Text & LLM & Deepseek-V3.2, GPT-5.1... & Code & - & Training \\
%  LOGIGEN\cite{LOGIGEN} & \Text & LLM & DeepSeek-V3.2 & Domain-Specific & - & Training \\

% \bottomrule
% \end{tabular}}
% \end{table*}

Large-scale, high-quality training environments are crucial for improving the capabilities of agents. However, relying on manual construction severely limits the scalability of both environment quantity and diversity. Consequently, recent research increasingly explores automated approaches to synthesize training environments at scale. Based on the form of the resulting environments, we categorize these methods into \textbf{Symbolic Synthesis} (\cref{sec:Symbolic Synthesis}) and \textbf{Neural Synthesis} (\cref{sec:Neural Synthesis}). Symbolic synthesis constructs environments using symbolic rules such as code, while the environment in neural synthesis is represented by a neural model (e.g., a world model).

\subsection{Symbolic Synthesis}
\label{sec:Symbolic Synthesis}

Formally, a symbolic environment can be defined as a tuple $\mathcal{E} = \langle \mathcal{S}, \mathcal{A}, \mathcal{P}, \mathcal{R} \rangle$. Here $\mathcal{S}$ and $\mathcal{A}$ represent the state and action spaces. The core characteristic of symbolic synthesis is that the state transition ${\mathcal{P}}: \mathcal{S} \times \mathcal{A} \rightarrow \mathcal{S}$ and the feedback mechanism $\mathcal{R}$ are explicitly controlled by a set of symbolic rules or executable code $\mathcal{C}$. These environments are driven by underlying rules to ensure reliable feedback. To facilitate agent training, symbolic synthesis is widely used to obtain diverse, high-quality, and verifiable environments at scale \cite{DeepSeek-V3.2, LongCat-Flash-Thinking-2601, MiMo-V2-Flash}. As shown in Fig. \ref{fig:symbolic_synthesis}, based on the source of synthesis logic and the degree of design freedom, we categorize these methods into three types: \textbf{(1) Task-Driven Synthesis} (\cref{sec:task driven synthesis}), which transforms static datasets such as mathematics and coding problems into interactive environments through code encapsulation; \textbf{(2) Real-World-Driven Synthesis} (\cref{sec:real world driven synthesis}), which constructs simplified virtual mappings of real-world complex interactive environments like web pages and operating systems; \textbf{(3) De Novo Synthesis} (\cref{sec:synthesis from scratch}), which leverages generative models to autonomously build environments equipped with tool and verification logic without relying on predefined instances.

\subsubsection{Task-Driven Synthesis}
\label{sec:task driven synthesis}

Task-driven synthesis denotes transforming existing static tasks \cite{Generalized_Planning, BYTESIZED32, Exploration_Walk, Text2World, WorldCoder} and data such as tool-calling \cite{EnvScaler, AgentScaler, Agent_CPT} or mathematical data \cite{Nemotron-Terminal, LLM-in-Sandbox} into interactive environments by wrapping them with programmatic rules, as shown in Fig. \ref{fig:symbolic_synthesis}. This approach leverages the abundance of high-quality static data to ensure the correctness and scalability of the environments. Next, we introduce in detail the developmental trajectory of these methods.

% \noindent
\textbf{A significant portion of task-driven synthesis methods focus on coding agent environments \cite{CodeGym, Agent2World}.} Early work includes SWE-Gym \cite{SWE-Gym} which packages issues and pull requests (PRs) from 11 Python repositories sourced from GitHub into executable Python environments using Docker. To overcome the limited scale of early work, recent research explores fully automated scaling \cite{R2E-Gym,Nemotron-Terminal}. For example, Scale-SWE \cite{Scale-SWE} and SWE-Hub \cite{SWE-Hub} propose an automated environment synthesis paradigm based on multi-agent collaboration. Scale-SWE \cite{Scale-SWE} decomposes the synthesis task into three agents: Environment Builder, Unit-test Creator, and Problem Statement Writer. SWE-Hub \cite{SWE-Hub} proposes a similar concept (e.g., Env Agent, Bug Agent) and further emphasizes the automatic generation of more realistic and complex bug tasks at the system level. SCALER \cite{SCALER} utilizes competitive programming data to synthesize verifiable environments with adjustable difficulty, providing stable reward signals for RL. As environment volumes grow, environment configuration costs escalate, prompting efficiency optimizations. MEnvAgent \cite{MEnvAgent} employs an incremental patching mechanism to enable environment reuse, while SWE-smith \cite{SWE-smith} proposes sharing a single environment image across thousands of tasks within the same repository. Conversely, DockSmith \cite{DockSmith} trains specialized models to automate the generation and repair of Dockerfiles to improve construction efficiency. Beyond standard software engineering, CLI-Gym \cite{CLI-Gym} shifts focus to command line interfaces through an agentic environment inversion paradigm where agents intentionally break environments to generate historical error data. 

% \noindent
\textbf{Beyond coding tasks, other studies explore broader tool usage environments.} AgentScaler \cite{AgentScaler} focuses on the number and breadth of environments by using 30,000 heterogeneous APIs gathered from existing datasets to construct diverse environments. EnvScaler \cite{EnvScaler} focuses on synthesizing complex state dependencies to increase the difficulty and depth of environments through programmatic synthesis and dual agent filtering strategies. Additionally, Agentic CPT \cite{Agent_CPT} extends this paradigm by transforming unstructured corpora from Wikipedia or CommonCrawl into interactive environments for continual pre-training.

% \noindent
\textbf{Further research investigates knowledge intensive environments in vertical domains.} MedAgentGym \cite{MedAgentGym} standardizes static biomedical datasets into code centric reasoning tasks. SciAgentGym \cite{SciAgentGym} introduces execution grounded synthesis by equipping existing scientific benchmarks with real tools and verification mechanisms. PaperArena \cite{PaperArena} constructs literature analysis environments for complex cross document reasoning. FinMTM \cite{FinMTM} synthesizes financial environments based on industrial reports for interaction through MCP. V-GameGym \cite{V-GameGym} incorporates visual rendering as feedback, marking the evolution of environment synthesis from text-only to multimodal streams. 

% \noindent
\textbf{Finally, recent efforts integrate multi domain tasks to explore idealized environment synthesis.} LLM-in-Sandbox \cite{LLM-in-Sandbox} presents a unified perspective by utilizing standard operating systems as universal sandboxes for all tasks. This demonstrates that non-coding tasks like mathematics can also achieve substantial performance gains through interaction within encapsulated environments.

\subsubsection{Real-World-Driven Synthesis}
\label{sec:real world driven synthesis}

Real-world-driven synthesis projects real-world complex interaction mediums (e.g., Web and Game) into simplified virtual environments for training and evaluation \cite{AgentSynth, AI_GAMESTORE, LMGAME-BENCH, GameDevBench, VeriEnv, AutoWebWorld, GLM-5, MiMo-V2-Flash, DIVE, KAT-Coder-V2}. Given the immense volume and vast action spaces of real world systems, this paradigm achieves significantly greater scalability than task driven approaches. Illustrated by Fig. \ref{fig:symbolic_synthesis}, these methods basically project complex, diverse real-world environments into simplified forms.

% \noindent
\textbf{Traditional real world computing systems include GUI, OS, and complex software tools.} Some frameworks leverage these authentic applications to construct massive scale environments \cite{Agent2World, SWE-Universe} designed to generate high quality and long horizon trajectory data. To automate scaling, AgentSynth \cite{AgentSynth} exploits information asymmetry where progressively generating and executing simple subtasks is significantly easier than solving complex long horizon problems in a single attempt. This mechanism enables the low cost synthesis of extensive planning tasks with controllable difficulty. TaskCraft \cite{TaskCraft} similarly automates the synthesis of tool invocation environments with adjustable difficulty levels grounded in external web environments. Recently, the emergence of the MCP addresses the fragmentation of real-world interfaces. OSWorld-MCP \cite{OSWorld-MCP} pioneers the integration of GUI with standardized protocol tool invocations within synthetic environments to bridge the gap between pure visual interactions and text based tool usage. VeriEnv \cite{VeriEnv} utilizes generative models to directly clone real-world websites for virtual environments. To stress test agent capabilities in realistic settings, MCPMark \cite{MCPMark} synthesizes complex tasks with profound operational depth directly upon authentic protocol servers (e.g., GitHub and Notion). Applying this technology to vertical domains, MedMCP-Calc \cite{MedMCP-Calc} constructs highly rigorous healthcare environments by integrating real electronic medical record databases and clinical guideline retrieval systems. 

% \noindent
\textbf{Recent research also simulates environments with visual feedback and physical constraints.} AI Gamestore \cite{AI_GAMESTORE} proposes a multiverse of human games as a novel standard for evaluating models by synthesizing open-ended evaluation environments from digital platforms like Steam. lmgame-Bench \cite{LMGAME-BENCH} encapsulates diverse video games into standardized Gymnasium style APIs through modular harnesses. Shifting focus to game development, GameDevBench \cite{GameDevBench} leverages video tutorials and testing frameworks inherent to game engines. This approach translates hard to quantify visual feedback into executable testing tasks to evaluate agent proficiency in multimodal code generation. Finally, EmbodiedBench \cite{EmbodiedBench} achieves the systematic synthesis ranging from high level semantic planning to low level micro control (e.g., manipulation of robotic arms). 

\subsubsection{De Novo Synthesis}
\label{sec:synthesis from scratch}

Emerging de novo synthesis methods aim to directly generate diverse interactive environments utilizing minimal seed examples or zero shot approaches, as demonstrated in Fig. \ref{fig:symbolic_synthesis}. Compared to previous paradigms, this method offers the highest degree of freedom and the most expansive environmental space \cite{DeepSeek-V3.2, LongCat-Flash-Thinking-2601, AutoEnv, LOGIGEN, InfiniteWeb, NL2Plan, RandomWorld, MiMo-V2-Flash}. It closely approximates the ideal scenario of freely scalable environments.

% \noindent
\textbf{The primary challenge of de novo synthesis involves ensuring rigorous internal logic within generated systems.} AutoForge \cite{AutoForge} addresses the instability of previous reverse synthesis by prompting generative models to first construct scalable state structures and model underlying logic graphs based on tool invocation relationships before producing code. LOGIGEN \cite{LOGIGEN} advances this by introducing logic driven forward deductive synthesis. This framework utilizes an architect agent to compile rules into physical environments supported by SQLite databases, fundamentally altering traditional reverse synthesis pathways. AutoEnv \cite{AutoEnv} focuses on cross domain generalized generation by hierarchically decoupling the underlying logic of the environment from the surface interaction, successfully yielding datasets for numerous highly heterogeneous environments entirely from scratch.

% \noindent
\textbf{Automated verification is equally crucial since generated environments inevitably suffer from model hallucinations.} Within SWE domains, ScaleEnv \cite{ScaleEnv} proposes methodologies based on procedural testing and graph expansion to guarantee code correctness and task diversity without human intervention. Agent World Model \cite{Agent_world_model} ensures quality stability during massive scaling by accurately restoring the software development process and embedding execution level self correction mechanisms. Furthermore, SWE-Playground \cite{SWE-Playground} constructs a pioneering purely synthetic training pipeline that completely removes the dependency on external data sources like GitHub. 

% \noindent
\textbf{Fully automated synthesis extends broadly to other domains.} Endless Terminals \cite{Endless_Terminals} synthesizes thousands of terminal environments from scratch by sampling dimensions such as file operations and network configurations. To mitigate data contamination, gg-bench \cite{gg-bench} employs random sampling to invent entirely novel two-player games. These games are then automatically encapsulated into standard Gymnasium interfaces to establish dynamic benchmarks.

\subsubsection{Summary}

% Overall, from task-driven to real-world-driven, and then to synthesis from scratch, the degrees of freedom and synthetic environment space of these three types of environment synthesis methods are gradually increasing. This trajectory closely aligns with the core vision and ultimate objective of environment scaling in contemporary agent research \cite{silver2025welcome}. However, as they gradually move away from dependence on real-world data, synthetic environments also encounter critical challenges. To ensure robust training, critical issues concerning \textbf{task diversity}, environmental \textbf{complexity}, execution \textbf{correctness}, generation \textbf{quality}, and \textbf{distributional bias} relative to real world scenarios must be concurrently resolved.

\begin{tcolorbox}[takeaway,title={Takeaway 5.1}]
\begin{itemize}
    \item \textbf{Evolution of Scalability:} The progression from task driven to real world driven and to de novo synthesis methods represents a continuous expansion in synthesis freedom and overall environment space. This trajectory closely aligns with the core vision and ultimate objective of environment scaling in contemporary agent research \cite{silver2025welcome}.
    
    \item \textbf{Future Directions:} As paradigms gradually reduce dependence on real-world data, synthetic environments also encounter critical challenges. To ensure robust training, critical issues concerning \textbf{task diversity}, environmental \textbf{complexity}, execution \textbf{correctness}, generation \textbf{quality}, and \textbf{distributional bias} relative to real world scenarios must be concurrently resolved.
\end{itemize}
\end{tcolorbox}

\renewcommand{\arraystretch}{1.5} 
\begin{table*}[htbp]
 \caption{
   Representative works for symbolic synthesis. \textbf{Base Model} refers to the model used for environment synthesis; ``-'' indicates that no model was used or that it was not specified. \textbf{Data Source} refers to the data on which the environment synthesis was based.
    }
\resizebox{\textwidth}{!}{
\begin{tabular}{lccccccc}
\toprule
\textbf{Name} &
  \textbf{Modality} &
 \textbf{Architecture} &   \textbf{Base Model} &
  \textbf{Task Domain} &
  \textbf{Data Source}  & \textbf{Evaluation}\\ \midrule
\rowcolor{gray!10!white} \multicolumn{7}{c}{\textbf{\textit{Task-Driven Synthesis (\cref{sec:task driven synthesis})}}}\\ 
 Planetarium\cite{Planetarium} &\Text & LLM & - &  Code & IPC & Correctness, Complexity  \\ 
 SWE-Gym\cite{SWE-Gym} &\Text & LLM & - &  Code & GitHub & Correctness \\
 R2E-Gym\cite{R2E-Gym} & \Text & LLM & Sonnet-3.5-V2 & Code & GitHub & Correctness, Complexity \\
 VML PDDL\cite{VML_PDDL} &\Text & LLM & Qwen2.5-Coder &  Code & IPC & Correctness \\
 TheoryCoder\cite{TheoryCoder} &\Text & LLM & GPT-4o &  Game & VGDL, BabyAI, etc. & Correctness \\ 
 SWE-smith\cite{SWE-smith} & \Text & LLM & o3, Claude 3.7 & Code & GitHub & Correctness, Complexity \\
 MedAgentGym\cite{MedAgentGym} & \Text & LLM & GPT-4.1-mini, etc. & Domain-Specific & MIMIC-III, eICU, etc. & Correctness, Complexity, Fidelity \\
 V-GameGym\cite{V-GameGym} & \Text\hspace{-0.4em}\Image\hspace{-0.4em}\Video & VLM & Claude-Sonnet-4 & Game & OpenCoder, The Stack v2 & Correctness, Diversity, Complexity \\
 Agentic CPT\cite{Agent_CPT} & \Text & LLM & LLM & Tool & CommonCrawl, Wikipedia,etc. & Correctness \\
 AgentScaler\cite{AgentScaler} & \Text & LLM & LLM & Tool & ToolBench, API-Gen & Correctness \\
 PaperArena\cite{PaperArena} & \Text\hspace{-0.4em}\Image & VLM & GPT-5, Claude 4, Gemini-2.5 & Domain-Specific & OpenReview, Open Access & Diversity, Complexity \\
 Agent2World\cite{Agent2World} & \Text & LLM & GPT-4.1-mini, Llama-3.1-8b & Domain & LIMA & Correctness, Fidelity \\
 SCALER\cite{SCALER} & \Text & LLM & GLM-4.6 & Domain & CodeContests & Correctness, Diversity, Complexity \\
 MEnvAgent\cite{MEnvAgent} & \Text & LLM & Kimi-K2, Gemini-3, Claude-4.5 & Code & GitHub & Correctness, Complexity \\
 EnvScaler\cite{EnvScaler} & \Text & LLM & GPT-4.1, Qwen3-235B & Tool & API-Bank, ToolACE & Correctness, Diversity, Complexity \\
 LLM-in-Sandbox\cite{LLM-in-Sandbox} & \Text & LLM & - & Domain & Instruction Pre-Training, etc. & Correctness \\
 DockSmith\cite{DockSmith} & \Text & LLM & Qwen3* & Code & GitHub & Correctness, Complexity \\
 Scale-SWE\cite{Scale-SWE} & \Text & LLM & DeepSeek, Gemini-3 & Code & GitHub, PyPI & Correctness, Diversity \\
 SciAgentGym\cite{SciAgentGym} & \Text\hspace{-0.4em}\Image & VLM & - & Domain-Specific & SciInstruct, GPQA, BMMR, etc. & Correctness, Diversity, Complexity, Fidelity \\
 CLI-Gym\cite{CLI-Gym} & \Text & LLM & LLM & Code & SWE-smith, GitHub & Correctness \\
 FinMTM\cite{FinMTM} & \Text\hspace{-0.4em}\Image & VLM & Gemini-3-Pro & Domain-Specific & - & Correctness, Diversity, Complexity \\
 Nemotron-Terminal\cite{Nemotron-Terminal} & \Text & LLM & DeepSeek-V3.2 & Code & OpenMathReasoning, etc. & Correctness \\
 Hybrid-Gym\cite{HYBRID-GYM} & \Text & LLM & GPT-4o-mini, Claude-4.5, etc. & Code & GitHub, SWE-Gym, RepoST & Correctness, Diversity \\
 SWE-Hub\cite{SWE-Hub} & \Text & LLM & LLM & Code & GitHub & Correctness, Diversity \\

\midrule
\rowcolor{gray!10!white} \multicolumn{7}{c}{\textbf{\textit{Real-World-Driven Synthesis (\cref{sec:real world driven synthesis})}}}\\ 

 EnvGen\cite{EnvGen} & \Text\hspace{-0.4em}\Image & VLM & GPT-4 & Embodied & Crafter,Heist & Correctness, Fidelity \\
 EmbodiedBench\cite{EmbodiedBench} & \Text\hspace{-0.4em}\Image & VLM & GPT-4 & Embodied & ALFRED, AI2-THOR, etc. & Correctness, Fidelity \\
 lmgame-Bench\cite{LMGAME-BENCH} & \Text\hspace{-0.4em}\Image & VLM & o3 & Game & video games & Correctness, Complexity \\
 AgentSynth\cite{AgentSynth} &\Text\hspace{-0.4em}\Image & VLM & GPT-4.1 & Domain & OSWorld,persona hub & Correctness, Complexity \\
 TaskCraft\cite{TaskCraft} & \Text\hspace{-0.4em}\Image & VLM & GPT-4.1 & GUI, Tool & Web & Correctness, Complexity \\
 OSWorld-MCP\cite{OSWorld-MCP} & \Text\hspace{-0.4em}\Image & VLM & o3 & GUI & OSWorld & Correctness, Complexity \\
 MedMCP-Calc\cite{MedMCP-Calc} & \Text & LLM & GPT-OSS, DeepSeek-V3.1, etc. & Domain-Specific & MDCalc, MIMIC-IV, etc. & Correctness \\
 AI GameStore\cite{AI_GAMESTORE} & \Text\hspace{-0.4em}\Image & VLM & Claude-4.5, Gemini-2.5 & Game & App Store, Steam & Correctness, Complexity, Fidelity \\
 GameDevBench\cite{GameDevBench} & \Text\hspace{-0.4em}\Image\hspace{-0.4em}\Video & VLM & GPT-5 & Game & YouTube, Godot 4 & Correctness, Complexity, Fidelity \\
 AutoWebWorld\cite{AutoWebWorld} & \Text\hspace{-0.4em}\Image & VLM & GPT-5.1, Gemini-3, etc. & GUI & Facebook, Gmail, etc. & Correctness, Complexity \\
 SWE-Universe\cite{SWE-Universe} & \Text & LLM & Qwen-Next-80B-A3B* & Code & GitHub & Correctness \\
 VeriEnv\cite{VeriEnv} & \Text\hspace{-0.4em}\Image  & VLM & GPT-5.2 & GUI & Web & Correctness, Complexity, Fidelity \\
 DIVE\cite{DIVE} & \Text  & LLM & Claude-4-Sonnet & Tool & Wikipedia, PubMed, etc. & Correctness, Complexity \\
 KAT-Coder-V2\cite{KAT-Coder-V2} & \Text  & LLM & Claude & Code & GitHub, etc. & Correctness, Complexity \\
 \midrule
\rowcolor{gray!10!white} \multicolumn{7}{c}{\textbf{\textit{De Novo Synthesis (\cref{sec:synthesis from scratch})}}}\\ 

 RandomWorld\cite{RandomWorld} & \Text & LLM & GPT-4o & Tool & - & Correctness, Complexity \\
 NL2Plan\cite{NL2Plan} & \Text & LLM & GPT-4o & Tool & - & Correctness, Complexity, Fidelity \\
 gg-bench\cite{gg-bench} & \Text & LLM & OpenAI o1 & Game & - & Correctness, Diversity, Complexity \\
 AutoEnv\cite{AutoEnv} & \Text & LLM & Claude-4-Sonnet & Domain & - & Correctness, Diversity, Complexity \\
 AutoForge\cite{AutoForge} & \Text & LLM & Qwen3-Thinking & Code, Tool & - & Correctness, Diversity, Complexity \\
 SWE-Playground\cite{SWE-Playground} & \Text & LLM & Gemini-2.5, GPT-4.1, etc. & Code & - & Correctness \\
 InfiniteWeb\cite{InfiniteWeb} & \Text\hspace{-0.4em}\Image & VLM & GPT-5 & GUI & - & Correctness, Diversity \\
 Endless Terminals\cite{Endless_Terminals} & \Text & LLM & o3 & Code & - & Correctness, Complexity \\
Agent World Model\cite{Agent_world_model} &\Text\hspace{-0.4em}\Image & LLM & GPT-5 & Tool & - & Correctness, Diversity \\
 ScaleEnv\cite{ScaleEnv} & \Text & LLM & Deepseek-V3.2, GPT-5.1, etc. & Code & - & Correctness, Diversity, Complexity \\
 LOGIGEN\cite{LOGIGEN} & \Text & LLM & DeepSeek-V3.2 & Tool & - & Correctness, Complexity \\

\bottomrule

\end{tabular}
}
\end{table*}

\subsection{Neural Synthesis}
\label{sec:Neural Synthesis}

\begin{figure*}[t]
    \centering
    \includegraphics[width=1\textwidth]{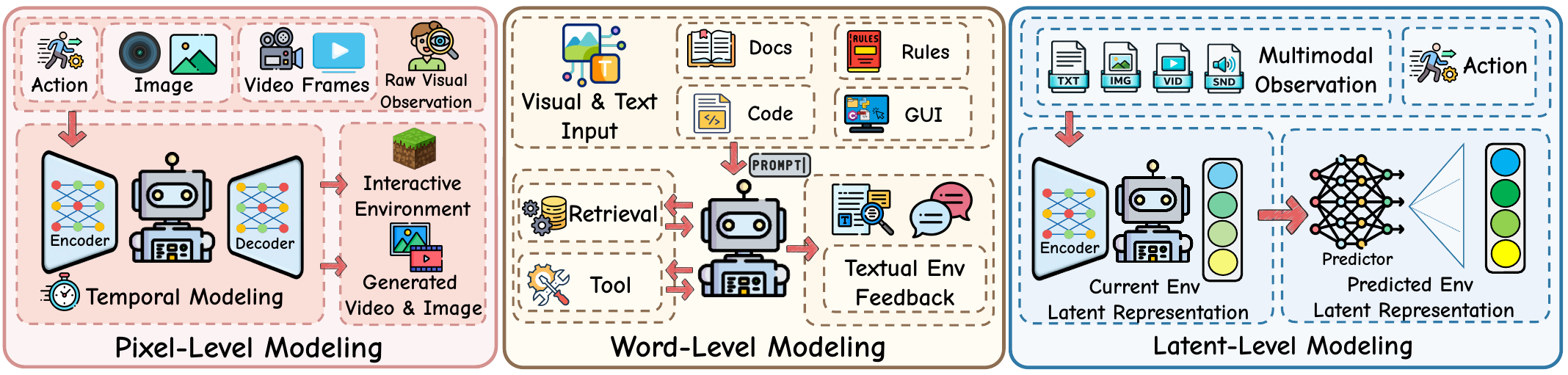 }
    \caption{
    Three neural environment synthesis paradigms are presented: Pixel-Level Modeling, Word-Level Modeling, and Latent-Level Modeling, trading off fidelity and abstraction.
    }
    \label{fig:Neural_synthesis}
\end{figure*}

Neural synthesis refers to modeling the environment using a neural network. Such approaches are typically centered around world models, which are neural models that learn to simulate the environment's state transitions and observations, enabling agents to interact with a learned environment instead of the real one. Depending on the form of representation used for environment modeling, we categorize them into three types: \textbf{(1) Pixel-Level Modeling} (\cref{sec:Pixel-level Modeling}), which directly models environment dynamics from pixel-level observations such as images or videos; \textbf{(2) Word-Level Modeling} (\cref{sec:Word-level Modeling}), which represents environment states in natural language; and \textbf{(3) Latent-Level Modeling} (\cref{sec:Latent-level Modeling}), which learns structured representations of the environment in a latent space and captures the dynamics through low-dimensional embeddings. An overview of these paradigms is illustrated in Fig. \ref{fig:Neural_synthesis}.

\subsubsection{Pixel-Level Modeling}
\label{sec:Pixel-level Modeling}
Pixel-level modeling refers to modeling the environment directly in the raw visual observation space. Rather than relying on semantic abstraction or representation compression, this paradigm learns environment dynamics from high-dimensional visual signals \cite{DIAMOND, DreamGen, Pandora}, thereby preserving richer physical detail and scene information. 

\textbf{Some approaches focus primarily on modeling the temporal evolution of the visual world with limited emphasis on complex interaction.} 
A representative approach \cite{Recurrent_World_Models_Facilitate_Policy_Evolution} simulates game environments using a world model that compresses spatial representations with a VAE and models temporal dynamics as a probability distribution with an MDN-RNN, where the stochasticity is controlled via a temperature parameter. Subsequent research shifts toward using world models for agent decision-making and interaction, introducing action-conditioned modeling so that the model can predict future visual observations given agent behavior. In embodied tasks, EVA \cite{EVA} integrates a vision-language model with a video generative model, enabling the framework to generate future visual observations conditioned on language-described actions. In GUI tasks, ViMo \cite{ViMo} uses natural language instructions rather than abstract action codes, and trains an image predictor with both visual features and text instructions as dual conditioning inputs.

\textbf{Other approaches extend world models to support long-horizon interactive environments.} 
Matrix-Game \cite{Matrix-Game} trains on a large-scale Minecraft dataset to enable interactive video generation conditioned on continuous keyboard and mouse inputs, supporting temporally extended trajectories rather than short clips. Its successor Matrix-Game 2.0 \cite{Matrix-Game_2.0} further improves long-horizon consistency by introducing a scalable data production pipeline and a frame-level action injection module, enabling stable interactive generation at the minute scale. 
NeuralOS \cite{NeuralOS} targets operating system environments with persistent and evolving states. It models long-range dependencies by maintaining system state with a hierarchical RNN, while combining diffusion-based rendering and multi-stage training to mitigate exposure bias over extended interaction sequences.

\subsubsection{Word-Level Modeling}
\label{sec:Word-level Modeling}

Word-level modeling benefits from the expressive and general-purpose nature of natural language, making it well-suited for modeling environments across diverse domains \cite{GTM, CWM, ITP}. Compared to pixel-level modeling, it provides a higher-level representation with lower computational cost, and is more capable of handling complex reasoning and long-horizon planning tasks.

\textbf{A line of work focuses on using language models to represent environment knowledge.} WKM \cite{WKM} synthesizes task knowledge (i.e., knowledge required to accomplish a task) and state knowledge (i.e., summaries of environment states based on historical actions), providing prior guidance and constraints for action generation during agent reasoning. Complementary to this, retrieval-augmented approaches extend language world models with external knowledge sources. Specifically, R-WoM \cite{R-WoM} enhances environment understanding by retrieving tutorial knowledge and integrating it into reasoning, improving long-horizon simulation through reasoning-aware retrieval and listwise reward estimation. 

\textbf{Other research turns to using language world models to perform dynamic reasoning over the environment.} MobileDreamer \cite{MobileDreamer} represents key GUI elements in structured text and introduces a rollout imagination strategy that constructs a tree of predictions to enable multi-step lookahead, thereby improving action selection. Also targeting GUI tasks, Code2World \cite{Code2World} proposes representing GUI states as renderable code, and using a language model to generate code to express GUI state transitions. However, reasoning with language world models inevitably suffers from hallucination, inconsistency, and long-horizon drift, making verifiability a critical concern. RAP \cite{RAP} incorporates Monte Carlo Tree Search to explore multiple reasoning trajectories and selects the optimal path, ensuring that the output is traceable and more reliable.

\subsubsection{Latent-level Modeling}
\label{sec:Latent-level Modeling}
Unlike pixel-level methods that directly model raw visual observations and word-level approaches that describe environments through linguistic symbols, latent-level modeling operates in a learned representation space \cite{IWM, DINO-WM, seq-JEPA}. 
\textbf{By shifting prediction from the observation space to the latent space, models can focus on the structural information that is most relevant to semantics and dynamics.} I-JEPA \cite{I-JEPA} predicts representations of image patches in latent space rather than pixels. By masking large target blocks and leveraging spatially distributed context blocks, it encourages the model to learn semantic representations, achieving improvements in both performance and efficiency. seq-JEPA \cite{seq-JEPA} extends this framework from single images to action-observation sequences, introducing inductive biases to separate equivariant and invariant representations within a single model. This enables the model to perform classification with invariant representations and fine-grained discrimination with equivariant representations, while generalizing to sequential tasks such as gaze prediction and path integration.

\textbf{However, learning latent representations from scratch is costly.} Subsequent research therefore turns to reusing representations from pretrained foundation models, enabling cross-domain generalization and diverse applications. A line of work \cite{DINO-WM, DINO-world, DINO-Foresight} leverages frozen pretrained visual representations from DINOv2 to model environment and temporal dynamics in latent space. These methods avoid learning visual representations from scratch, and train dynamics models on offline video data or trajectories, supporting efficient future forecasting, generalization, and task-adaptive planning. V-JEPA 2 \cite{V-JEPA_2}, pretrained on one million hours of internet video, requires only 62 hours of robot data to fine-tune and achieves zero-shot robot planning, highlighting the potential of pretrained foundation models to enable scalable and generalizable world modeling.

\subsubsection{Summary}

% In summary, neural synthesis based on world models can be categorized into pixel-level, word-level, and latent-level modeling. Pixel-level modeling operates at low abstraction with high fidelity, but suffers from information redundancy; word-level modeling provides high abstraction with lower fidelity, often leading to information loss due to compression; latent-level modeling lies between the two, aiming to balance representation compactness and predictive capability, while facing challenges in interpretability and reliance on learned representations. As world models continue to evolve, maintaining long-horizon consistency, preserving essential information, and achieving efficient modeling remain fundamental challenges. Addressing these issues, together with improving controllability and integration across different representation spaces, is critical for building robust and general-purpose neural environment modeling systems.

% \vspace{-3pt}
\begin{tcolorbox}[takeaway,title={Takeaway 5.2}]
\begin{itemize}
\item \textbf{Pros and Cons:} Each modeling paradigm involves inherent trade-offs.
Pixel-level modeling operates at low abstraction with high fidelity, but suffers from information redundancy. 
Word-level modeling provides high abstraction with lower fidelity, often leading to information loss due to compression. 
Latent-level modeling aims to balance representation compactness and predictive capability, while facing challenges in interpretability and reliance on learned representations.

\item \textbf{Future Directions:} Maintaining long-horizon consistency, preserving essential information, and achieving efficient modeling remain fundamental challenges. Addressing these issues, together with improving controllability and integration across different representation spaces, is critical for building robust and general-purpose neural environment modeling systems.
\end{itemize}
\end{tcolorbox}
% \vspace{-5pt}
\renewcommand{\arraystretch}{1.5} 
\begin{table*}
\caption{Representative methods in neural synthesis. \textbf{Modality} refers to the output modality of the model. \textbf{Base Model} marked as "-" indicates that no pre-existing foundation model is used, while "base model*" denotes that the base model is fine-tuned.}
\resizebox{\textwidth}{!}{
\begin{tabular}{lccccccc}
\toprule
\textbf{Name} &
  \textbf{Modality} &
 \textbf{Architecture} &   \textbf{Base Model} &
  \textbf{Task Domain} &
  \textbf{Data Source}  & \textbf{Evaluation}\\ \midrule
\rowcolor{gray!10!white} \multicolumn{7}{c}{\textbf{\textit{Pixel-Level Modeling (\cref{sec:Pixel-level Modeling})}}}\\ 
 EVA \cite{EVA} &\Video & VLM,Diffusion & ChatUniVi*,Dynamicrafter* &  Embodied & COCO,CC3M,etc. & Correctness, Fidelity \\
 ViMo \cite{ViMo} &\Text\hspace{-0.4em}\Image & Diffusion,VLM & Stable Diffusion*,GPT-4o &  GUI & Android Control,etc. & Correctness, Complexity, Fidelity \\
 Matrix-Game \cite{Matrix-Game} &\Video & Diffusion & HunyuanVideo* &  Game & MineDojo,etc. & Correctness, Diversity, Fidelity \\
 Matrix-Game 2.0 \cite{Matrix-Game_2.0} &\Video & Diffusion & SkyReelsV2* &  Game & Sekai,etc. & Correctness, Diversity, Fidelity \\
 DIAMOND \cite{DIAMOND} &\Image & Diffusion & - &  Game & Atari 100k & Correctness, Diversity, Complexity, Fidelity \\
 GameNGen \cite{GameNGen} &\Video & Diffusion & Stable Diffusion v1.4* &  Game & ViZDoom & Correctness, Complexity, Fidelity \\
 DreamGen \cite{DreamGen} &\Video & Diffusion & WAN2.1* &  Embodied & RoboCasa,DROID & Correctness, Diversity, Fidelity \\
 EnerVerse \cite{EnerVerse} &\Video & Diffusion & DynamicCrafter* &  Embodied & RT1,Taco-Play,etc. & Correctness, Complexity, Fidelity \\
 Genie Envisioner \cite{Genie_Envisioner} &\Video & Diffusion & LTX-Video 2B*,COSMOS2 2B* &  Embodied & AgiBot-World-Beta & Correctness, Diversity, Fidelity \\
 KeyWorld \cite{KeyWorld} &\Video & Diffusion & - &  Embodied & LIBERO & Correctness, Complexity, Fidelity \\
 MineWorld \cite{MINEWORLD} &\Video & LLM,VAE & LLaMA*,VQ-VAE* &  Game & Video PreTraining & Correctness, Fidelity \\
 NeuralOS \cite{NeuralOS} &\Image & RNN,Diffusion & - &  GUI & private & Correctness, Complexity, Fidelity \\
 FlowDreamer \cite{FlowDreamer} &\Video & Diffusion & Stable Diffusion 2.1* &  Embodied & RT-1,Language Table,etc. & Correctness, Fidelity \\
 Cosmos Policy \cite{Cosmos_Policy} &\Video & Diffusion & Cosmos-Predict2-2B* &  Embodied & LIBERO,RoboCasa,etc. & Correctness, Fidelity \\
 DreamZero \cite{DreamZero} &\Video & Diffusion & Wan2.1-I2V-14B* &  Embodied & DROID,etc. & Correctness, Diversity, Fidelity\\
 GAIA-2 \cite{GAIA-2} &\Video & Diffusion & - & Embodied & private & Correctness, Diversity, Fidelity \\
 Cosmos-Drive \cite{Cosmos-Drive} &\Video & Diffusion & Cosmos-7B-Text2World* & Embodied & Real Driving Scene,etc. & Complexity, Diversity, Fidelity\\
 PEWM \cite{PEWM} &\Video & Diffusion & DynamiCrafter* & Embodied & RLBench,LIBERO,etc. & Correctness, Fidelity \\
 AdaWorld \cite{AdaWorld} &\Video & Diffusion & Stable Video Diffusion* & Embodied,Game & Gym Retro,etc. & Correctness, Diversity, Fidelity\\
 LingBot-VA \cite{LingBot-VA} &\Video & Diffusion & Wan2.2-5B* & Embodied & Agibot,RoboMind,etc. & Correctness, Complexity, Fidelity\\
 MeWM \cite{MeWM} &\Text\hspace{-0.4em}\Image & VLM,Diffusion & - & Domain-Specific & HCC-TACE-Seg,etc. & Correctness, Fidelity\\
 Cosmos \cite{Cosmos} &\Video & Diffusion & - & Embodied & private & Correctness, Diversity, Fidelity\\
 Pandora \cite{Pandora} &\Video & VLM,Diffusion & Chat-Univi*,DynamiCrafter* & Cross-Domain & WebVid-10M,etc. & Correctness\\
 LIVE \cite{LIVE} &\Video & Diffusion & NFD* & Embodied,Game & RealEstate10K,etc. & Correctness, Fidelity\\
 \midrule
\rowcolor{gray!10!white} \multicolumn{7}{c}{\textbf{\textit{Word-Level Modeling (\cref{sec:Word-level Modeling})}}}\\ 
 GTM \cite{GTM} &\Text & LLM & Qwen2.5-1.5B* &  Tool & KernelBench,etc. & Correctness, Fidelity \\
 WKM \cite{WKM} &\Text & LLM & Mistral-7B-Instruct-v0.2* &  Cross-Domain & ALFWorld,WebShop,etc. & Correctness \\
 Code2World \cite{Code2World} &\Text & VLM & Qwen3-VL-8B-Instruct* &  GUI & Android Control & Correctness, Fidelity \\
 CWM \cite{CWM} &\Text & LLM & - &  Code & OpenCodeReasoning,etc. & Correctness, Complexity \\
 Dyna-Think \cite{Dyna-Think} &\Text & LLM & Qwen2.5-32B-Instruct* &  GUI & private & Correctness, Fidelity \\
 ITP \cite{ITP} &\Text & LLM & Qwen3-8B* &  Embodied & ALFWorld,ScienceWorld & Correctness \\
 WebDreamer \cite{WebDreamer} &\Text & VLM & GPT-4o,Qwen2-VL-7B-Instruct* &  GUI & private & Correctness, Complexity, Fidelity \\
 MobileDreamer \cite{MobileDreamer} &\Text & LLM & Qwen3-8B* &  GUI & Android Control & Correctness, Complexity, Fidelity \\
 VLWM \cite{VLWM} &\Text & VLM & PerceptionLM-8B* &  Cross-Domain & COIN,CrossTask,etc.  & Correctness, Complexity \\
 R-WoM \cite{R-WoM} &\Text & VLM & Qwen2.5-VL-72B-Instruct &  GUI & -  &  Correctness, Complexity \\
 RAP \cite{RAP} &\Text & LLM & LLaMA-33B &  Game & -  & Correctness, Complexity \\
 SimuRA \cite{SimuRA} &\Text & LLM & GPT-4o &  Cross-Domain & -  & Correctness \\
 FPWC \cite{FPWC} &\Text & VLM & GPT-4V &  GUI & -  & Correctness, Complexity \\
 WWM \cite{WWM} &\Text & LLM & Gemini 2.5 Flash &  GUI & -  & Correctness, Diversity \\
 DyMo \cite{DyMo} &\Text & LLM & Cohere's R7B* &  Tool & BFCL  & Correctness \\
 Simia \cite{Simia} &\Text & LLM & o4-mini & Cross-Domain & - & Correctness, Diversity, Fidelity \\
 WebWorld \cite{WebWorld} &\Text & LLM & Qwen3-8/14/32B* & GUI & private & Correctness, Diversity, Fidelity \\
 gWorld \cite{gWorld} &\Text & VLM & Qwen3-VL-8/32B* & GUI & Android in the Wild,etc. & Correctness, Fidelity \\
 SWE-World \cite{SWE-World} & \Text & LLM & Qwen2.5-32B/72B* & Code & GitHub, SWE-bench, etc. & Correctness \\

\midrule
\rowcolor{gray!10!white} \multicolumn{7}{c}{\textbf{\textit{Latent-Level Modeling (\cref{sec:Latent-level Modeling})}}}\\ 
V-JEPA 2 \cite{V-JEPA_2} &\Embed & JEPA & - & Cross-Domain & VideoMix22M & Correctness, Fidelity \\
I-JEPA \cite{I-JEPA} &\Embed & JEPA & - & Cross-Domain & ImageNet-1K & Correctness, Diversity \\
IWM \cite{IWM} &\Embed & JEPA & - & Cross-Domain & ImageNet & Correctness, Fidelity \\
DINO-world \cite{DINO-world} &\Embed & ViT & DINOv2 & Cross-Domain & private & Correctness \\
EchoJEPA \cite{EchoJEPA} &\Embed & JEPA & - & Domain-Specific & MIMIC-IV-Echo,etc. & Correctness \\
DINO-WM \cite{DINO-WM} &\Embed & ViT & DINOv2 & Cross-Domain & private & Correctness, Fidelity \\
DINO-Foresight \cite{DINO-Foresight} &\Embed & ViT & DINOv2 & Embodied & Cityscapes,nuScenes & Correctness, Complexity, Fidelity \\
seq-JEPA \cite{seq-JEPA} &\Embed & JEPA & - & Cross-Domain & 3DIEBench,CIFAR100,etc. & Correctness, Complexity \\

\bottomrule

\end{tabular}}
\end{table*}

\subsection{Quality Control and Evaluation of Environments}

After synthesizing environments, a central question is how to evaluate and control their quality, namely whether these environments can serve as reliable substrates for training and evaluation. We summarize existing evaluation practices along four complementary dimensions: \textbf{correctness}, \textbf{diversity}, \textbf{complexity}, and \textbf{fidelity}. We next discuss how prior work assesses environment quality from these perspectives.

\subsubsection{Correctness}

Correctness is the most fundamental requirement for synthesized agentic environments. It requires not only that the environment has valid state transitions, but also that the synthesized tasks can be legally executed, have valid solutions, and are paired with reliable verifiers that provide correct reward signals \cite{SWE-smith, R2E-Gym, CLI-Gym}.

A common strategy for symbolic environment is program execution and unit testing \cite{SWE-Hub}. In coding and software environments, many works use sandboxes or unit tests to check whether generated environments can run, while filtering runtime errors and timeout cases \cite{SWE-Universe, SWE-Hub, MCP-Universe}. For example, SWE-Gym \cite{SWE-Gym} and Scale-SWE \cite{Scale-SWE} validate patches or generated tasks through execution tests. GameDevBench \cite{GameDevBench} uses deterministic tests in the Godot scripting framework, V-GameGym \cite{V-GameGym} executes and repairs generated code in a UI sandbox. ScaleEnv \cite{ScaleEnv} verifies generated tools and database code through procedural testing, while Endless Terminals \cite{Endless_Terminals} validates container construction, initial-state tests, and final completion tests before retaining a task.

Other works focus more on solvability and trajectory validation. Tool-based environments often execute golden tool sequences or expert trajectories, and use final-state comparison to determine whether the task can be truly completed \cite{Agent2World, SciAgentGym, AgentGen}. For instance, AutoForge \cite{AutoForge} obtains the target state by executing golden tool sequences, while AgentSynth \cite{AgentSynth} uses a verifier to estimate long-horizon task completion.

Some works also evaluate the reliability of the verifier itself. In OS, GUI, and Web environments, task completion often cannot be judged by simple string matching. Execution-based evaluators or expert reviews are therefore used to prevent failed trajectories from being falsely accepted \cite{OSWorld-MCP, EnvScaler, AutoWebWorld}. For example, MCP-Universe \cite{MCP-Universe} uses static and dynamic evaluators instead of relying on unstable LLM-as-a-judge evaluation, while OSWorld-MCP \cite{OSWorld-MCP} combines execution validation with expert review to ensure tool correctness. InterCode \cite{InterCode} checks custom reward functions with gold commands, and Mobile-Env \cite{Mobile-Env} combines system signals such as screen text and user responses to improve state estimation and outcome evaluation. Some works further measure the agreement between LLM or human verifiers and ground-truth task completion \cite{AgentSynth, Terminal-Bench, SciAgentGym}.

% Thus, correctness in environment synthesis is not merely about whether an environment can run. It requires a complete validation loop covering execution, trajectory feasibility and reward reliability.

Unlike the aforementioned evaluation approaches based on explicit execution and verifiers, correctness in neural environment synthesis is difficult to assess through executable tests due to the widespread multi-modal content (e.g., videos and images) in neurally synthesized environments. As a result, correctness is often reformulated as measuring whether the synthesized outputs satisfy predefined constraints. Such methods typically perform evaluation from semantic consistency, interaction consistency, and trajectory deviation. For instance, DreamGen \cite{DreamGen} employs Qwen-VL-2.5 to score generated videos and determine whether they faithfully follow the given language instructions, further incorporating human annotations to improve reliability. Matrix-Game \cite{Matrix-Game}, on the other hand, utilizes a pretrained inverse dynamics model to infer actions from generated videos and compares them with ground-truth keyboard inputs using average accuracy to evaluate interaction correctness. At a finer granularity, Genie Envisioner \cite{Genie_Envisioner} combines LLM-based semantic evaluation with rule-based metrics, using symmetric Hausdorff distance to measure spatial deviations between generated and reference trajectories, and normalized dynamic time warping (NDTW) to capture temporal alignment of action sequences. For GUI scenarios, MobileDreamer \cite{MobileDreamer} introduces the mIoU metric to assess spatial accuracy by computing the overlap between predicted and ground-truth element bounding boxes. In addition, some works \cite{WKM, Dyna-Think} directly adopt final task completion as an overall criterion to evaluate the validity of synthesized environments.

\subsubsection{Diversity}

Diversity concerns whether a synthesized environment collection provides broad and non-redundant coverage over task spaces, state spaces, tool spaces, and linguistic expressions. This is crucial for agent training and evaluation: if environments are too homogeneous, agents may overfit to surface patterns rather than acquire generalizable interaction abilities. A high-quality environment set should therefore cover different task types, capabilities, tools, and interaction patterns \cite{SWE-World}.

Some methods in symbolic environments use similarity filtering to avoid generating many semantically redundant samples \cite{InfiniteWeb}. For example, Agent World Model \cite{Agent_world_model} adopts embedding-based deduplication to prevent scenario collapse, EnvScaler \cite{EnvScaler} uses embedding similarity and t-SNE visualization to analyze topic dispersion, and V-GameGym \cite{V-GameGym} applies clustering to select diverse game seeds from large code repositories.

Other methods expand diversity through structured coverage. Tool environments often track tool categories and API combinations \cite{ToolBench, API-Bank}, while OS, GUI, and Web environments emphasize coverage over web tasks, operation types and distractor tools \cite{Mind2Web, WebArena, Mobile-Env}. For example, AutoForge \cite{AutoForge} uses random walks over tool dependency graphs to generate diverse tool sequences. MCP-Universe \cite{MCP-Universe} covers multiple real MCP servers and task domains. And other methods improve diversity through the generation process itself \cite{AgentBench, AI_GAMESTORE}. TaskCraft \cite{TaskCraft} synthesizes multi-hop tool-use tasks from Web, PDF, and image sources.

In neural environment synthesis, beyond improving diversity through expanded task and environment coverage as discussed above, diversity is also commonly characterized by the richness of the model’s output distribution under identical input conditions, namely, whether it can produce multiple valid outputs that are semantically consistent yet diverse in details. For example, Genie Envisioner \cite{Genie_Envisioner} generates multiple videos under the same language instruction and computes pairwise similarities based on CLIP global video embeddings as a diversity score, thereby evaluating the model’s ability to produce diverse yet valid trajectories. GAIA-2 \cite{GAIA-2} perturbs latent representations extracted from real videos and performs conditional denoising to generate diverse environment variants while preserving core semantics and ego-motion trajectories. Furthermore, some works decouple actions from context to verify the model's ability to synthesize diverse environments with consistent core actions. For instance, AdaWorld \cite{AdaWorld} transfers identical actions to different contexts to evaluate generalization across diverse environments. Similarly, I-JEPA \cite{I-JEPA} performs multiple stochastic decodings for the same input and target location using the RCDM framework, and assesses diversity by comparing variations in core semantics, object structures, and low-level details across samples.

\subsubsection{Complexity}

Complexity measures whether synthesized environments have appropriate difficulty. A useful environment should be neither too simple nor systematically impossible. If tasks are too easy, they may fail to improve agent capabilities and can even lead to overfitting; if tasks are too hard or unsolvable, evaluation results provide little useful feedback. Thus, complexity evaluation should not simply increase task length, but construct solvable task distributions that can distinguish agents with different abilities \cite{AgentSynth, TaskCraft, VeriEnv}.

Some methods quantify complexity through structure parameters. Many symbolic environments report interaction steps, number of tools, API calls, state entities, or lines of code to describe structural difficulty \cite{PaperArena, MedAgentBench}. For example, AutoForge \cite{AutoForge} increases the logical complexity of tool-use tasks through DAG structures. OSWorld-MCP \cite{OSWorld-MCP} characterizes difficulty by the number of tool-calling rounds. LOGIGEN \cite{LOGIGEN} controls complexity by injecting multi-variable conditions, state-dependent constraints, role permissions, and irreversible transitions into the compiled policy environment. NL2Plan \cite{NL2Plan} stratifies planning tasks by expected optimal plan length, dividing problems into short, medium, and long settings.

Other methods calibrate complexity using strong-model or human performance \cite{SWE-smith, gg-bench, AI_GAMESTORE}. gg-bench \cite{gg-bench} filters games using win-rate disparity from RL self-play to remove tasks with insufficient discriminative power. In more realistic task settings, AI GameStore \cite{AI_GAMESTORE} uses expert cognitive ratings and human median performance to calibrate game difficulty.

Complexity in neural environment synthesis is often reflected in the reasoning depth and width explored during the generation process. For example, WebDreamer \cite{WebDreamer} adjusts the planning horizon (1/2/3 steps) to control multi-step prediction depth, thereby influencing task complexity. MobileDreamer \cite{MobileDreamer} varies the number of candidate actions per node to modulate the branching factor of the search space. VLWM \cite{VLWM} regulates complexity by adjusting the number of candidate plans in the System-2 stage, while RAP \cite{RAP} increases complexity by scaling the number of Monte Carlo Tree Search iterations, thereby expanding search depth.

\begin{figure*}[htbp]
    \centering
    \includegraphics[width=1\textwidth]{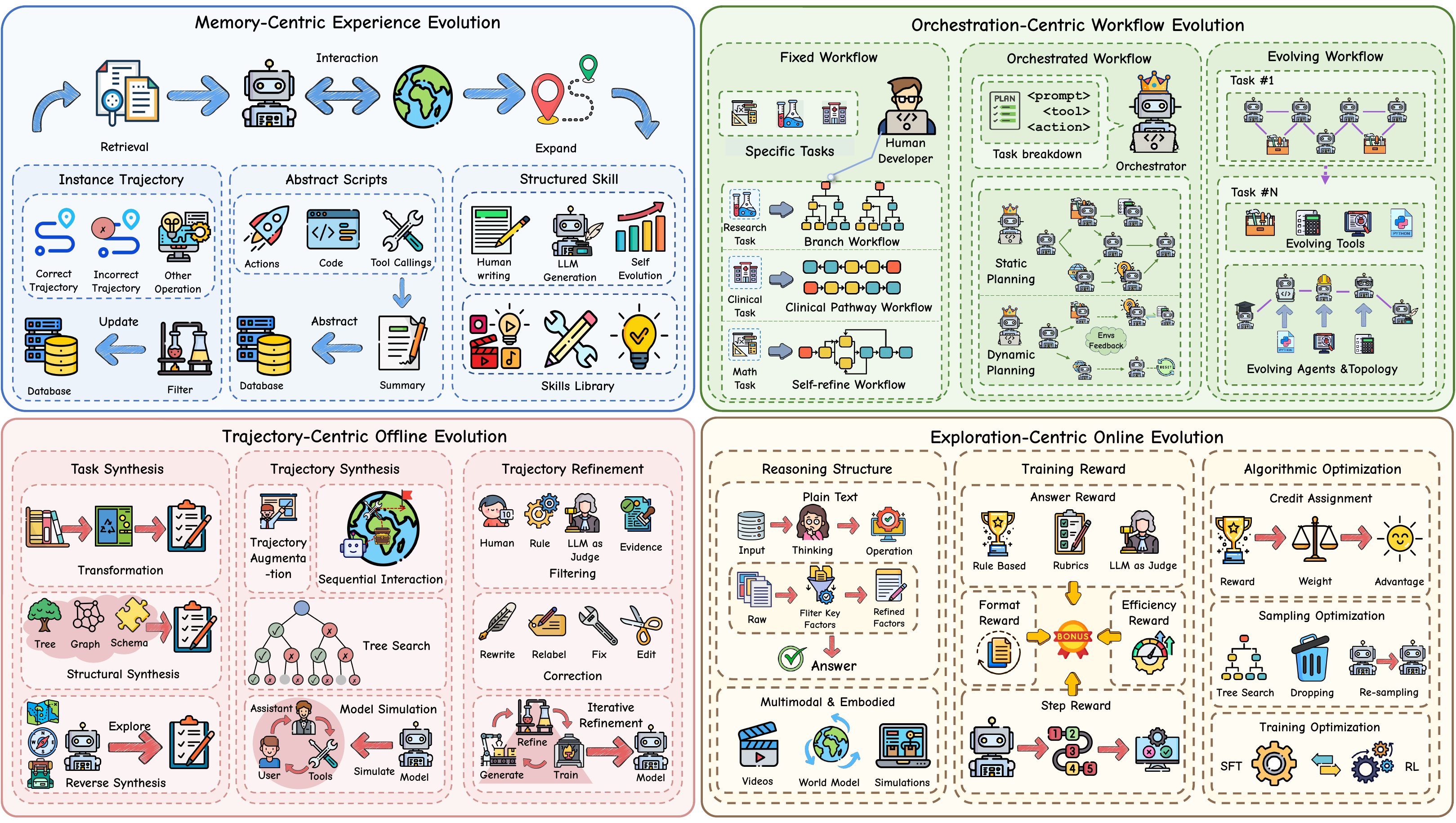 }
    \caption{
    Overview of agent evolution paradigms. Existing methods are organized into four categories: Memory-Centric Experience Evolution, which enhances agent capabilities through accumulated experiences; Orchestration-Centric Workflow Evolution, which adapts agent workflows to optimize task performance; Trajectory-Centric Offline Evolution, which refines agent behavior through synthetic task interaction data; and Exploration-Centric Online Evolution, which strengthens agent capabilities through real-time learning and adaptation via reinforcement.
    }
    \label{fig:agent_evo}
\end{figure*}
\subsubsection{Fidelity}

Fidelity concerns whether synthesized environments faithfully reflect the real systems, physical dynamics or user behaviors they aim to model. It is related to, but distinct from, correctness. A task may be executable inside a synthesized simulator, but if its dynamics, API behavior, or user workflow diverges from the real world, it may still have low fidelity. Fidelity therefore asks a higher-level question: an environment should not only work, but work in a way that resembles the target setting \cite{WebArena, Genie_Envisioner, Matrix-Game}.

For symbolic synthesis, fidelity is mainly evaluated through consistency between symbolic rules and reference environments. In PDDL and rule-based game environments, fidelity comes from the alignment between synthesized rules and the original task intent or human intuition \cite{VML_PDDL, Exploration_Walk, GVGAI-LLM}. For example, some PDDL-generation works manually compare generated instances with original natural language intents and measure how many edits are needed to make the task semantically correct \cite{VML_PDDL}. OSWorld-MCP \cite{OSWorld-MCP} uses expert reviews to check whether synthesized tools have real practical value. Therefore, fidelity in symbolic environments mainly emphasizes realistic task logic and real user needs, rather than visual or physical realism.

In neural environment synthesis, fidelity primarily depends on the consistency between synthesized and real environments in terms of perceptual appearance and dynamic evolution, namely, whether generated outputs resemble real-world environments in visual, physical, and temporal aspects. In embodied, gaming, and GUI scenarios, some works \cite{DIAMOND, GameNGen, GAIA-2} employ metrics such as FID, FVD, and LPIPS to measure the similarity between synthesized and real environments in perceptual distributions. DreamGen \cite{DreamGen} further evaluates fidelity from a physical perspective by quantifying the realism of rigid-body motion, collisions, gravity, and contact forces, thereby capturing dynamic consistency. Additionally, some approaches assess fidelity from a subjective perspective. For example, WebWorld \cite{WebWorld} introduces the Web Turing Score, in which a large language model is tasked with distinguishing between real and synthesized environments, thereby measuring perceptual realism and overall plausibility. From a temporal perspective, dynamic consistency is also considered a key factor. For instance, GAIA-2 \cite{GAIA-2} adopts Frechet Video Motion Distance (FVMD) to compare motion patterns between generated and real videos based on keypoint trajectories, while EnerVerse \cite{EnerVerse} incorporates expert evaluations of continuity and stability in robotic behaviors to assess fidelity in long-horizon interactions.

\subsubsection{Summary}
\begin{tcolorbox}[takeaway,title={Takeaway 5.3}]
\begin{itemize}
\item \textbf{Dimensions:} Quality covers not only \textbf{instance-level} correctness, but also \textbf{distribution-level} diversity, complexity, and fidelity.

\item \textbf{Direction:} Existing work has moved from post generation filtering toward closed-loop generation with continuous validation and refinement for quality-controlled environment synthesis.

\item \textbf{Limitations:} While correctness has developed well-established evaluation frameworks, the study of diversity, complexity, and fidelity remains preliminary, which affects the credibility.
\end{itemize}
\end{tcolorbox}
\section{Agent Evolution} \label{sec:agent_evo}
\begin{table*}[tbp]
\caption{The statistics of Memory-Centric Experience Evolution methods. }
\centering
\renewcommand{\arraystretch}{1.5}
\resizebox{\textwidth}{!}{%
\begin{tabular}{lccccc}\toprule
\textbf{Name} &
 \textbf{Task} &
 \textbf{Model} &
  \textbf{Generalization} &
  \textbf{Experience Update} &
 \textbf{Train} \\ \midrule

\rowcolor{gray!10!white} \multicolumn{6}{c}{\textbf{\textit{Instance Trajectory Experience (\cref{sec:instance_trajectory_experience})}}} % \multicolumn{6}{c}{}
\\
OpenAgent \cite{OpenAgent} &
  Code &
  GPT-4, GPT-3.5-Turbo &
  \Yes &
  \Yes &
  \No \\

CoPS \cite{CoPS} &
  Cross-Domain &
  Llama 3.1 8B Instruct, Llama 3.1 70B Instruct &
  \Yes &
  \Yes &
  \No \\

ELLMER \cite{ELLMER} &
  Embodied &
  GPT-4, DALL-E &
  \Yes &
  \Yes &
  \No \\

Memp \cite{Memp} &
  Cross-Domain &
  GPT-4o, Qwen2.5-72B &
  \No &
  \Yes &
  \No \\

Synapse \cite{Synapse} &
  GUI &
  CodeLlama-7B, GPT-3.5 &
  \Yes &
  \Yes &
  \No \\

WorldMM \cite{WorldMM} &
  Image \& Video &
  Qwen3-VL-8B-Instruct &
  \No &
  \Yes &
  \No % &
  \\ 

DGM \cite{DGM} &
  Code &
  o3-mini, Claude-3.5-Sonnet &
  \Yes &
  \Yes &
  \No % &
  \\ 
   \midrule
\rowcolor{gray!10!white} \multicolumn{6}{c}{\textbf{\textit{Abstract Scripts Experience (\cref{sec:abstract_scripts_experience})}}}   \\
Reasoning Bank \cite{Reasoning-Bank} &
  Cross-Domain &
  Gemini-2.5-flash, Gemini-2.5-pro, Claude-3.7-sonnet &
  \Yes &
  \Yes &
  \No % &
  \\
  
Agent-Pro \cite{Agent-Pro} &
  Game &
  GPT-3.5-Turbo-0613, GPT-4-0613, etc. &
  \No &
  \Yes &
  \No \\

Agent-KB \cite{AGENTKB_2} &
  Cross-Domain &
  GPT-4o, GPT-4.1, etc. &
  \Yes &
  \Yes &
  \No \\

DeepAgent \cite{DeepAgent} &
  Tool &
  QwQ-32B &
  \Yes &
  \Yes &
  \Yes \\

FLEX \cite{FLEX} &
  Cross-Domain &
  Claude-Sonnet-4, DeepSeek-V3.1-Terminus, etc. &
  \No &
  \Yes &
  \No \\

O-Mem \cite{O-Mem} &
  Cross-Domain &
  GPT-4.1, GPT-4o-mini &
  \No &
  \Yes &
  \No \\

H-EPM \cite{H-EPM} &
  Tool &
  GPT-4.1-mini, GPT-4.1, etc. &
  \Yes &
  \Yes &
  \Yes \\

Reme \cite{Reme} &
  Tool &
  Qwen3-8B, Qwen3-32B &
  \No &
  \Yes &
  \No \\

BIFROST \cite{BIFROST} &
  DeepResearch &
  Llama-3.2-3B-Instruct, etc. &
  \Yes &
  \No &
  \No \\

PhysMem \cite{PhysMem} &
  Embodied &
  Gemini-3.0-Flash, Qwen3-VL &
  \Yes &
  \Yes &
  \No \\

AWM \cite{AWM} &
  GUI &
  GPT-3.5, GPT-4 &
  \Yes &
  \Yes &
  \No \\
  \midrule
\rowcolor{gray!10!white} \multicolumn{6}{c}{\textbf{\textit{Structured Skill Experience (\cref{sec:structure_skill_experience})}}} % \multicolumn{6}{c}{} 
\\
SAGE \cite{SAGE} &
  Tool &
  Qwen2.5-32B-Instruct &
  \No &
  \Yes &
  \Yes \\

ASI \cite{ASI} &
  Cross-Domain &
  GPT-4o, Claude-3.5-sonnet &
  \Yes &
  \Yes &
  \No \\

SkillWeaver \cite{SkillWeaver} &
  DeepResearch &
  GPT-4o, GPT-4o-mini &
  \No &
  \Yes &
  \No \\

SkillRL \cite{SkillRL_2} &
  Cross-Domain &
  Qwen2.5-7B-Instruct &
  \Yes &
  \Yes &
  \Yes \\

SkillOrchestra \cite{SkillOrchestra} &
  Tool &
  Qwen2.5-3B &
  \Yes &
  \Yes &
  \Yes \\

\bottomrule
\end{tabular}%
}
\label{tab:experience}
\end{table*}
\textbf{Agent Evolution} refers to the process by which an agent’s capabilities advance through various mechanisms, typically realized through continuous interaction with the external environment. This process encompasses both external structural adaptations and internal parametric changes. Specifically, external structural adaptations include \textbf{Memory-Centric Experience Evolution} (\cref{sec:agent_evo_experience_utilization}), which enhances the model's task-processing capabilities by accumulating and leveraging experience from the external environment, and \textbf{Orchestration-Centric Workflow Evolution} (\cref{sec:agent_evo_agentic_workflow_design}), which adapts to diverse environmental tasks through the design of specialized architectures. The internal parametric changes include \textbf{Trajectory-Centric Offline Evolution} (\cref{sec:agent_evo_synthetic_data_generation}), which improves the model’s task adaptation by synthesizing and designing complex data generation pipelines, and \textbf{Exploration-Centric Online Evolution} (\cref{sec:agent_evo_reinforcement_learning_optimization}), which strengthens the model's capabilities through reinforcement learning processes within specific environments. The framework is shown as Fig. \ref{fig:agent_evo}.

\subsection{Memory-Centric Experience Evolution}
\label{sec:agent_evo_experience_utilization}
Experience helps agents better understand and master their current capabilities \cite{survey_Memory_in, survey_Rethinking_Memory_in, survey_Rethinking_Memory,Think_While_Watching,Learning_How_to_Remember}. These methods typically involve storing trajectories, experiences, or process knowledge in an external memory base, and then retrieving relevant information to enhance the model’s ability to handle problems. Through leveraging experience at different levels of granularity and diverse utilization strategies, the model can maintain long-term memory, strengthen its ability to handle specific problems, and enable the co-evolution of an agent’s memory and capabilities across tasks. Based on how experience is summarized and represented, we categorize it into the following types: (1) \textbf{Instance Trajectory Experience}: This refers to the complete interaction trajectories and actions of an agent with the environment. (2) \textbf{Abstract Scripts Experience}: This type of experience abstracts and summarizes multiple trajectories into reusable script-like knowledge, enabling more efficient reuse across similar tasks. (3) \textbf{Structured Skill Experience}: Skills represent a highly structured form of externalized experience that can be modularly stored and invoked on demand.

\subsubsection{Instance Trajectory Experience}
\label{sec:instance_trajectory_experience}
\textbf{Instance Trajectory Experience represents the most fundamental and specific type of experience, referring to the complete record of actions an agent takes during its interaction with the environment}. Instance trajectories directly reflect the agent's decision-making process in a specific task context by recording the correct or incorrect actions, or other agent operations, providing the most detailed information. However, they are highly context-dependent and have limited generalization ability. 
For example, Synapse \cite{Synapse} provides the LLM with entire trajectories to improve the performance in Web tasks. WorldMM \cite{WorldMM} utilizes the multiple categories of memory, where the visual memory provides the detailed scenes of the targets. The maintenance of experience is also crucial. Existing methods typically organize experiences into databases and perform maintenance operations to ensure their effectiveness \cite{Memp, CoPS}. For example, Memp \cite{Memp} refines historical task trajectories into both fine-grained instructions and high-level abstract scripts, systematically exploring strategies for building, retrieving, and updating procedural memories.

\subsubsection{Abstract Scripts Experience}
\label{sec:abstract_scripts_experience}
\textbf{Abstract Scripts Experience is derived by generalizing across multiple instance trajectories to capture reusable task execution patterns}. Compared to individual trajectories, script-level experience no longer focuses on specific environment states, but instead extracts the typical sequence of steps or operational logic underlying a class of tasks. By abstracting trajectories into scripts, such experience achieves stronger generalization and enables cross-domain retrieval \cite{AGENTKB_2, BIFROST, AWM}. Reasoning-Bank \cite{Reasoning-Bank} extracts generalizable reasoning strategies from the agent’s successful and failed experiences and stores them in a structured memory bank. During testing, relevant memories are retrieved to guide decision-making. Besides, some works \cite{MemGPT, O-Mem} focus on utilizing the memory to maintain the long-horizon information of previous works, rather than to handle specific problems. For instance, O-Mem \cite{O-Mem} further organizes experience into structured categories, including persona memory, working memory, and episodic memory, thereby supporting personalized and consistent long-term interactions. 
In addition, some approaches focus on how memory is utilized during decision-making \cite{Voyager, H-EPM}. For example, H-EPM \cite{H-EPM} integrates tool graphs with procedural memory to guide execution, where each edge in the graph serves as a compact representation of procedural summaries. 

Other works emphasize the maintenance and evolution of such experiences to enable continuous self-improvement \cite{Agent-Pro, FLEX, Training-Free_GRPO}. For instance, Agent-Pro \cite{Agent-Pro} enhances performance through iteratively reflecting on previous actions and fine-tuning to correct the wrong steps, while FLEX \cite{FLEX} and Reme \cite{Reme} reflect on both successful and failed experiences to construct and maintain structured experience repositories, facilitating ongoing agent evolution.

\subsubsection{Structured Skill Experience}
\label{sec:structure_skill_experience}
\textbf{Skills are the experience organization mechanism to expand the LLM's capabilities}. Skills have gained widespread attention due to their detail and comprehensiveness. Many methods focus on exploring the automatic synthesis of skills to enhance the capabilities and efficiency of intelligent agents.

Some methods automatically induce and generate procedural skills through interaction with the environment to help the agent complete tasks. ASI \cite{ASI} induces procedural skills automatically through interaction with the environment, helping the agent execute tasks. SkillWeaver \cite{SkillWeaver} enables the agent to autonomously discover and transform website features into reusable Python API skills. 

As research progresses, more works are focused on using skills to complete tasks. By building a skill library, SAGE \cite{SAGE} enhances the self-evolving capabilities of large language model agents. It uses a sequential rollout mechanism to reuse programming skills and integrates skill-based rewards to motivate the generation and utilization of high-quality skills. SkillRL \cite{SkillRL_2} distills redundant raw interaction trajectories into a hierarchical skill bank and co-evolves it with policy learning, thus improving the agent's task success rate and reasoning efficiency.

\subsubsection{Summary}

\begin{tcolorbox}[takeaway,title={Takeaway 6.1}]
\begin{itemize}
\item \textbf{Comparative Analysis:} The granularity of experience and its utilization strategies have become central research focuses. The former concerns how experience is acquired and represented, while the latter emphasizes how experience is adapted and evolved for downstream tasks. 
In this context, both the scale and effectiveness of experience are critical, which has led to increasing attention on \textbf{skills} as a structured and reusable form of experience. 

\item \textbf{Future Directions:} An important direction is to develop more systematic and fine-grained frameworks for experience management and utilization. Such systems should be capable of aggregating and integrating large-scale experiences from human contributions and open-source resources. 
Especially, current experience management mechanisms remain relatively underdeveloped, with limited support for comprehensive operations such as insertion, deletion, updating, and retrieval, and lacking principled design for scalable and efficient management. This is an important direction for experience utilization.
\end{itemize}
\end{tcolorbox}

% update by hongbang in 20260407
\renewcommand{\arraystretch}{1.1} 
\begin{table*}[htbp]
\caption{The statistics of Orchestration-Centric Workflow Evolution methods.}
\resizebox{\textwidth}{!}{
\begin{tabular}{lccccc}\toprule
\textbf{Name} & \textbf{Task} & \textbf{Collaboration} & \textbf{Model} & \textbf{Train} & \textbf{Multi-Agent} \\ \midrule
\rowcolor{gray!10!white} 
\multicolumn{6}{c}{\textbf{\textit{Fixed Workflow (\cref{sec:fixed_workflow})}}} \\
Simulate Before Act \cite{Simulate_Before_Act} & DeepResearch & Tree & GPT-4o & \No & \No \\
WebVoyager \cite{WebVoyager} & DeepResearch & Sequential & GPT-4-V, Claude-3-O, GPT-4o & \No & \No \\
ManuSearch \cite{ManuSearch} & DeepResearch & Sequential & QwQ-32B, DeepSeek-R1, etc. & \No & \Yes \\
% Multimodal DeepResearcher \cite{Multimodal_DeepResearcher} & DeepResearch & Sequential + Iterative & Claude-3.7-S, Qwen3-235B, etc. & \No & \Yes \\
Open Deep Search \cite{Open_Deep_Search} & DeepResearch & Sequential & DeepSeek-R1, Llama-3.1-70B & \No & \No \\
Search-o1 \cite{Search_o1} & DeepResearch & Sequential & QwQ-32B-P & \No & \No \\
SearchAgent-X \cite{SearchAgent_X} & DeepResearch & Sequential & Qwen-7/14B & \No & \No \\
Thought of Search \cite{Thought_of_Search} & DeepResearch & Sequential & GPT-4 & \No & \No \\
WKM \cite{WKM} & Domain-Specific & Sequential & Mistral-7B, Llama-3-8B, etc. & \Yes & \No \\
MedAgent-Zero \cite{MedAgent_Zero} & Domain-Specific & Iterative & GPT-3.5/4/o, o1-P & \No & \Yes \\
QuantAgent \cite{QuantAgent} & Domain-Specific & Iterative & GPT-4 (0125-P) & \No & \Yes \\
Agentless \cite{Agentless} & Code & Sequential & GPT-4o, o1 & \No & \No \\
AutoCodeRover \cite{AutoCodeRover} & Code & Sequential & GPT-4 (0125-P) & \No & \Yes \\
CodeNav \cite{CodeNav} & Code & Iterative & GPT-4, Mixtral-8x22B-IT, etc. & \No & \No \\
CodeR \cite{CodeR} & Code & Graph & GPT-4 (1106) & \No & \Yes \\
MAGIS \cite{MAGIS} & Code & Sequential + Iterative & GPT-4, GPT-3.5, Claude-2, etc. & \No & \Yes \\
MetaGPT \cite{METAGPT} & Code & Sequential & GPT-4, GPT-3.5-T, etc. & \No & \Yes \\
% OpenHands \cite{OpenHands} & Code & Iterative & Claude-4.5/4-S, GPT-5, etc. & \No & \No \\
SWE-agent \cite{SWE_agent} & Code & Sequential & GPT-4-T, Claude-3-O & \No & \No \\
VideoAgent \cite{VideoAgent} & Video & Sequential & GPT-4, CogAgent, LaViLa & \No & \No \\
AgileThinker \cite{AgileThinker} & Game & Sequential & DeepSeek-V3/R1, Gemini-2.5-Flash & \No & \No \\
LVAgent \cite{LVAgent} & Video & Iterative & Qwen2-VL, InternVL-2.5, etc. & \Yes & \No \\
Explorer \cite{Explorer} & GUI & Sequential & Phi-3.5V, Qwen2-VL-7B & \Yes & \Yes \\
\midrule
\rowcolor{gray!10!white} 
\multicolumn{6}{c}{\textbf{\textit{Automated Workflow (\cref{sec:automated_workflow})}}} \\
AutoFlow \cite{AutoFlow} & Cross-Domain & Graph & GPT-4-1106-preview, Mixtral-8x7B & \Yes & \Yes \\
ScoreFlow \cite{ScoreFlow} & Cross-Domain & Graph & Llama-3.1-8B-Instruct, GPT-4o-mini, etc. & \Yes & \Yes \\
RobustFlow \cite{RobustFlow} & Cross-Domain & Graph & Qwen3-32B, GPT-4o-mini, etc. & \Yes & \Yes \\
DyFlow \cite{DyFlow} & Cross-Domain & DAG & Phi-4, GPT-4.1, GPT-4o-mini, etc. & \Yes & \Yes \\
AgentFLow \cite{AgentFlow} & Cross-Domain & Sequential + Iterative & Qwen2.5-7B-IT & \Yes & \Yes \\
MaAS \cite{MaAS} & Cross-Domain & DAG & GPT-4o-mini, Qwen-2.5-72B-Instruct, etc. & \Yes & \Yes \\
FlowReasoner \cite{FlowReasoner} & Code & Graph & DeepSeek-R1-Distill-Qwen-7B, etc. & \Yes & \Yes \\
% Kimi K2.5 \cite{Kimi_K2.5} & Cross-Domain & Graph & Kimi K2.5 & \Yes & \Yes \\
Mindsearch \cite{Mindsearch} & DeepResearch & DAG & GPT-4o, InternLM2.5-7B, etc. & \No & \Yes \\
WideSeek-R1 \cite{wideseek_r1} & DeepResearch & Tree & Qwen3-4B & \Yes & \Yes \\
Workflow-R1 \cite{Workflow_R1} & DeepResearch & Sequential & Qwen2.5-7B-Instruct, etc. & \Yes & \Yes \\
ResearStudio \cite{ResearStudio} & DeepResearch & Sequential & GPT-4.1, o4, o3, GPT-4o, etc. & \No & \No \\
Webpilot \cite{Webpilot} & DeepResearch & Iterative & GPT-4o, GPT-3.5 & \No & \Yes \\
Plan-and-act \cite{PLAN_AND_ACT} & Domain-Specific & Sequential & Llama-3.3-70B-I, QWQ-32B & \Yes & \Yes \\
AOrchestra \cite{AORCHESTRA} & Domain-Specific & Sequential & Gemini-3-Flash, DeepSeek-V3.2, etc. & \Yes & \Yes \\
Dr. MAS \cite{Dr_MAS} & Domain-Specific & Tree + Iterative & Qwen3-4/8B, Qwen2.5-3/7B-I & \Yes & \Yes \\
Workforce \cite{OWL} & Domain-Specific & Sequential & Qwen2.5-32B-I, GPT-4o, Claude-3.7-S & \Yes & \Yes \\
ControlLLM \cite{ControlLLM} & Embodied & DAG & ChatGPT, Llama-7/13B, GPT-4 & \Yes & \No \\
HuggingGPT \cite{HuggingGPT} & Embodied & DAG & GPT-3.5-T, GPT-4, Alpaca, etc. & \No & \No \\
HyperAgent \cite{HYPERAGENT} & Code & DAG & Claude-3-S/H, Llama-3-70B, etc. & \Yes & \No \\
RepairAgent \cite{RepairAgent} & Code & Graph & GPT-3.5 & \No & \No \\
SWE-Search \cite{SWE_Search} & Code & Tree & GPT-4o, Qwen2.5-72B-IT, etc. & \Yes & \No \\
Alibaba LingmaAgent \cite{Alibaba_LingmaAgent} & Code & Tree & GPT-4-T, GPT-4o, Claude-3-O/S & \No & \No \\
MindAgent \cite{MINDAGENT} & Embodied & Tree & GPT-4 (0613) & \No & \Yes \\
Code Researcher \cite{Code_Researcher} & Code & Graph & GPT-4o, o1 & \No & \No \\
SCLPlan \cite{SCLPlan} & Embodied & Sequential + Iterative & Llama-3.1-8B, Llama-3-70B & \No & \No \\
MGA \cite{MGA} & GUI & Sequential + Iterative & O3 & \Yes & \Yes \\
\midrule
\rowcolor{gray!10!white} 
\multicolumn{6}{c}{\textbf{\textit{Evolving Workflow (\cref{sec:evolving_workflow})}}} \\
AFlow \cite{AFlow} & Cross-Domain & Graph & GPT-4o-mini, DeepSeek-V2.5, etc. & \No & \No \\
AgentSquare \cite{AgentSquare} & Cross-Domain & Graph & GPT-4o, GPT-3.5-turbo & \No & \No \\
Puppeteer \cite{Puppeteer} & Cross-Domain & Graph & Qwen2.5, Llama-3.1, etc. & \No & \Yes \\
ADAS \cite{ADAS} & Cross-Domain & Graph & GPT-4, GPT-3.5, Claude-Haiku, etc. & \No & \No \\
ReasonRAG \cite{ReasonRAG} & DeepResearch & Iterative & Qwen2.5-7B-IT & \Yes & \No \\
MUSE \cite{MUSE} & Domain-Specific & Sequential + Iterative & Gemini-2.5-Flash, DeepSeek-V3 & \No & \No \\
AvaTaR \cite{AVATAR} & Domain-Specific & Sequential + Iterative & Claude-3-O, GPT-4/o & \No & \No \\
ReCreate \cite{ReCreate} & Domain-Specific & Iterative & GPT-5, Claude-4.5-O & \Yes & \No \\
Yunjue Agent \cite{Yunjue_Agent} & Domain-Specific & Iterative & Gemini-3-Pro, GPT-5, GPT-5 & \No & \Yes \\
Chain-of-Agents \cite{Chain-of-Agents} & Domain-Specific & Sequential & Qwen2.5-3/7/32B-IT & \Yes & \Yes \\
ClinicalReTrial \cite{ClinicalReTrial} & Domain-Specific & Iterative & DeepSeek-V3, Claude & \No & \Yes \\
EvoClinician \cite{EvoClinician} & Domain-Specific & Iterative & Gemini-3-Pro, GPT-5.1, MedGemma & \No & \Yes \\
Grammar-based Search \cite{Grammar_based_Search} & Domain-Specific & Sequential & GPT-4o, GPT-4.1, GPT-5 & \No & \Yes \\
CASCADE \cite{CASCADE} & Domain-Specific & DAG & GPT-5, O3, Claude-4.5-S, etc. & \No & \Yes \\
STEVE \cite{STEVE} & Embodied & Sequential & Llama-2-7/13B & \Yes & \Yes \\
LATM \cite{LATM} & Cross-Domain & Iterative & GPT-4, GPT-3.5, etc. & \No & \Yes \\
Criticize-Reflect \cite{Criticize_Reflect} & Embodied & Graph & Llama2-70B, GPT-4, GPT-3.5 & \No & \Yes \\
\bottomrule
\end{tabular}
}
\end{table*}

\subsection{Orchestration-Centric Workflow Evolution}
\label{sec:agent_evo_agentic_workflow_design}
A \textbf{Workflow} is a graph-based topology of multi-step processes powered by LLMs, where nodes encompass LLM calls, tool executions, subagents, or functions, and edges dictate the fixed, conditional, or cyclic routing \cite{ToT,ReAct,ReWoo,Self-Refine,LLM+P}. The approach centers on decomposing objectives into detailed steps, enabling the agent to navigate complex environments. Harness frameworks like AutoGen  \cite{autogen} and OpenHands \cite{OpenHands} facilitate the deployment of these systems by providing the essential infrastructure and interfaces for inter-agent communication. Based on the degree of structural autonomy, we categorize these methods into three types: \textbf{(1) Fixed Workflow}, characterized by explicit data flow and predefined roles; \textbf{(2) Automated Workflow}, where a designated orchestrator agent dynamically coordinates the available worker agents; and \textbf{(3) Evolving Workflow}, where the entire structure undergoes persistent evolution through interaction with the environment.

\subsubsection{Fixed Workflow}
\label{sec:fixed_workflow}
\textbf{Fixed Workflow} constitutes a task execution framework with a deterministic logical topology, predefined by developers during the system design phase. The framework incorporates hard-coded sequential logic \cite{Explorer, Simulate_Before_Act}, conditional branching \cite{CodeR, VideoAgent}, and local loops including prespecified error retry protocols \cite{LVAgent, QuantAgent}. Within this configuration, the agent functions as a modular unit tasked with performing specific atomic operations. Consequently, the agent has no authority over the global workflow topology and its decision making is bounded by the localized task node. Recent studies demonstrate the utility of this framework in complex engineering scenarios. MetaGPT \cite{METAGPT} codifies standard operating procedures for software development into multi-agent pipelines, leveraging role specialization to ensure stability and efficiency. For knowledge-intensive domains, works such as Multimodal DeepResearcher \cite{Multimodal_DeepResearcher} structure report generation as a four-stage pipeline ``Researching, Textualization, Planning, and Generation'' to replicate the procedural standards of human experts. Moreover, Agentless \cite{Agentless} avoids complex autonomous interactions, showing that a fixed three stage pipeline of localization, repair, and validation is more effective than dynamic agent planning for software bug fixing. Overall, Fixed Workflow aims to codify human expertise into explicit execution logic. By decomposing tasks into multi-stage pipelines, it moves beyond single-shot invocation and enables the model to handle complex tasks in specific scenarios.

\subsubsection{Automated Workflow}
\label{sec:automated_workflow}
\textbf{Automated Workflow} typically consists of a central orchestrator agent and a series of worker agents. Based on the input task, the orchestrator autonomously constructs the workflow \cite{HuggingGPT, Mindsearch, Cognitive_Kernel_Pro}, or intervenes to adjust the existing topology \cite{PLAN_AND_ACT, Agent_e,Webpilot}. By decomposing the objective into specific execution steps, the orchestrator manages the coordination logic among execution nodes and may also adjust the subsequent path based on real-time feedback. For tool invocation, works such as ControlLLM, Tool-Planner, and ToolChain* employ task decomposition and structured search for tool workflow planning, while frameworks like OctoTools facilitate extensible tool use through standardized tool cards \cite{ControlLLM,ToolPlanner,ToolChainStar,OctoTools}. In terms of domain generalization and multi-agent system scalability, Workforce~\cite{OWL} separates strategic planning from specialized execution. It relies on a domain-agnostic ``Planner Agent'' to decompose tasks, a ``Coordinator Agent'' to orchestrate subtasks, and specialized ``Worker Nodes'' to carry out execution. AORCHESTRA \cite{AORCHESTRA} further extends this flexibility by creating specific configurations for each task node, allowing the orchestrator to instantiate customized subagents on demand rather than relying on a static pool of predefined workers. Overall, Automated Workflow aims to handle open-ended tasks where execution paths cannot be exhaustively listed. By decoupling high-level planning from low-level execution, it adapts to dynamic environments.

\subsubsection{Evolving Workflow}
\label{sec:evolving_workflow}
\textbf{Evolving Workflow} refers to a persistently evolving framework where the topological structure adapts to the environment as tasks accumulate, distinguishing it from patterns that merely change states in response to different tasks. This evolution manifests as long-term modifications to task topology \cite{MUSE, ReCreate, Criticize_Reflect, GPTSwarm}, the autonomous introduction and persistent storage of new tools or roles during runtime \cite{Yunjue_Agent, STEVE}, or behavioral shifts in agents simulating the entire workflow after training \cite{Chain-of-Agents, ReasonRAG}. In this setting, the workflow is no longer a static program, but a dynamic system that expands functional boundaries or optimizes coordination paths through autonomous iteration. Work such as LATM \cite{LATM} enables an agent to act as a tool maker, coding Python functions into a persistent cache to broaden the range of tasks the system can handle. Criticize-Reflect \cite{Criticize_Reflect} and ReCreate \cite{ReCreate} further extend this approach by allowing the system to iteratively rewrite its own organizational prompts and execution logic based on feedback, dynamically altering the entire workflow structure. Alternatively, Chain-of-Agents \cite{Chain-of-Agents} demonstrates that such multi-agent workflows can even be fully internalized into a single model through distillation. Overall, Evolving Workflow aims to break the boundaries between predefined logic and individual tasks. By internalizing empirical insights from execution into the workflow, the agent system develops the ability to improve itself across tasks.

\subsubsection{Summary}

\begin{tcolorbox}[takeaway,title={Takeaway 6.2}]
\begin{itemize}
\item \textbf{Comparative Analysis:} Beyond model training, workflows serve as a method to enhance agent capabilities. In \textbf{Fixed Workflow}, since role functions are relatively decentralized and static, improving overall coordination through training faces challenges. In contrast, \textbf{Automated Workflow} offers a more explicit entry point: by specifically training the central orchestrator, the collective performance of the entire system can be elevated more directly.  

\item \textbf{Future Directions:} \textbf{Evolving Workflow} represents an advanced direction for managing complex and dynamic environments. In highly variable environments, training individual agents in isolation often struggles to meet practical demands. Rather than fragmented training at single nodes, this collective evolution allows the entire system to better adapt to increasingly complex future environments.
\end{itemize}
\end{tcolorbox}

\begin{table*}[!htbp]
\label{tab:trajectory}
\caption{The statistics of Trajectory-Centric Offline Evolution Methods. \protect\Filtering refers to Filtering, \protect\Correction refers to Correction, \protect\Iterative refers to Iterative Refinement.}
\setlength{\tabcolsep}{3pt}
\resizebox{\textwidth}{!}{
\begin{tabular}{lcccccc}
\toprule
\textbf{Name} & \textbf{Domain}          & \textbf{Data Size}  & \textbf{Teacher Model}                             & \textbf{Task Source} & \textbf{Trajectory Source} & \textbf{Refinement} \\ 

\midrule
\rowcolor{gray!10!white} 
\multicolumn{7}{c}{\textbf{\textit{Task Synthesis (\cref{sec:task_synthesis})}}} \\ 
BAGEL~\cite{BAGEL}                                 & Cross-Domain       & 260                  & PaLM-2                                     & Reverse                                 & Sequential Interaction                              & \Filtering                                      \\
ToolACE~\cite{ToolACE}                               & Tool            & 180,000                 & -                                          & Structural                         & Model Simulation                              & \Filtering                                   \\
OS-Genesis~\cite{OS-Genesis}                            & GUI             & 1,000                   & GPT-4o                                     & Reverse                                  & Sequential Interaction                              & \Filtering                                      \\
Insta~\cite{Insta}                                 & GUI             & 150,000                 & Qwen3-235B                                 & Reverse                           & Sequential Interaction                              & \Filtering                                      \\
APIGen-MT~\cite{APIGen-MT}\                             & Tool            & 5,000                   & GPT-4o, DeepSeek V3                        & Structural                         & Model Simulation                              & \Filtering                                      \\
% WebDancer~\cite{WebDancer}                             & Deep Research   & 14,228                 & GPT-4o, QwQ-Plus                           & Transformation                                     & Sequential Interaction                              & \Filtering                                      \\
WebSailor~\cite{WebSailor}                             & Deep Research   & 2,000                   & QwQ-32B                                    & Structural                         & Sequential Interaction                    & \Filtering                                      \\
WebShaper~\cite{WebShaper}                             & Deep Research   & 5,000                   & QwQ-32B                                        & Structural                          & Sequential Interaction                              & \Filtering                                      \\
WebWatcher~\cite{WebWatcher}                            & Deep Research   & 8,000                   & GPT-4o                                     & Transformation                                     & Sequential Interaction                              & \Filtering                                      \\
Cognitive Kernel-Pro~\cite{Cognitive_Kernel_Pro}                  & Deep Research   & 47,314               & GPT-4.1                                    & Reverse                                             & Sequential Interaction                              & \Correction                                      \\
Mobile-Agent-v3.5~\cite{Mobile-Agent-v3.5}                     & GUI             & -                    & GUI-Owl                                    & Structural                         & Sequential Interaction                              & \Iterative                           \\
WebExplorer~\cite{WebExplorer}                           & Deep Research   & 13,000                  & -                                          & Reverse                            & Sequential Interaction                              & \Filtering                                      \\
DeepDive~\cite{DeepDive}                              & Deep Research   & 858                  & Claude-4-Sonnet-T   & Structural                        & Sequential Interaction                              & \Filtering                                      \\
% WebSailor-V2~\cite{WebSailor-V2}                          & Deep Research   & -                    & QwQ-32B                                    & Structural                          & Sequential Interaction                              & \Filtering                                      \\
% AgentScaler~\cite{AgentScaler}                           & Tool            & -                    & -                                          & Structural                          & Model Simulation                              & \Filtering                                      \\
AutoPlay~\cite{AutoPlay}                              & GUI             & 11,500                & GPT-4o, UI-TARS-1.5-7B                     & Reverse                                 & Sequential Interaction                              & \Filtering                                   \\
AgentFounder~\cite{AgentFounder}                          & Cross-Domain    & -                    & -                                          & Transformation                                     & Augmentation                                                 & \Filtering\hspace{-0.4em}\Correction                          \\
WebAggregator~\cite{WebAggregator} & Deep Research   & 6,184                & GPT-4.1                                    & Reverse                                     & Sequential Interaction                              & \Filtering                                      \\
CRMWeaver~\cite{CRMWeaver}                             & Domain-Specific & 3,000                   & GPT-4.1                                    & Structural                          & Sequential Interaction                              & \Filtering                                      \\
WebLeaper~\cite{WebLeaper}                             & Deep Research   & 20,000                  & -                                          & Structural                          & Sequential Interaction                              & \Filtering                                      \\
AgentEvolver~\cite{AgentEvolver}                          & Cross-Domain    & -                    & Qwen-Plus                                  & Reverse                                  & Sequential Interaction                              & \Filtering                                      \\
% MagicAgent~\cite{MagicAgent}                            & Tool            & -                    & -                                          & Structural                          & Model Simulation                              & \Filtering\hspace{-0.4em}\Correction                          \\
VSearcher~\cite{VSearcher}                             & Deep Research   & 1,308                 & Gemini-3-Pro-T                     & Transformation                                     & Sequential Interaction                              & \Filtering                                      \\
OpenSeeker~\cite{OpenSeeker}  & Deep Research   & 11,700  & - & Reverse & Sequential Interaction & - \\
\midrule
\rowcolor{gray!10!white} 
\multicolumn{7}{c}{\textbf{\textit{Trajectory Synthesis (\cref{sec:Trajectory_synthesis})}}} \\ 
ToolCoder~\cite{ToolCoder}                             & Code            & 53,000                  & GPT-3.5-turbo                              & Transformation                                     & Augmentation                                                 & \Filtering                                      \\
ToolBench~\cite{ToolBench}                               & Tool            & 126,486              & GPT-3.5-turbo                              & Transformation                                     & Tree Search                                & \Filtering                                      \\
ToolAlpaca~\cite{ToolAlpaca}  &  Tool   &  3,938 &  
GPT-3.5 & Transformation & Model Simulation& \Filtering \\
AgentTuning~\cite{AgentTuning}                           & Cross-Domain    & 1,866                & GPT-3.5, GPT-4                             & Reuse                                              & Sequential Interaction                              & \Filtering                                      \\
Lingma SWE-GPT~\cite{Lingma_SWE-GPT}                        & Code            & -                    & GPT-4o    & Transformation                                     & Sequential Interaction                              & \Filtering                                      \\
Aguvis~\cite{Aguvis}                                & GUI             & 35,000                & GPT-4o                                     & Transformation                                              &  Augmentation                                               & -                                              \\
% FlowReasoner~\cite{FlowReasoner}                          & Code            & 1,400                 & DeepSeek-R1-671B                           & Reuse                                              & Sequential Interaction                              & -                                              \\
MaskSearch~\cite{MaskSearch}                            & Deep Research   & 58,000                  & Qwen-Max                                  & Transformation                                     & Sequential Interaction                              & \Filtering                                      \\
DreamGen~\cite{DreamGen}                              & Embodied        & 240,000                 & WAN2.1, IDM                                & Human\&Reuse                                       & Model Simulation                              & \Filtering                                   \\
% WebResearcher~\cite{WebResearcher}                         & Deep Research   & -                    & -                                          & Transformation                                     & Sequential Interaction                              & \Filtering                                      \\
% WebWeaver~\cite{WebWeaver}                             & Deep Research   & 6,400                 & -                                          & Transformation                                     & Sequential Interaction                              & \Filtering                                      \\
Tongyi DeepResearch~\cite{Tongyi_DeepResearch}                   & Deep Research   & -                    & -                                          & Structural                         & Sequential Interaction                              & \Filtering                                      \\
% Game-TARS~\cite{Game-TARS}                             & Game            & 20,000                  & Human                                      & Human                                              & Human                                            & \Filtering                                                    \\
AgentFold~\cite{AgentFold}                             & Deep Research   & -                    & GLM-4.5, DeepSeek-V3.1                     &   Reuse                                            & Sequential Interaction                              & \Filtering                                      \\
SIMA2~\cite{SIMA2}                                & Embodied        & -                    & Human                                      & Transformation                              & Human\&Augmentation                                  & \Iterative                           \\
O-Researcher~\cite{O-Researcher}                          & Deep Research   & 3,500                 & -                                          & Transformation                              & Sequential Interaction                              & \Filtering                                      \\
Pixels2Play~\cite{Pixels2Play}                           & Game            & -         & Human                                      & Reverse                                  & Human\&Augmentation                                  & \Filtering                                      \\
ToolACE-MCP~\cite{ToolACE-MCP}                           & Tool            & 15,092               & GPT-4o                                     & Structural                          & Model Simulation                              & -                                              \\
ProAct~\cite{ProAct}                                & Game    & 33,000                  & -                                          & Reuse                                              & Tree Search                                & \Correction                                     \\
OpenResearcher~\cite{OpenResearcher}                        & Deep Research   & 97,000                  & GPT-OSS-120B                               & Reuse                                              & Model Simulation                              & \Filtering                                      \\
HATS~\cite{HATS}                                  & GUI             & 1,000                   & GPT-4o                                     & Reverse                                  & Tree Search                                & \Correction                                 \\
\midrule
\rowcolor{gray!10!white} 
\multicolumn{7}{c}{\textbf{\textit{Trajectory Refinement (\cref{sec:trajectory_refinement})}}} \\ 
Toolformer~\cite{Toolformer}                            & Tool            & 68,000                 & GPT-J-6.7B                               & Transformation                                     & Augmentation                                                 & \Filtering                                      \\
ETO~\cite{ETO}       & Cross-Domain    & 7651                 & GPT-4                                      & Reuse                                              & Sequential Interaction                              & \Iterative                           \\
Agent-FLAN~\cite{Agent-FLAN}                            & Cross-Domain            & 24,703               & GPT-3.5-turbo                              & Transformation                              & Augmentation                                                 & \Filtering                                      \\
Self-Improvement~\cite{Self-Improvement}                      & GUI             & 58                   & Qwen-1.5-72B-Chat                          & Transformation                              & Sequential Interaction                              & \Filtering                                      \\
UI-TARS~\cite{UI-TARS}                               & GUI             & -               & UI-TARS                                    & Human                                              & Human\&Augmentation                                  & \Iterative                           \\
\(\mu\)Code~\cite{muCode}                                 & Code            & -                    & Llama-3.1-8B-I                     & Reuse                                              & Sequential Interaction                              & \Correction                                     \\
SimpleDeepSearcher~\cite{SimpleDeepSearcher}                    & Deep Research   & 871                  & QwQ-32B                                    & Reuse                                              & Sequential Interaction                              & \Filtering                                      \\
EvolveSearch~\cite{EvolveSearch}                          & Deep Research   & 80,000                  & Qwen2.5-7B-I                       & Reuse                                              & Sequential Interaction                              & \Iterative                           \\
GUI-Reflection~\cite{GUI-Reflection}                        & GUI             & 32,951               & Gemini-2.0-Flash, Gemini-2.5-Pro           & Transformation                                     & Augmentation                                                 & \Correction                                     \\
Agent-Reward~\cite{Agent-Reward}                          & Cross-Domain    & 1,136                & GPT-4o                                     & Transformation                                     & Augmentation                                                 & \Filtering                                      \\
WebSynthesis~\cite{WebSynthesis}                          & GUI             & 4,000                   & GPT-4o                                     & Reuse                                              & Model Simulation                              & \Filtering\hspace{-0.4em}\Correction                          \\
TiG~\cite{TiG}                                   & Game            & -                    & DeepSeek-R1                                & Human                                              & Human                                            & \Correction                                     \\
ToolRM~\cite{ToolRM}                                & Tool            & 180,000 & Qwen2.5-32B-I, Granite-20b, etc.                     & Structural                         & Human\&Augmentation                                  & \Filtering                                   \\
Tool-Reflection~\cite{Tool-Reflection}                       & Tool            & 6,000                   & Human                                      & Transformation                              & Model Simulation                              & \Correction                                     \\
% UI-TARS-2~\cite{UI-TARS-2}                             & GUI             & -                    & UI-TARS-2                                  & Transformation                              & Human               & \Iterative                           \\
AgentFrontier~\cite{AgentFrontier}                         & Deep Research   & 12,000                  & -                                          & Structural                         & Sequential Interaction                              & \Iterative                           \\
WebSTAR~\cite{WebSTAR} & GUI & 13,300 & GPT-4o & Reuse & Sequential Interaction & \Filtering \\ 
SynthAgent~\cite{SynthAgent}                            & GUI             & 2,500                   & GPT-4.1                                    & Reverse                                  & Sequential Interaction                              & \Correction                                     \\
SERA~\cite{SERA}                                  & Code            & 200,000                 & GLM-4.5-Air, GLM-4.6                       & Reverse                                  & Sequential Interaction                              & \Filtering\hspace{-0.4em}\Correction                        \\
% UI-Venus-1.5~\cite{UI-Venus-1.5}                          & GUI             & 30,000                  & Qwen3-VL-235B                              & Reverse                                  & Sequential Interaction                              & \Filtering\hspace{-0.4em}\Correction                         \\
TopoCurate~\cite{TopoCurate}  &  Tool  & 2,400 & Claude-4.5-Sonnet & Reuse  & Sequential Interaction & \Filtering \\
\bottomrule
\end{tabular}
}
\end{table*}

\subsection{Trajectory-Centric Offline Evolution}
\label{sec:agent_evo_synthetic_data_generation}
\begin{figure*}[t]
    \centering
    \includegraphics[width=1\textwidth]{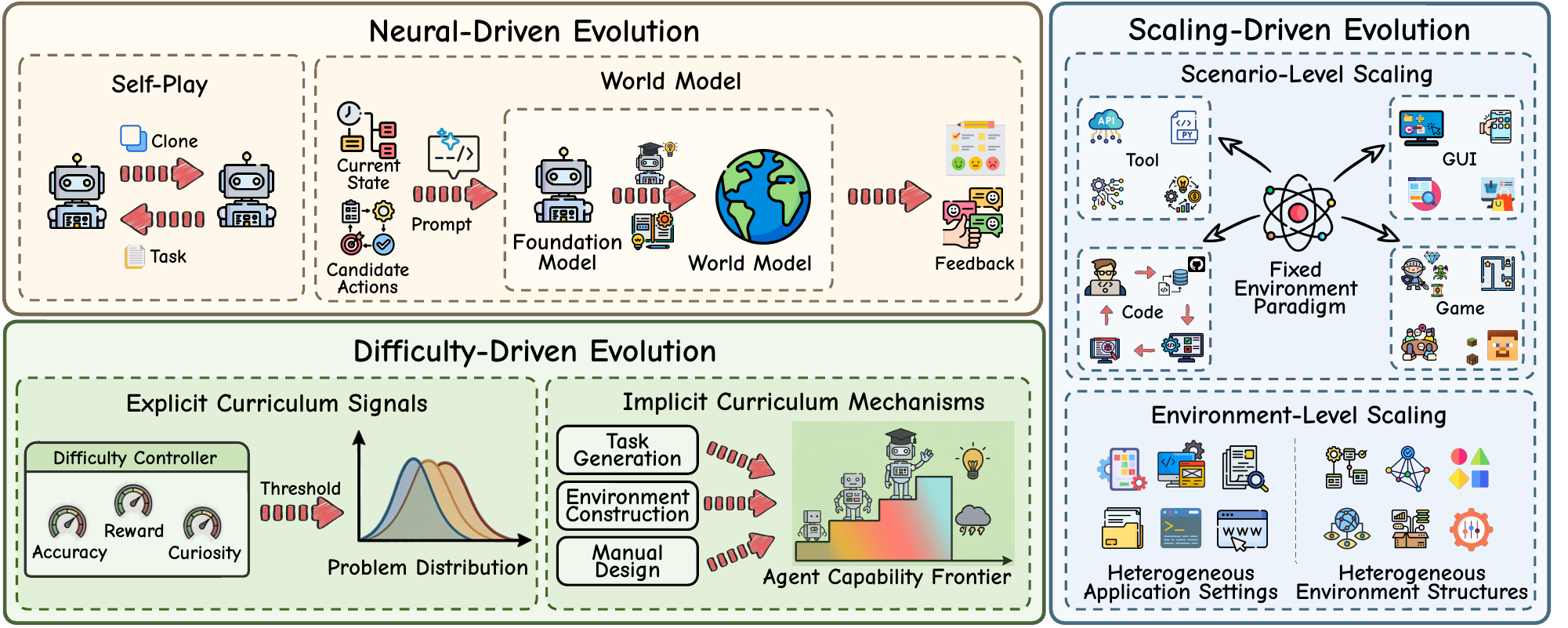}
    \caption{
    \textbf{Overview of environment evolution paradigms.} Existing methods are organized into three categories: Neural-Driven Evolution, which evolves environments through self-play or world models; Difficulty-Driven Evolution, which adapts task difficulty via explicit curriculum signals or implicit curriculum mechanisms; and Scaling-Driven Evolution, which expands environment diversity at the scenario or environment level.
    }
    \label{fig:parametric_synthesis}
\end{figure*}

\textbf{Trajectory-Centric Offline Evolution} primarily refers to agentic SFT, which leverages interaction trajectories to enhance an agent's capabilities. These trajectories may include environment states, action choices, tool calls, external feedback, intermediate reasoning, and even error recovery. In practice, the process of trajectory synthesis usually consists of three stages, as illustrated in Fig.~\ref{fig:parametric_synthesis}: \textbf{(1) Task Synthesis} (\cref{sec:task_synthesis}), which determines what tasks the agent should solve; \textbf{(2) Trajectory Synthesis} (\cref{sec:Trajectory_synthesis}), which turns these tasks into interaction trajectories; and \textbf{(3) Trajectory Refinement} (\cref{sec:trajectory_refinement}), which improves the quality and reliability of the synthesized trajectories before training.

\subsubsection{Task Synthesis}
\label{sec:task_synthesis}
Based on the task construction paradigm, existing approaches can be categorized into three types: \textbf{Resource Transformation}, \textbf{Reverse Synthesis}, and \textbf{Structure-based Synthesis}.

%Reuse&Transformation
\textbf{Resource Transformation} refers to performing transformations based on prior resources. Typical approaches include adding tool-use annotations via LLMs~\cite{Toolformer, ToolCoder, ToolBench}, masking answers~\cite{MaskSearch}, replacing entities with target information~\cite{VSearcher, WebWatcher}, or utilizing existing resources as seeds and prompting LLMs to synthesize novel tasks~\cite{Self-Improvement, Lingma_SWE-GPT}.
% \textbf{Reuse\&Transformation}: Early works often reuse prior datasets\cite{AgentTuning, Aguvis}
% or do some transformation based on existing resources. Many early works use LLMs to annotate existing texts\cite{Toolformer,ToolCoder,ToolBench}. Typically, for example, Toolformer\cite{Toolformer} uses a language model to annotate a large language modeling dataset with potential API calls. Similarly, ToolCoder\cite{ToolCoder} uses GPT-3.5 to add tool usage information into the source code data. Self-Improvement\cite{Self-Improvement} uses WebArena examples as seeds and prompt LLM to synthesize out-of-domain tasks. Lingma SWE-GPT\cite{Lingma_SWE-GPT} constructs task instances by collecting issues, corresponding PRs, and codebases from public GitHub repositories. WebDancer\cite{WebDancer} generates tasks in two ways: synthesizing QA pairs from recursively crawled subpages of knowledgeable websites, and iteratively rewriting simple questions with newly retrieved entity-related information to make them progressively more challenging while preserving the answer. Starting from authoritative sources such as Wikipedia, arXiv, and GitHub, WebWatcher\cite{WebWatcher} recursively traverses hyperlinks to build multi-hop QA pairs, rewrites target entities into visual references, and grounds the resulting VQA queries in authentic web images. VSearcher\cite{VSearcher} generates tasks by starting from a rare Wikidata seed, iteratively replacing selected entities with rarely known information, and then injecting a critical image to form a challenging multimodal browsing task.

\textbf{Reverse Synthesis} denotes a paradigm where agents first explore the environment and then synthesize task instructions from observations or interaction trajectories. This is particularly common in the GUI~\cite{OS-Genesis, AutoPlay, HATS} and Deep Research~\cite{BAGEL, Explorer, Insta, Cognitive_Kernel_Pro, WebExplorer, WebAggregator, SynthAgent, AgentEvolver} domains. To be more specific, WebExplorer~\cite{WebExplorer} leverages LLMs to navigate the web and synthesize corresponding QA pairs. OS-Genesis~\cite{OS-Genesis} performs step-wise interactions in GUI environments and retrospectively derives tasks from the resulting transitions. Going beyond naive exploration, HATS~\cite{HATS} generates tasks through hardness-driven exploration, targeting ambiguous interactions during exploration and validating synthesized instructions through iterative alignment.

% For GUI and Search tasks, a common way is environment exploration\cite{WebExplorer,WebAggregatora, SynthAgent, AgentEvolver}. 
% Explorer\cite{Explorer} uses a bottom-up web trajectory synthesis pipeline that starts with an abstract task proposal and iteratively refines it into a more specific task through web exploration. Insta\cite{Insta} uses a language model task proposer that guides exploration on a website via an initial easy task and then, conditioned on a trajectory that deeply explores the website, creates a harder, grounded task.  AutoPlay\cite{AutoPlay} explores GUI environments with an MLLM explorer, then uses the resulting trajectories to synthesize environment-grounded tasks.

% Typically, BAGEL\cite{BAGEL} let the agent explore the environment without conditioning on any natural language instruction, and then use a language model to relabel the whole trajectory. Further, OS-Genesis\cite{OS-Genesis} first performs step-wise interactions in GUI environments and then retrospectively derives high-quality tasks from the observed state changes. HATS\cite{HATS} generates tasks via hardness-driven exploration with an HD-MCTS policy that targets under-represented yet semantically challenging and informative interactions, then iteratively replays, verifies, and refines the instruction until it is executable and semantically aligned.

\textbf{Structure-based Synthesis} synthesizes tasks based on explicit structural representations, including graphs, trees, schemas, abstract syntax trees (ASTs), etc. Some methods leverage trees~\cite{ToolACE,WebLeaper}, such as ToolACE~\cite{ToolACE}, which builds a hierarchical API context tree and synthesizes APIs through a speciation-adaptation-evolution process. Many instead rely on graphs~\cite{WebSailor,DeepDive,AgentScaler,Tongyi_DeepResearch,CRMWeaver,ToolACE-MCP,MagicAgent,OpenSeeker}, including API dependency graph, knowledge graph, database graph, etc. For instance, DeepDive~\cite{DeepDive} operates over knowledge graphs, constructing QA pairs via random walks and enriching paths with node attributes before using an LLM to obfuscate key cues into complex questions. In contrast, Mobile-Agent-v3.5 ~\cite{Mobile-Agent-v3.5} relies on mobile DAGs annotated by humans, sampling realistic interaction paths and synthesizing user instructions with LLMs/VLMs. Specifically, WebShaper ~\cite{WebShaper} uses set operations over knowledge projections, where an agentic Expander progressively increases task complexity.

\subsubsection{Trajectory Synthesis}
\label{sec:Trajectory_synthesis}
In terms of how trajectories are generated, recent studies mainly adopt four kinds of approaches: \textbf{Trajectory Augmentation}, \textbf{Sequential Interaction}, \textbf{Tree Search}, and \textbf{Model Simulation}.

\textbf{Trajectory Augmentation} includes rewriting~\cite{Agent-FLAN}, enriching~\cite{Aguvis,UI-TARS,SIMA2} and expanding~\cite{AgentFounder} existing trajectories. Typically, Agent-FLAN~\cite{Agent-FLAN} rewrites ReAct-style trajectories into multi-turn chat-style conversations. Aguvis ~\cite{Aguvis} first generates trajectories through automated programs and then augments them with reasoning traces produced by GPT-4o. AgentFounder~\cite{AgentFounder} first generates multi-step reasoning and tool-action sequences, and then expands them into diverse decision paths through higher-order action synthesis.

\textbf{Sequential Interaction} means that trajectories are primarily generated through a single forward interaction chain. Among relevant approaches, the ReAct framework is widely adopted~\cite{AgentTuning, WebDancer, VSearcher}, while some methods further optimize via complex workflow pipelines~\cite{Lingma_SWE-GPT} or multi-agent systems~\cite{FlowReasoner, MaskSearch, WebWeaver, O-Researcher}. As an example, Lingma SWE-GPT~\cite{Lingma_SWE-GPT} synthesizes development trajectories with a three-stage workflow: Repository Understanding, Fault Localization, and Patch Generation. MaskSearch~\cite{MaskSearch} uses a multi-agent system involving a planner, rewriter, and observer. To manage context, Tongyi DeepResearch~\cite{Tongyi_DeepResearch} summarizes context into an evolving report to preserve essential information. Moreover, AgentFold~\cite{AgentFold} adopts a more fine-grained context-folding strategy, performing either granular condensation of the Latest Interaction or deep consolidation of it with prior summaries into a coarser state summary.

\textbf{Tree Search} denotes a class of work that identifies higher-quality trajectories using tree search algorithms such as depth-first search (DFS) and Monte Carlo Tree Search (MCTS). In particular, ToolBench~\cite{ToolBench} uses DFS to search for a valid solution path to avoid being trapped in a single ReACT trajectory. ProAct~\cite{ProAct} generates trajectories by running MCTS from the current state, sampling both optimal and suboptimal/dead-end paths.

\textbf{Model Simulation} is another type of method that generates trajectories in simulated environments through components including simulated users~\cite{APIGen-MT, AgentScaler,ToolACE-MCP}, tools~\cite{ToolAlpaca,ToolACE,ToolRM,MagicAgent}, and world models~\cite{DreamGen, WebSynthesis}. Specifically, ToolACE-MCP~\cite{ToolACE-MCP} generates multi-turn dialogue trajectories through role-based simulation, with planner, user, assistant, and tool roles simulated by LLMs. ToolRM~\cite{ToolRM} uses an LLM to simulate the tool’s error feedback. DreamGen~\cite{DreamGen} generates neural trajectories by prompting a robot-adapted video world model. 

\subsubsection{Trajectory Refinement}
\label{sec:trajectory_refinement}
\textbf{Trajectory Refinement} determines which trajectories are ultimately incorporated into the training set. We broadly classify it into three methods, namely, \textbf{Filtering}, \textbf{Correction}, and \textbf{Iterative Refinement}.

\textbf{Filtering} removes low-quality samples using rules~\cite{Toolformer,Lingma_SWE-GPT,SimpleDeepSearcher, WebResearcher, WebLeaper}, structure-based scoring~\cite{TopoCurate}, LLM-as-judge~\cite{OS-Genesis, Insta, MaskSearch, WebSTAR, Agent-Reward, AutoPlay}, or model self-scoring~\cite{Self-Improvement}. In detail, Lingma SWE-GPT~\cite{Lingma_SWE-GPT} applies rejection sampling based on two criteria: fault localization accuracy and patch similarity. TopoCurate~\cite{TopoCurate} computes three structure-aware metrics on an interaction topology to filter trajectories.  WebSTAR~\cite{WebSTAR} performs step-level filtering using a grading model. Self-Improvement~\cite{Self-Improvement} filters out trajectories in which the model self-detected a failure, producing a refusal.

\textbf{Correction} chooses to correct~\cite{GUI-Reflection,AgentFounder}, rewrite~\cite{ToolRM,UI-Venus-1.5,HATS}, relabel~\cite{muCode} or edit~\cite{SynthAgent} erroneous trajectories instead of simply discarding them. 
For example, GUI-Reflection~\cite{GUI-Reflection} identifies the first incorrect action in a trajectory and generates post-error reflection and correction actions for it. WebSynthesis~\cite{WebSynthesis} leverages GPT-4 to synthesize reflections for failed branches, tracing back to the common ancestor node and identifying "go\_back" as the corrective action for rollback trajectory construction. $\mu$Code~\cite{muCode} relabels the collected data with the optimal solutions selected by a local search expert. SynthAgent~\cite{SynthAgent} applies Remove, Reorder, Drop, or Keep edits to trajectories to eliminate task-irrelevant, redundant, or misaligned actions.

\textbf{Iterative Refinement} connects generation, filtering, verification, correction, and retraining into a closed loop, allowing the refinement process to directly shape the data distribution of the next iteration~\cite{ETO, UI-TARS, EvolveSearch, Mobile-Agent-v3.5, UI-TARS-2, AgentFrontier}. Among these, ETO~\cite{ETO} updates the agent policy through an iterative cycle of trajectory collection and training. EvolveSearch~\cite{EvolveSearch} iteratively alternates between RL exploration and SFT optimization, using high-quality and diverse rollouts collected during RL as the SFT training set. AgentFrontier~\cite{AgentFrontier} generates complex-reasoning data within the LLM’s Zone of Proximal Development (ZPD), targeting tasks situated at the model’s capability frontier and, as the model learns, its ZPD advances, enabling a continuously adaptive curriculum.

\subsubsection{Summary}
\begin{tcolorbox}[takeaway,title={Takeaway 6.3}] %
\begin{itemize}
\item \textbf{Comparative Analysis:} For task synthesis, current research mainly focuses on making tasks more complex and realistic. Thus, methods have evolved from corpus transformation to reverse and structure-based synthesis. Trajectories are expected to better capture high-level abilities, such as planning, reflection, and recovery, resulting in increasingly sophisticated workflows. At the same time, maintaining verifiable quality boundaries while scaling up requires refinement strategies to shift from simple filtering toward a closed-loop pipeline that includes generation, verification, correction, and retraining.

\item \textbf{Future Directions:} It is likely that the next competitive edge will no longer lie in simply scaling up data, but in several fundamental directions: how to improve the alignment between synthesized trajectories and training objectives, how to build low-cost but high-trust verification mechanisms for long-horizon trajectories, how to turn failure and recovery processes into stable training assets, and how to prevent models from continually reinforcing their own biases in a self-bootstrapping data flywheel. Whoever can solve these problems more effectively will be more likely to truly advance trajectory synthesis from ``sample production'' to ``capability'' engineering.”

\end{itemize}
\end{tcolorbox}

% Nemotron-Cascade-2

\renewcommand{\arraystretch}{1.5} 
\begin{table*}[]
\label{tab:rl}
\caption{The statistics of Exploration-Centric Online Evolution Methods. \protect\Result refers to outcome reward, \protect\Process refers to process reward, \protect\Efficient refers to efficient reward and \protect\Format refers to format reward.}
\resizebox{\textwidth}{!}{
\begin{tabular}{lcccccc}
\toprule
\textbf{Name} &
 \textbf{Task} &
 \textbf{RL Algorithm} &
  \textbf{Model} &
  \textbf{Training Data} &
  \textbf{Reward} \\ \midrule

\rowcolor{gray!10!white} \multicolumn{6}{c}{\textbf{\textit{Reasoning Structure Design (\cref{sec:reasoning_structure})}}} \\
DeepRetrieval \cite{Deepretrieval} &
  DeepResearch &
  PPO &
  Qwen2.5-3B-I, LLaMA-3.2-3B, etc. &
  NQ, HotpotQA, etc. &
  \Result\hspace{-0.4em}\Format \\

Search-R1 \cite{Search-r1} &
  DeepResearch &
  PPO &
  Qwen2.5-3/7B, etc. &
  NQ, HotpotQA &
  \Result  \\

ReSearch\cite{ReSearch_2} &
  DeepResearch &
  GRPO &
  Qwen2.5-7/32B, etc. &
  MuSiQue &
  \Result\hspace{-0.4em}\Format \\

AutoRefine \cite{AutoRefine} &
  DeepResearch &
  GRPO &
  Qwen2.5-3/7B, &
  Natural Questions, HotpotQA &
  \Result\hspace{-0.4em}\Process \\

SEEA-R1 \cite{SEEA-R1} &
  Embodied &
  Tree-GRPO &
  Qwen2.5-VL-7B-I &
  ALFWorld, Customize &
  \Result \\

M3-Agent \cite{M3-Agent} &
  Image \& Video &
   DAPO &
  Qwen2.5-Omni-7B, Qwen3-32B, etc. &
  Customize &
  \Result \\

Video-Thinker \cite{Video-Thinker} &
  Image \& Video &
  GRPO &
  Qwen2.5-VL-7B-I &
  ActivityNet, YouCook2, etc. &
  \Result\hspace{-0.4em}\Format \\

\midrule  
\rowcolor{gray!10!white} \multicolumn{6}{c}{\textbf{\textit{Training Reward Design (\cref{sec:reward_shaping})}}} \\
  
MaskSearch \cite{MaskSearch} &
  DeepResearch &
  DAPO &
  Qwen2.5-1.5/3B, etc. &
  Wikipedia, HotpotQA, etc. &
  \Result\hspace{-0.4em}\Format  \\

R1-Searcher \cite{R1-Searcher} &
  DeepResearch &
  Reinforce++ &
  Qwen-2.5-7B, Llama-3.1-8B-I&
  HotpotQA, 2WikiMultiHopQA &
  \Result\hspace{-0.4em}\Format\hspace{-0.4em}\Process \\

ToolRL \cite{ToolRL} &
  Tool &
  GRPO &
  Qwen2.5-1.5/3B-I, &
  ToolACE, xLAM, etc. &
  \Result\hspace{-0.4em}\Format\hspace{-0.4em}\Process \\

OTC-PO \cite{OTC-PO} &
  Cross-Domain &
  PPO, GRPO &
  Qwen2.5-3/7B &
  NQ, HotpotQA, etc. &
  \Result\hspace{-0.4em}\Format\hspace{-0.4em}\Process\hspace{-0.4em}\Efficient \\

Tool-N1 \cite{Tool-N1} &
  Tool &
  GRPO &
  Qwen2.5-7/14B-I &
  xLAM, ToolACE &
  \Result\hspace{-0.4em}\Format \\

ARTIST \cite{ARTIST} &
  Cross-Domain &
  GRPO &
  Qwen2.5-7/14B-I &
  NuminaMath, BFCL v3 &
  \Result\hspace{-0.4em}\Format\hspace{-0.4em}\Process \\

VLA-RL \cite{VLA-RL} &
  Embodied &
  PPO &
  OpenVLA-7B &
  LIBERO &
  \Result\hspace{-0.4em}\Process \\

VRAG-RL \cite{VRAG-RL} &
  Image \& Video &
  GRPO &
  Qwen2.5-VL-3/7B-I &
  Customize &
  \Result\hspace{-0.4em}\Format\hspace{-0.4em}\Efficient \\

R-Search \cite{R-Search} &
  DeepResearch &
  PPO, GRPO &
  Qwen2.5-3/7B-I &
  2WikiMultiHopQA &
  \Result\hspace{-0.4em}\Format\hspace{-0.4em}\Process \\

Chain-of-Agents \cite{Chain-of-Agents} &
  Cross-Domain &
  DAPO &
  Qwen2.5-3/7/32B-I &
  NQ, HotpotQA, etc. &
  \Result\hspace{-0.4em}\Format \\

Video-MTR \cite{Video-MTR} &
  Image \& Video &
  PPO &
  Qwen2.5-VL-7B &
  NExT-GQA, QVHighlights &
  \Result\hspace{-0.4em}\Format\hspace{-0.4em}\Process \\

Agent-R1 \cite{Agent-R1} &
  Cross-Domain &
  PPO, GRPO, etc. &
  Qwen2.5-3B-I &
  HotpotQA, 2WikiMultiHopQA &
  \Result\hspace{-0.4em}\Process \\

ToolOrchestra \cite{ToolOrchestra} &
  Tool &
  GRPO &
  Qwen3-8B &
  ToolScale, GeneralThought-430K &
  \Result\hspace{-0.4em}\Efficient\hspace{-0.4em}\Process \\

VideoMem \cite{VideoMem} &
  Image \& Video &
  PRPO &
  Qwen3-VL-8B&
  VideoMarathon, LLaVA-Video-178K &
  \Result\hspace{-0.4em}\Process\hspace{-0.4em}\Efficient \\

GDPO \cite{GDPO} &
  Tool &
  GDPO &
  Qwen2.5-1.5/3B-I &
  ToolACE, xLAM, etc. &
  \Result\hspace{-0.4em}\Format \\

FlowSteer \cite{FlowSteer} &
  Tool &
  CWRPO &
  Qwen3-8B &
  GSM8K, MATH, etc. &
  \Result\hspace{-0.4em}\Format\hspace{-0.4em}\Process \\

IntentRL \cite{IntentRL} &
  DeepResearch &
  GRPO &
  Qwen2.5-7B &
  DeepResearch Bench &
  \Result\hspace{-0.4em}\Format\hspace{-0.4em}\Process\hspace{-0.4em}\Efficient \\

SHARP \cite{SHARP} &
  Tool &
   GRPO &
  LLaMA-3.1-8B, Qwen3-8B &
  MuSiQue &
  \Result\hspace{-0.4em}\Process \\

\midrule

\rowcolor{gray!10!white} \multicolumn{6}{c}{\textbf{\textit{Training Algorithm Optimization (\cref{sec:algorithmic_optimization})}}}\\

RAGEN \cite{RAGEN} &
  Cross-Domain & StarPO &
  Qwen2.5-0.5/3B-I &
  WebShop &
  \Result\hspace{-0.4em}\Format  \\
ZeroSearch \cite{ZeroSearch} &
  DeepResearch & ZeroSearch &
  Qwen2.5-3/7B, etc. &
  NQ, HotpotQA &
  \Result \\
GiGPO \cite{GiGPO} &
  Cross-Domain & GiGPO &
  Qwen2.5-1.5/3/7B-I &
  ALFWorld, WebShop, etc. &
  \Result\hspace{-0.4em}\Process \\

WebAgent-R1 \cite{WebAgent-R1} &
  GUI & M-GRPO &
  Qwen-2.5-3B,Llama-3.1-8B &
  WebArena &
  \Result \\
EvolveSearch \cite{EvolveSearch} &
  DeepResearch &  GRPO &
  Qwen2.5-7B-I &
  NQ, HotpotQA, etc. &
  \Result\hspace{-0.4em}\Format \\
Embodied Planner-R1 \cite{Embodied_Planner-R1} &
  Embodied & IPO &
  Qwen2.5-7B-I &
  ALFWorld, ScienceWorld &
  \Result \\
SPIRAL \cite{SPIRAL} &
  Game & SPIRAL &
  Qwen3-4/8B, etc. &
  Customize &
  \Result \\
MobileGUI-RL \cite{MobileGUI-RL} &
  GUI & 
  MobGRPO & 
  Qwen2.5-VL-7/32B &
  AndroidWorldAvd &
  \Result\hspace{-0.4em}\Efficient \\
ARPO \cite{ARPO} &
  Cross-Domain &
  ARPO &
  Qwen2.5-3B-I, Llama3.1-8B-I, etc. &
  Tool-Star, STILL, etc. &
  \Result\hspace{-0.4em}\Format\hspace{-0.4em}\Process \\
Thinking With Videos \cite{Thinking_With_Videos} &
  Image \& Video &
  DGRPO &
  Qwen2.5-VL-7B &
  MTVR-CoT, MTVR-RL, etc. &
  \Result\hspace{-0.4em}\Format\hspace{-0.4em}\Process \\
ComputerRL \cite{ComputerRL} &
  GUI &
  ComputerRL &
  GLM-4-9B-0414, GLM-4.1V-9B-T &
  Customize &
  \Result \\
ExIt \cite{ExIt} &
  Cross-Domain &
  ExIt &
  Llama-3.2-3B-I, Qwen2.5-7B-I, etc. &
  NuminaMath, BFCL-v3, etc. &
  \Result \\
AgentGym-RL \cite{AgentGym-RL} &
  Cross-Domain &
   ScalingInter-RL &
  Qwen2.5-3/7B &
  WebArena, NQ, etc. &
  \Result \\
UI-S1 \cite{UI-S1} &
  GUI &
  SOPO &
  Qwen2.5VL-3/7/32B &
  AndroidControl-Train, Amex &
  \Result\hspace{-0.4em}\Format\hspace{-0.4em}\Process \\
SPEAR \cite{SPEAR} &
  Cross-Domain &
  SPEAR &
  Qwen2.5-1.5/7B-I, etc. &
  ALFWorld, WebShop, etc. &
  \Result\hspace{-0.4em}\Format\hspace{-0.4em}\Process \\
AgentRL \cite{AgentRL} &
  Cross-Domain &
  AgentRL &
  Qwen2.5-3/7B-I, etc. &
  ALFWorld, WebShop, etc. &
  \Result\hspace{-0.4em}\Format \\
Stronger-MAS \cite{STRONGER-MAS} &
  Cross-Domain &
  AT-GRPO &
  Qwen3-1.7/8B &
  Sudoku, Sokoban, etc. &
  \Result\hspace{-0.4em}\Process \\
VAGEN \cite{VAGEN} &
  Cross-Domain &
  PPO &
  Qwen2.5-VL-3B &
  Sokoban, FrozenLake, etc. &
  \Result\hspace{-0.4em}\Format\hspace{-0.4em}\Process \\

GTR-Turbo \cite{GTR-Turbo} &
  Cross-Domain &
  GTR &
  Qwen2.5-VL-7B, Qwen3-VL-8B &
  Points24, ALFWorld &
  \Result\hspace{-0.4em}\Process \\

SeeUPO \cite{SeeUPO} &
  Tool &
  SeeUPO &
  Qwen3-14B, Qwen2.5-14B-I &
  AppWorld, BFCL v4 &
  \Result\hspace{-0.4em}\Process\\

HGPO \cite{HGPO} &
  Cross-Domain &
  HGPO &
  Qwen2.5-1.5/7B-I & % , Qwen2.5-B-I
  ALFWorld, WebShop &
  \Result \\

DeepSWE \cite{DeepSWE} &
  Code &
  GRPO++ &
  Qwen3-32B &
  R2E-Gym &
  \Result\hspace{-0.4em}\Efficient  \\

 \bottomrule

\end{tabular}
}

\end{table*}

\subsection{Exploration-Centric Online Evolution}
\label{sec:agent_evo_reinforcement_learning_optimization}
\textbf{Reinforcement learning} is a crucial method for agent evolution \cite{PPO, GRPO, RLVE}. Through reinforcement learning, the model can not only enhance its capabilities but also mitigate catastrophic forgetting, thereby achieving better performance across multiple tasks \cite{SayCan, ReTool, GUI-R1, ARLArena, CM2, SimpleTIR, GEM, WebRL, WebArbiter}. By modifying the model's reasoning paradigms, setting up reward mechanisms, or improving training algorithms, reinforcement learning encourages the model to achieve targeted improvements in specific tasks or behaviors \cite{Agent-RRM, MetaClaw, ZeroGUI, DistRL}. We summarize the reinforcement learning training process into three parts: \textbf{(1) Reasoning Structure Design}: Modifying the model's reasoning structure in downstream tasks and designing corresponding reinforcement learning algorithms. \textbf{(2) Training Reward Design}: Designing more effective reward signals to encourage the model to explore and optimize more efficiently. \textbf{(3) Training Algorithm Optimization}: Optimizing the training algorithm to enhance the model's stability and effectiveness in complex tasks.

\subsubsection{Reasoning Structure Design}
\label{sec:reasoning_structure}
\textbf{The research on inference structure design primarily focuses on how agents organize decision-making and reasoning processes in complex tasks}. In recent years, various methods have been proposed to enhance task-solving capabilities by designing specialized inference structures, often integrated with the training process. For example, Search-R1 \cite{Search-r1} introduces special tags in the reasoning process to differentiate between the thinking, search, feedback, and answer processes, while ReSearch \cite{ReSearch_2} and DeepRetrieval \cite{Deepretrieval} enhance the model's reasoning capability by incorporating format correctness rewards to ensure the accuracy of generated labels and the correct reasoning format, to promote the effectiveness of structured output. 

As research progresses, more works are designing complex inference structures or workflows to further improve model performance. For instance, AutoRefine \cite{AutoRefine} introduces an explicit refinement step into the inference loop. In this paradigm, after obtaining retrieved documents, the model must first filter, refine, and organize the evidence, extracting key facts. Based on this refined information, the model then decides whether to conduct another search or generate the final answer.

Additionally, in multimodal and embodied systems, Video-Thinker \cite{Video-Thinker} enables the model to think about videos by integrating time localization and content understanding into the reasoning process using special tokens. M3-Agent \cite{M3-Agent} enhances the handling of audiovisual tasks by incorporating external memory tools to establish connections between roles and memories before the reasoning process. ITP \cite{ITP} introduces world models to simulate environmental dynamics, supporting more complex reasoning tasks.

\subsubsection{Training Reward Design}
\label{sec:reward_shaping}
\textbf{The core of reward design lies in modeling refined signals to guide agents in exploring and identifying optimal strategies}. The reward mechanism has evolved from early, purely result-oriented systems that focused on the correctness of the outcome to a multi-dimensional, task-specific optimization framework.

In terms of efficiency optimization, researchers have introduced a cost penalty mechanism to reduce the agent's over-reliance on tools or redundant searches. For example, OTC-PO \cite{OTC-PO} penalizes redundant calls by rewarding the product of correctness and efficiency, ReasonRAG \cite{ReasonRAG} drives minimalist reasoning by applying a path-length penalty factor, and VRAG-RL \cite{VRAG-RL} and VideoMem \cite{VideoMem} impose sparsity constraints on the RAG process or memory module, respectively, through retrieval time rewards and memory capacity penalties, aiming to solve problems with minimal cost.

In the domain of goal alignment and stability, many methods combine rewards to help models better understand tasks. For example, IntentRL \cite{IntentRL} combines answer reward, format compliance, and repetition penalties, while SHARP \cite{SHARP} uses the Shapley Value for marginal contribution attribution, effectively solving the credit assignment problem. GDPO \cite{GDPO} introduces a decoupled normalization mechanism, independently processing accuracy, format, and length rewards to prevent signal collapse. At the same time, FlowSteer \cite{FlowSteer} employs a conditional release strategy. Final answer rewards are activated only when intermediate metrics of diversity are satisfied. Similarly, Video-MTR \cite{Video-MTR} uses this strategy, where process answer rewards are triggered only when final answer validity meets the required criteria. For multi-agent frameworks like Chain-of-Agents \cite{Chain-of-Agents}, the correctness reward and format reward are composed to formulate the final reward.

In terms of process supervision, format rewards have become a universal constraint across tasks, while VLA-RL \cite{VLA-RL} uses a robot process reward model (RPRM), which provides high-density pseudo-reward signals through visual milestones.

\subsubsection{Training Algorithm Optimization}
\label{sec:algorithmic_optimization}
\textbf{The optimization design of reinforcement learning algorithms focuses on improving training stability and sample efficiency in complex tasks through innovations in underlying architectures}.
In the area of credit allocation, DigiRL \cite{DigiRL} proposes an advantage-weighted RL with advanced advantage estimators to account for stochasticity, coupled with an automatic curriculum that derives maximal learning signals. To address the challenge of sparse rewards due to long-range interactions, GiGPO \cite{GiGPO} introduces Anchor State Grouping, which aggregates different actions originating from the same state into step-level groups for optimization. The contextual consistency cluster of historical contexts and adaptive weighting in HGPO \cite{HGPO}, along with the world modeling reward and Bi-layer Generalized Advantage Estimation mechanism in VAGEN \cite{VAGEN}, enables precise reward propagation from macro-level episode rewards to micro-level step-level rewards. SPIRAL \cite{SPIRAL}, through Role-Conditioned Advantage Estimation (RAE), effectively mitigates asymmetric variance risk in multi-agent games.

For objective function optimization, AEPO \cite{AEPO} introduces entropy-balanced rollout mechanism and optimization to prevent traditional algorithms from excessively suppressing high-entropy exploration signals, while CEMs \cite{CEMs} impose an information-theoretic metric term to encourage the model to learn efficient communication topologies.

Regarding sampling structures, RAGEN \cite{RAGEN} retains trajectories with high reward variance during the sampling process, avoiding overfitting to specific samples. Stronger-MAS \cite{STRONGER-MAS} employs tree-structured sampling to ensure statistical validity deep within conversations, while WebSailor \cite{WebSailor} improves gradient efficiency by dynamically replicating non-zero variance samples, and AgentRL \cite{AgentRL} uses cross-policy sampling to enable stabilization. 

Finally, in terms of training mechanisms, ComputerRL \cite{ComputerRL} introduces the novel concept of alternating RL and supervised fine-tuning updates, forcibly injecting successful experiences to tackle the issue of entropy collapse in online RL during later stages. This establishes a systematic optimization paradigm that evolves from local credit alignment to global strategy robustness.

\subsubsection{Summary}
\begin{tcolorbox}[takeaway,title={Takeaway 6.4}] % RL
\begin{itemize}
\item \textbf{Comparative Analysis:}  The design of the reinforcement learning process remains one of the key challenges in post-training. Current research focuses on improving more efficient reasoning paradigms, designing fine-grained reward signals, enhancing training stability, and enabling more effective exploration. Among these, training algorithm optimization is particularly critical, as it is more tightly coupled with parameter updates and thus offers a more general and scalable solution.

\item \textbf{Future Directions:} Looking ahead, there are two promising directions. One is to integrate new concepts and capabilities into the training process, such as experience, skills, and personalization, enabling more adaptive and intelligent agents. The other is to develop more general-purpose algorithms that can be reliably deployed in real-world systems, bridging the gap between research and practical applications.
\end{itemize}
\end{tcolorbox}

%%%%%%%%%%%%
%%%%%%%%%%%%
%%%%%%%%%%%%
%%%%%%%%%%

%%%%%%%%%%%%%%%%%%%%%%%%%%%%%%%

%%%%%%%%%%%%%%%%%%%%%%%%%%%%%%%%%%%%%%%%%%%%%
%%%%%%%%%%%%%%%%%%%%%%%%%%%%%%%%%%%%%%%%%%%%%
\section{Environment Evolution}\label{sec:env_evo}
% Environment evolution focuses on how training environments can be adapted to improve agent capability.
Environment evolution focuses on how training environments are progressively evolved to support the continuous improvement of agent capability.
Existing approaches can be categorized into three paradigms according to how the environment is evolved:
% \textbf{(1) Parameter-Driven Evolution} (\cref{sec:param_evo}) focuses on modeling environment dynamics through world models, enabling agents to simulate interactions before execution;
\textbf{(1) Neural-Driven Evolution} (\cref{sec:neural_evo}) instantiates the environment as a learned neural model, and evolves the environment by optimizing this model to provide adaptive interactions or simulated transitions for agent learning;
\textbf{(2) Difficulty-Driven Evolution} (\cref{sec:diff_evo}) regulates environment complexity to match the agent's capability through curriculum learning;
\textbf{(3) Scaling-Driven Evolution} (\cref{sec:scaling_evo}) expands the environment distribution itself by increasing scenario diversity or introducing new environment structures.
This categorization highlights different ways in which environments can be adapted to support agent learning, and provides a structured view of the design space of environment evolution.
\begin{table*}
\caption{The statistics of Environment Evolution Methods. \textbf{Str} stands for structured text, \textbf{Err} for error messages, and \textbf{NA} means not applicable or unspecified.}
\resizebox{\textwidth}{!}{
\begin{tabular}{lccccc}
\toprule
\textbf{Name} &
\textbf{Trait} &
\textbf{Domain} &
\textbf{Feedback} &
\textbf{Evo Criterion} &
\textbf{Process Reward} \\
\midrule

\rowcolor{gray!10!white}
\multicolumn{6}{c}{\textbf{\textit{Neural-Driven Evolution (\cref{sec:neural_evo})}}} \\

Absolute zero~\cite{Absolute_zero} & Self-Play & Code & Str & Success Rate & \No \\
Self-Challenging~\cite{Self-Challenging} & Self-Play & Tool & Str, Tool Response & NA & \No \\
Vision-zero~\cite{Vision-zero} & Self-Play & Chart, Vision & Image, Str & Accuracy, N/A Rate & \No \\
SSR~\cite{SSR} & Self-Play & Code & Str, JSON & Success Rate & \No \\
Active Zero~\cite{Active_Zero} & Self-Play & Vision, Math & Image, Str & Solver Uncertainty, Repetition Penalty & \No \\

WebDreamer~\cite{WebDreamer} & World Model & GUI & Image, Str & NA & \No \\
UI-Simulator~\cite{UI-Simulator} & World Model & GUI & Str, List & Teacher-forcing Loss & \No \\
WebEvolver~\cite{WebEvolver} & World Model & GUI & Str, List & NA & \No \\
Code2World~\cite{Code2World} & World Model & GUI & Str, Image & NA & \No \\
WAC~\cite{WAC} & World Model & GUI & Str & NA & \No \\
WebWorld~\cite{WebWorld} & World Model & GUI & Str, List & NA & \No \\

\midrule
\rowcolor{gray!10!white}
\multicolumn{6}{c}{\textbf{\textit{Difficulty-Driven Evolution (\cref{sec:diff_evo})}}} \\

POET~\cite{POET} & Implicit Signals & Robotics & State & NA & \Yes \\
PAIRED~\cite{PAIRED} & Explicit Signals & Game & State & Regret & Partial \\
DCD~\cite{DCD} & Explicit Signals & Game, Robotics & State, Image & Regret & Partial \\
ACCEL~\cite{ACCEL} & Explicit Signals & Game, Robotics & State & Regret & Partial \\
MAESTRO~\cite{MAESTRO} & Explicit Signals & Game & State & Regret & Partial \\
ReMiDi~\cite{ReMiDi} & Explicit Signals & Game & State & Regret & Partial \\
EnvGen~\cite{EnvGen} & Explicit Signals & Game & Image & Success Rate & \No \\
AgentGen~\cite{AgentGen} & Implicit Signals & Embodied & Str & NA & \No \\
DataEnvGym~\cite{DataEnvGym} & Explicit Signals & Math, Code, Vision, Tool & Str & Student Feedback & \No \\
Eurekaverse~\cite{Eurekaverse} & Explicit Signals & Embodied & NA & Success Rate & \Yes \\
Environment Tuning~\cite{Environment_Tuning} & Implicit Signals & Tool & Hint, Str, Err & Curriculum Schedule & \Yes \\
SEC~\cite{SEC} & Implicit Signals & Math, Reasoning & Str & NA & \No \\
Reasoning Core~\cite{Reasoning_Core} & Implicit Signals & Reasoning, Planning, Logic & Str, Err & External Curriculum Policy & \No \\
DreamGym~\cite{DreamGym} & Implicit Signals & GUI, Embodied & Str & NA & \No \\
RLVE~\cite{RLVE} & Explicit Signals & Math, Algorithm & Str & Success Rate & \No \\
CuES~\cite{CuES} & Explicit Signals & Tool, GUI & Str, Err & Curiosity & \No \\
GenEnv~\cite{GenEnv} & Explicit Signals & Tool, Embodied, Reasoning & Str, Err & $\alpha$-Curriculum Reward & \No \\
SCALER~\cite{SCALER} & Explicit Signals & Math & Str, Err & Accuracy & \No \\

\midrule
\rowcolor{gray!10!white}
\multicolumn{6}{c}{\textbf{\textit{Scaling-Driven Evolution (\cref{sec:scaling_evo})}}} \\

FTRL~\cite{FTRL} & Scenario-Level & Tool & Str, Err & NA & \Yes \\
ARE~\cite{ARE} & Environment-Level & GUI, Tool & Str, Tool Response & NA & \No \\
AgentScaler~\cite{AgentScaler} & Scenario-Level & Tool & Str, Err & NA & \No \\
AutoEnv~\cite{AutoEnv} & Environment-Level & Reasoning, Game & Str & NA & \No \\
AutoForge~\cite{AutoForge} & Scenario-Level & Tool & Str, Err & NA & \No \\
EnvScaler~\cite{EnvScaler} & Scenario-Level & Tool & Str, Err & NA & \Yes \\
InfiniteWeb~\cite{InfiniteWeb} & Scenario-Level & GUI & Image, Str & NA & \Yes \\
Agent World Model~\cite{Agent_world_model} & Scenario-Level & Tool & Dict, Str, Err & NA & \Yes \\
AutoWebWorld~\cite{AutoWebWorld} & Scenario-Level & GUI & Image, Str & NA & \Yes \\

\bottomrule
\end{tabular}}
\end{table*}
\subsection{Neural-Driven Evolution}\label{sec:neural_evo}
Neural-Driven environment evolution treats the environment itself as a learned neural model.
In this paradigm, the environment is no longer a fixed external simulator or a manually specified interaction process, but a trainable model that directly embodies either self-generated interactive dynamics or learned environment dynamics.
Accordingly, environment evolution is realized through the optimization of this model, so that the environment can adapt together with agent learning by providing adaptive interactions or simulated transitions.
Existing approaches in this direction can be broadly divided into two categories.
The first relies on \textbf{self-play}, where the environment is effectively instantiated by the agent itself through self-generated tasks, challenges, or role-based interactions~\cite{Absolute_zero,Self-Challenging,Vision-zero,SSR,Active_Zero}.
The second relies on \textbf{world models}, where the environment is instantiated as a separately learned simulator that approximates environment dynamics and provides environment-like feedback~\cite{Code2World,WebEvolver,WebWorld,WebDreamer,UI-Simulator,WAC}.
\subsubsection{Self-Play}\label{sec:self-play}
Self-play constructs an effective training environment from the agent itself or its role-specialized variants.
Through self-generated tasks, the agent not only learns within the environment but also participates in shaping the environment it learns from, making environment evolution endogenous to self-play.
Absolute zero~\cite{Absolute_zero} represents the clearest form of this paradigm: a unified model simultaneously plays the roles of proposer and solver, generating tasks that maximize its own learning progress and then improving by solving them.
Self-Challenging~\cite{Self-Challenging} extends this paradigm to tool-use settings, where the same model first acts as a challenger to synthesize verifiable tasks and then switches to an executor role to learn from them.
Beyond direct task generation, some methods realize self-play through structured role interaction.
Vision-zero~\cite{Vision-zero} formulates self-improvement as a gamified visual reasoning process, where role-based interactions over visual inputs create increasingly informative training challenges.
SSR~\cite{SSR} grounds self-play in software engineering, allowing a single model to alternate between bug injection and bug fixing so that increasingly realistic coding tasks emerge from the self-play loop itself.
Active Zero~\cite{Active_Zero} further generalizes this formulation by introducing multiple co-evolving roles, such as Searcher, Questioner, and Solver, enabling the agent to actively retrieve frontier examples and construct adaptive visual reasoning tasks.
\subsubsection{World Model}\label{sec:world_model}
World-model-based neural evolution instantiates the environment as a separately learned simulator that approximates environment dynamics and provides environment-like feedback for agent learning.
Existing approaches mainly differ in how the world model is obtained: some methods learn environment dynamics directly from interaction data~\cite{Code2World,WebEvolver,WebWorld}, while others leverage the prior knowledge of foundation models through prompting to simulate environment transitions at inference time~\cite{WebDreamer,UI-Simulator,WAC}.
These two directions introduce different trade-offs between training cost, scalability, and simulation fidelity.
\textbf{Learning-based World Models} obtain the simulator by explicitly learning environment transitions from interaction data.
Code2World~\cite{Code2World} models GUI transitions as renderable code, generating the next interface as HTML and aligning predictions with visual outcomes via render-aware reinforcement learning.
WebEvolver~\cite{WebEvolver} extends this paradigm by jointly evolving the world model and agent policy, where the learned model is used both to generate training trajectories and to simulate future interactions for planning.
WebWorld~\cite{WebWorld} further scales this line by training a web simulator on over one million real-world interaction trajectories, enabling long-horizon and structured state prediction at scale.
\textbf{Inference-based World Models} rely on the reasoning and prior knowledge of foundation models to simulate environment transitions without explicitly training a dedicated simulator.
WebDreamer~\cite{WebDreamer} demonstrates that a strong LLM can function as a web world model via direct prompting, predicting potential state changes for candidate actions and enabling model-predictive planning before real interaction.
UI-Simulator~\cite{UI-Simulator} constructs an LLM-based simulator that generates future UI states from prior context and actions, while guided rollouts and trajectory wrappers are introduced to synthesize scalable training data.
WAC~\cite{WAC} integrates a world model as an environment expert within a multi-agent pipeline, where simulated outcomes are used to support candidate action generation and pre-execution correction.
\subsection{Difficulty-Driven Evolution}\label{sec:diff_evo}
Complementary to neural-driven evolution, difficulty-driven environment evolution commonly adopts \textbf{curriculum learning}, where environments are progressively adjusted to match the evolving capabilities of agents.
% The goal is to maintain tasks near the agent’s capability frontier—avoiding tasks that are too easy to provide useful learning signals or too difficult to enable effective exploration.
Existing approaches can be broadly divided into two categories. 
The first relies on \textbf{explicit curriculum signals}, where environment difficulty is directly regulated by clearly defined signals such as accuracy, regret, reward or curiosity~\cite{RLVE,SCALER,GenEnv,CuES}. 
The second relies on \textbf{implicit curriculum mechanisms}, where curriculum progression is not explicitly controlled by a measurable signal but instead emerges from adaptive task generation, environment construction, or manually designed curricula~\cite{DreamGym,AgentGen,Reasoning_Core,Environment_Tuning}. 
\subsubsection{Explicit Curriculum Signals}\label{sec:Explicit Curriculum Signals}
Explicit curriculum signals regulate environment difficulty through clearly defined metrics that determine how tasks should be generated or adapted during training. 
%
% RLVE~\cite{RLVE}, SCALER~\cite{SCALER}, and GenEnv~\cite{GenEnv} all exemplify success-rate-aligned curriculum signals, though they differ in their adaptation mechanisms. 
% Most existing approaches adopt success rate as the primary signal to regulate environment difficulty, though they differ in their adaptation mechanisms.
Most existing approaches adopt performance-related signals as the primary basis for curriculum control, though they differ in their adaptation mechanisms.
A common strategy is to use success rate or accuracy as the main curriculum signal.
RLVE~\cite{RLVE} primarily performs forward difficulty evolution: each adaptive verifiable environment shifts its problem distribution toward harder instances once the empirical success rate on the current upper-bound difficulty exceeds a predefined threshold. 
SCALER~\cite{SCALER} adopts a broader co-adaptive strategy, using an online difficulty controller to adjust instance difficulty based on rollout accuracy so as to maintain training near a target success-rate band, while further reorganizing the training distribution through active environment curation. 
GenEnv~\cite{GenEnv} instead formulates this principle as a reward-driven curriculum policy, where the $\alpha$-Curriculum Reward encourages the generation of tasks whose empirical success rate stays close to a target band, thereby aligning difficulty with the agent's current capability.
EnvGen~\cite{EnvGen} extends this idea to game environments by adapting environment configurations according to task-specific success and failure statistics, so that the generated environments increasingly target the agent's weak skills.
Eurekaverse~\cite{Eurekaverse} similarly evolves embodied training terrains using policy training statistics, especially success rate and reward-related signals, to iteratively generate more suitable parkour environments.
Another important line of work comes from unsupervised environment design, where explicit signals such as regret are used to regulate the evolution of parameterized environments or replayable levels~\cite{PAIRED,DCD,MAESTRO,ACCEL,ReMiDi}.
PAIRED~\cite{PAIRED} uses regret between antagonist and protagonist returns to train an adversarial environment generator, thereby steering the environment toward tasks that are difficult yet still learnable.
ACCEL~\cite{ACCEL} further develops this paradigm in replay-based settings by editing and preserving high-value levels according to regret-style signals, enabling continual environment evolution while avoiding curriculum collapse.
Beyond direct performance signals, some methods rely on richer external feedback.
DataEnvGym~\cite{DataEnvGym} uses student feedback, such as errors and weak skills, to guide teacher-side data and environment generation, turning student performance into an explicit curriculum signal.
CuES~\cite{CuES} instead uses intrinsic curiosity as an explicit signal to drive exploration and task synthesis in environments without predefined tasks, enabling the agent to autonomously discover and organize useful training tasks.  
%
% Overall, these approaches make curriculum progression directly observable and controllable by grounding difficulty adaptation in explicit signals such as success rate, reward, or curiosity.
\subsubsection{Implicit Curriculum Mechanisms}\label{sec:Implicit Curriculum Mechanisms}
% In contrast, implicit curriculum mechanisms do not regulate difficulty through an explicitly defined signal. Instead, curriculum progression arises from the generation and organization of tasks and environments during training. 
Implicit curriculum mechanisms do not rely on explicit signals, with curriculum progression emerging from task generation and environment construction during training.
POET~\cite{POET} exemplifies this paradigm by jointly evolving environments and agents through mutation, minimal-criterion filtering, and transfer, allowing progressively more complex obstacle courses to emerge without an explicit difficulty controller.
DreamGym~\cite{DreamGym} induces curriculum through adaptive task generation, synthesizing tasks that challenge the current policy and prioritizing those with high reward entropy to construct progressively more informative training experiences.
AgentGen~\cite{AgentGen} similarly relies on task evolution by constructing tasks of increasing difficulty within generated environments and refining the progression through bidirectional evolution over both easier and harder instances.
Reasoning Core~\cite{Reasoning_Core} exemplifies curriculum emerging from environment construction. It emphasizes scalable symbolic reasoning environments with continuous difficulty control and large-scale problem generation, enabling increasingly challenging instances to arise as training progresses.
SEC~\cite{SEC} instead realizes curriculum progression through adaptive category selection, dynamically reallocating training focus across problem types during training without relying on a manually specified curriculum schedule.
Environment Tuning~\cite{Environment_Tuning} instead reflects manually designed curricula, organizing training through structured curricula, environment augmentation, and progress-based feedback.
%
% Overall, these approaches demonstrate that curriculum-like progression can emerge without explicit signal control, arising instead from task generation, environment construction, or manually designed curricula.

\subsection{Scaling-Driven Evolution}\label{sec:scaling_evo}
Beyond adjusting environment parameters or curriculum difficulty, scaling-driven environment evolution focuses on expanding the breadth of the \textbf{environment distribution} itself.
The goal is to expose agents to substantially richer interaction spaces by increasing either the diversity of scenarios within a given environment family or the coverage of fundamentally different domains and environment structures.
Existing approaches in this direction can be broadly divided into two categories.
The first emphasizes \textbf{scenario-level scaling}, where evolution is achieved by synthesizing more diverse tasks, trajectories, websites, or tool-use workflows within a shared interaction paradigm~\cite{EnvScaler,AutoForge,AgentScaler,InfiniteWeb,AutoWebWorld}.
The second emphasizes \textbf{environment-level scaling}, where evolution is driven by extending environments across multiple domains and applications, together with their transition, observation, and reward structures, thereby pushing agents toward broader cross-environment generalization~\cite{ARE,AutoEnv}.
\subsubsection{Scenario-Level Scaling}\label{sec:Scenario-Level Scaling}
Scenario-level scaling expands environment diversity primarily by increasing the number and variety of tasks, trajectories, and interaction scenarios within a shared environment paradigm.
In tool-use settings, several works~\cite{EnvScaler,AutoForge,AgentScaler} follow this paradigm by automatically constructing executable tool-interaction scenarios.
EnvScaler~\cite{EnvScaler} adopts a two-stage pipeline that first builds environment skeletons and then instantiates them into concrete scenarios with initial states, task descriptions, and rule-based validation.
AutoForge~\cite{AutoForge} further extends this pipeline by composing more challenging workflows from tool dependencies and reasoning constraints, while also introducing RL to stabilize training over synthesized environments.
AgentScaler~\cite{AgentScaler} provides a more structured formulation by abstracting function calling as read and write operations over databases, thereby systematically expanding the combinatorial space of tools, user intents, and execution paths.
A similar trend appears in GUI environments. 
InfiniteWeb~\cite{InfiniteWeb} enables scalable web-agent training by automatically generating functional websites from lightweight specifications, along with corresponding tasks and reward evaluators.
AutoWebWorld~\cite{AutoWebWorld} further improves controllability and verifiability by modeling websites as finite state machines, enabling systematic trajectory enumeration and verification over explicit state spaces.
%
% Overall, scenario-level scaling treats environment evolution as a problem of scenario expansion.
% By increasing the diversity of tasks, trajectories, and interaction patterns within a fixed environment paradigm, these methods provide agents with broader and more informative training experiences.
% \subsubsection{Domain-based Scaling}
\subsubsection{Environment-Level Scaling}\label{sec:Environment-Level Scaling}

By comparison, environment-level scaling extends the environment distribution itself by introducing diverse application settings and heterogeneous environment structures.
%
% ARE~\cite{ARE} and AutoEnv~\cite{AutoEnv} represent two complementary approaches to environment-level scaling, both focusing on expanding the environment space rather than increasing the number of task instances.
Current works focus on expanding the environment space rather than increasing the number of task instances.
ARE~\cite{ARE} provides a general platform for constructing and orchestrating agent environments across heterogeneous applications, supporting both synthetic environments and real-world application integration.
By enabling flexible configuration of rules, tools, content, and evaluation protocols, it expands the environment ecosystem in which agents are developed and evaluated.
AutoEnv~\cite{AutoEnv}, in contrast, offers a more fundamental formulation by modeling environments as factorizable distributions over transitions, observations, and rewards.
This abstraction allows heterogeneous environments with different dynamics and feedback structures to be generated in a unified manner, effectively scaling the space of environment factors and enabling systematic study of cross-environment learning and generalization.
%
% Overall, environment-level scaling reframes environment evolution as an expansion of the environment distribution itself.
% By exposing agents to diverse application settings and heterogeneous environment structures, these approaches move beyond scenario diversity toward a more fundamental form of generalization.
%
\subsection{Summary}

\begin{tcolorbox}[takeaway,title={Takeaway 7}]
\begin{itemize}
\item \textbf{Unified Perspective:}
% Taken together, these paradigms highlight a progressive shift in environment evolution: from improving how agents interact with a given environment (neural-driven), to controlling what difficulty of tasks agents experience (difficulty-driven), and ultimately to expanding which environments agents are exposed to (scaling-driven).
%
% Taken together, these paradigms highlight a progressive broadening of environment evolution:
% from representing the environment as a learned model (neural-driven),
% to adapting the difficulty of training environments to the agent's capability (difficulty-driven),
% and ultimately to expanding the environment distribution itself (scaling-driven).
%
Taken together, these paradigms reveal three complementary ways of evolving training environments for agent learning:
Neural-Driven Evolution represents the environment itself as a learned model,
Difficulty-Driven Evolution adapts environment difficulty to match the agent's capability,
and Scaling-Driven Evolution expands the environment distribution by broadening scenario diversity or environment structures.
%
% This progression provides a unified perspective on how environments can be systematically adapted to support agent learning.
% It also suggests that advancing agent capability requires not only stronger policies, but also treating environment evolution as a core component of the training process.
\item \textbf{Future Directions:} From this unified perspective, advancing agent capability requires not only stronger policies, but also treating environment evolution as a core component of the training process.
% Looking forward, training environments may need to evolve toward greater realism, with richer feedback, more diverse structures, and closer alignment with real-world interaction settings, eventually giving rise to \emph{environment engineering} as a systematic foundation for agent development.
Looking forward, training environments may need to evolve toward greater realism, with richer feedback, more diverse structures, and closer alignment with real-world interaction settings for agent development.

\end{itemize}
\end{tcolorbox}

\section{Challenges \& Future Directions}\label{sec:future}

\subsection{Environment-as-a-Service}

Current research faces a fundamental challenge due to the lack of standardization in environment interfaces. 
Existing environments differ substantially in their definitions of observations, action spaces, reward functions, and interaction paradigms, forcing agents to rely on environment-specific adaptations and limiting their ability to generalize across environments. 
Such heterogeneity also hinders unified benchmarking and undermines reproducibility.
In addition, environment deployment remains a major bottleneck for scaling agent systems.
Many environments rely on complex software infrastructures and tightly coupled runtime conditions, such as simulators, web browsers, graphical interfaces, or physical devices, making them difficult to reproduce and deploy across platforms. 
As a result, deployment overhead often becomes a system-level constraint, hindering large-scale training and evaluation.

To address these challenges, we advocate the concept of \textbf{Environment-as-a-Service (EaaS)}. The key idea is to standardize environment interfaces and provide them as cloud-hosted services, decoupling agent development from environment implementation and deployment. Under this paradigm, environments are encapsulated behind unified APIs, allowing researchers to interact with diverse environments through a consistent interface without managing underlying dependencies or runtime configurations.
% 统一规范化环境，环境即服务

\subsection{The Evolution of Environment Properties}
% 环境属性的演化

Current environments are still predominantly designed around static, short-horizon, closed-world, and single-modality settings.
While such formulations simplify environment construction and evaluation, they fail to reflect the complexity and variability of real-world scenarios.
In practice, real-world environments are often dynamic and continuously evolving.
Environment states may change while agents are interacting with them, requiring agents to make not only accurate but also timely decisions.
At the same time, many real-world tasks involve long-horizon interactions, where actions can influence future states far beyond immediate observations, significantly increasing the difficulty of planning and credit assignment.
Moreover, real environments are inherently open-ended, with continuously emerging tasks, expanding state spaces, and evolving interaction patterns, making it difficult for agents to rely solely on fixed task distributions.
In addition, the growing emergence of multimodal environments introduces diverse inputs, including text, images, videos, audio, and sensor signals, requiring agents to jointly understand and integrate heterogeneous modalities.
These trends fundamentally reshape the capability requirements of agents.
Accordingly, developing realistic environments that are dynamic, long-horizon, open-ended, and omni-modal will become a critical direction for advancing practical agent systems.

\subsection{From Single-agent to Multi-agent Environments}

Existing research has predominantly focused on single-agent environments, where the environment is typically modeled as a static and passive system. 
In such settings, an agent interacts with the environment independently, receives feedback, and optimizes its policy within a relatively controlled and static framework.
While this approach simplifies the problem, it fails to capture the complexity of real-world scenarios.
In practice, many tasks inherently involve multiple agents interacting within a shared environment, where decisions are interdependent and outcomes are jointly determined. Capturing such scenarios requires moving beyond single-agent environments to multi-agent environments, which better reflect the dynamics of cooperation, competition, and strategic interaction.

In multi-agent settings, the environment becomes a dynamic system shaped by interactions among agents. Each agent’s behavior influences others, leading to increased non-stationarity and complexity.
This shift introduces new challenges. Agents must continuously adapt their strategies in response to other agents, resulting in a more uncertain and evolving decision space. At the same time, inter-agent dependencies make credit assignment and stable reinforcement learning significantly more difficult. 
More importantly, complex multi-agent interactions may also give rise to emergent capabilities and behaviors that are difficult to observe in single-agent settings.
Understanding how such emergent behaviors arise from environment dynamics and agent interactions remains an important open problem for future research.

% 合成环境的可靠性，符号环境，神经环境的融合，sim2real gap

\subsection{Neural-Symbolic Environments}

Existing environment designs can be broadly categorized into neural and symbolic paradigms, each with distinct advantages and limitations.
Neural environments, typically built upon learned models or simulators, offer strong expressiveness and flexibility. They can capture complex, high-dimensional dynamics and support rich multimodal interactions. However, such environments are often opaque and lack interpretability, making it difficult to analyze underlying mechanisms or verify correctness. Moreover, neural environments may suffer from instability, distribution shift, and limited controllability, which can lead to unreliable feedback for agent learning.
In contrast, symbolic environments rely on rule-based systems or explicitly defined state transitions, providing clear semantics and high interpretability. Their structured nature enables precise control, reproducibility, and ease of analysis. However, symbolic environments are typically limited in expressiveness and scalability, making it challenging to model complex, real-world scenarios or handle high-dimensional observations.
These limitations highlight the need for neural-symbolic integration in environment design. By combining the expressiveness of neural models with the interpretability and controllability of symbolic systems, hybrid environments have the potential to provide both realistic dynamics and reliable structure. Developing such integrated environments remains an important direction for building robust and generalizable agent systems.

\subsection{Bridging the Sim-to-Real Gap in Environments}

Despite recent advances in generative and simulated environments, a substantial sim-to-real gap still exists between synthetic environments and real-world environments. 
Current environments are often constructed through hand-crafted rules or LLM-driven generation, aiming to provide scalable and controllable platforms for agent training and evaluation.
However, due to the inherent complexity, dynamics, and openness of the real world, existing synthetic environments still struggle to faithfully approximate real-world environments from multiple perspectives.

First, from the perspective of correctness, synthetic environments may contain factual errors and logical inconsistencies.
For example, webpages automatically generated by LLMs may exhibit invalid state transitions, incorrect feedback signals, or inconsistent interaction logic.
Such issues can lead agents to learn spurious policies that fail to generalize effectively to real-world environments.
Second, in terms of difficulty, many synthetic environments are simplified approximations of real-world problems.
To reduce construction and training costs, existing environments often simplify state spaces, action spaces, and task procedures, making it difficult to capture the long-horizon dependencies and complex interactions present in realistic scenarios.
As a result, agents that achieve strong performance in synthetic environments may still struggle to generalize effectively to more complex real-world tasks.
Moreover, regarding diversity, current synthetic environments are often constrained by predefined templates and fixed task distributions.
Compared with the continuously evolving and open-ended nature of real-world environments, synthetic environments typically provide limited scenarios and task distributions, which can easily cause agents to overfit to specific patterns.
From the perspective of fidelity, substantial discrepancies still exist between synthetic and real-world environments. 
In digital environments such as GUIs or web systems, real user behaviors, system latency, and dynamic interface changes are often insufficiently modeled. 
Therefore, an important future direction is to develop more realistic, diverse, and high-fidelity environments that better capture real-world complexity while maintaining scalability and controllability for large-scale agent training and evaluation.

\subsection{Co-Evolution of Agents and Environments}

Traditional research typically treats environments as fixed and passive systems, primarily focusing on how to optimize agent policies within predefined environments.
However, as environments become increasingly complex, the relationship between agents and environments is gradually evolving from one-way adaptation to bidirectional co-evolution.
Beyond requiring agents to adapt to environments, environments themselves should also evolve in response to agents by dynamically adjusting task difficulty and environment rules according to the agents' current capabilities, thereby forming a continual co-evolution mechanism.
For example, environments may progressively introduce more challenging tasks, longer interaction horizons, and increasingly complex feedback mechanisms as agent capabilities improve.
Environment generators may continuously create new tasks, tools, and interaction patterns based on the weaknesses or emerging capabilities of agents.

\subsection{Towards Unified Offline–Online Learning}
Current agent training methods can generally be categorized into two paradigms: offline training and online training.
Offline training relies on high-quality target trajectories generated by external expert models, allowing agents to imitate stronger policies while improving data efficiency and reducing the reliance on costly environment interaction.
However, it is limited by fixed data distributions and the lack of continual interaction with real environments, which can lead to distributional mismatch and reduced adaptability to evolving dynamics.
In contrast, online training relies on trajectories generated through the model’s direct interaction with environments, allowing policies to continuously adapt to evolving environment dynamics.
Such a paradigm more naturally reflects real-world agent behavior, but typically requires extensive interaction and often suffers from sparse rewards and high exploration costs.

Consequently, integrating offline and online paradigms is emerging as an increasingly important direction for agent training.
On-Policy Distillation (OPD) \cite{OPD} combines online trajectory generation with high-performing teacher supervision, where the student first explores through on-policy interaction and then receives dense corrective feedback from stronger teacher models.
However, the application of OPD to agentic settings remains largely underexplored.
Extending OPD to multi-turn agent environments introduces challenges fundamentally different from those in conventional single-turn tasks.
Errors made during early interaction steps not only affect the current output, but also alter subsequent environment states and future trajectories, leading to long-horizon error accumulation and increasingly inconsistent teacher supervision.
In addition, an important future direction is to investigate how to effectively integrate accurate environment feedback with fine-grained teacher supervision.

\subsection{Science of Environment Engineering}

Despite the rapid progress in agentic environments, current environment construction remains largely empirical and still lacks a systematic scientific foundation.
As environments become increasingly important for agent training, evaluation, and capability development, establishing a more principled understanding of environment engineering is emerging as a critical research direction.

One important future direction is to investigate the scaling laws of environments, namely how factors such as the number of environments, environment diversity, interaction horizon, and environment complexity influence the formation, generalization, and emergence of agent capabilities.
Understanding these relationships may help establish more principled guidelines for environment construction and scaling.
Beyond environment scale, another fundamental question is whether an environment itself is learnable by agents.
Not all environments are equally suitable for agent training, as overly sparse rewards, excessively large state spaces, or long interaction horizons may prevent agents from obtaining effective learning signals, leading to unstable training or even failure to converge.
Future research should therefore investigate environment learnability, namely what types of environments are more conducive to stable and efficient agent learning.
More importantly, environments and agent capabilities are inherently coupled.
Different environments may naturally encourage the development of different capabilities.
However, there is still a lack of systematic understanding of what types of environments cultivate what kinds of capabilities.
An important future direction is to establish the mapping between environments and agent capabilities, enabling targeted environment construction for developing specific meta-capabilities such as long-term memory, task decomposition, world modeling, and strategic planning.

\bibliography{reference}
% \bibliography{new_reference}
\bibliographystyle{IEEEtran}

% \appendix

% \section{Specific Evaluation Metric Formulas}\label{ap:metrics}

\vfill

\end{document}